\documentclass[journal]{IEEEtran}
\usepackage[utf8]{inputenc}
\usepackage[T1]{fontenc}

\usepackage{amsmath}
\usepackage{amsthm}
\usepackage{amsfonts}
\usepackage{amssymb}
\usepackage{mathrsfs}
\usepackage{picture}
\usepackage{graphicx}
\usepackage{times}
\usepackage{balance}
\usepackage{booktabs}
\usepackage{multirow}
\usepackage{slashbox}
\usepackage{rotating}
\usepackage{comment}
\usepackage{indentfirst}
\usepackage{enumerate}
\usepackage{lineno}
\usepackage{color}
\usepackage{url}
\usepackage{cite}
\usepackage{pdfpages}
\usepackage{indentfirst}
\usepackage{bm}
\usepackage{epsfig}
\usepackage{algorithm}
\usepackage{algorithmic}
\usepackage{tabu}
\usepackage{picins}

\newcommand{\ignore}[1]{}

\begin{document}

\title{Non-rigid Object Tracking via Deep Multi-scale Spatial-Temporal Discriminative Saliency Maps}
\author{Pingping~Zhang,
        Dong~Wang,
        Huchuan~Lu, \emph{Senior Member, IEEE},
        and~Hongyu~Wang, \emph{Member, IEEE}
\thanks{
Copyright (c) 2017 IEEE. Personal use of this material is permitted. However, permission to use this material for any other purposes must  be obtained from the IEEE by sending an email to \textcolor{blue}{\underline{pubs-permissions@ieee.org}}.

All the authors are with School of Information and Communication Engineering, Faculty of Electronic Information and Electrical Engineering, Dalian University of Technology, Dalian, 116024, P. R. China.
E-mail: jssxzhpp@mail.dlut.edu.cn; wdice@dlut.edu.cn; lhchuan@dlut.edu.cn; why@dlut.edu.cn. The corresponding author is Prof. Huchuan Lu.

PP. Zhang, D. Wang, and HC. Lu are supported in part by the Natural Science Foundation of China (NSFC), No. 61502070, No. 61528101 and No. 61472060. PP. Zhang is currently visiting the University of Adelaide supported by the China Scholarship Council (CSC) program.
}}
\markboth{Submitted to IEEE Transactions on Image Processing}%
{Shell \MakeLowercase{\textit{et al.}}:Interactive Video Segmentation via Local Appearance Model}
\maketitle
\begin{abstract}
In this paper we propose an effective non-rigid object tracking method based on spatial-temporal consistent saliency detection.
In contrast to most existing trackers that use a bounding box to specify the tracked target, the proposed method can extract the accurate regions of the target as tracking output, which achieves better description of the non-rigid objects while reduces background pollution to the target model.
Furthermore, our model has several unique features.
First, a tailored deep fully convolutional neural network (TFCN) is developed to model the local saliency prior for a
given image region, which not only provides the pixel-wise outputs but also integrates the semantic information.
Second, a multi-scale multi-region mechanism is proposed to generate local region saliency maps that effectively consider visual perceptions with different spatial layouts and scale variations.
Subsequently, these saliency maps are fused via a weighted entropy method, resulting in a final discriminative saliency map.
Finally, we present a non-rigid object tracking algorithm based on the proposed saliency
detection method by utilizing a spatial-temporal consistent saliency map (STCSM) model to conduct target-background classification and using a simple fine-tuning scheme for online updating.
Numerous experimental results demonstrate that the proposed algorithm achieves competitive performance in
comparison with state-of-the-art methods for both saliency detection and visual tracking, especially outperforming other related trackers on the non-rigid object tracking dataset.
\end{abstract}


\section{Introduction}

\IEEEPARstart{O}{bject} tracking aims to automatically identify the trajectories or locations of the moving objects in a sequence of images.
It is a very valuable research topic in the field of computer vision because of numerous advanced applications such as video surveillance, human-computer interaction and automatic driving.
Although decades of research on visual object tracking have emerged a diverse set of approaches and achieved satisfactory solutions under well-controlled environments, tracking generic objects has remained challenging.

In this paper we pay our attention to single-target tracking, which only provides the target region in the first frame and must inference new locations in the following frames.
Current efforts of single-target tracking~\cite{grabner2006real,ross2008incremental,babenko2009visual,
meixue_L1_2009,kwon2009tracking,kalal2010pn,zhang2014partial} mainly focus on building robust bounding box-based trackers
to overcome numerous inevitable factors, such as scale change, partial occlusion, illumination variation and pose
change.
To improve the tracking accuracy, a few researchers have shifted their efforts on non-rigid object tracking\footnote{Non-rigid object tracking also refers to segmentation-based tracking, which needs high pixel-wise accuracy.}, which is a more challenging task because this kind of tracking requires obtaining accurate target-background separations rather than coarse bounding boxes.
The existing non-rigid trackers often rely on the pixel~\cite{Aeschliman2010joint,Belagiannis2012Segmentation,Pixeltrack}, superpixel~\cite{Superpixel,multilevel} or patch-level~\cite{hough,ogbdt} classification.
For instance, PixelTrack~\cite{Pixeltrack} provides soft segmentations of the tracked object based on pixel-wise classification.
Superpixel tracker~\cite{Superpixel} treats the superpixels of specific objects as mid-level features and designs a simple discriminative model to generate a confidence map of the superpixels to achieve online tracking.
HoughTrack~\cite{hough} is able to identify the target area through patch-based classification and voting-based online Hough forest.
Recently, Son et al.~\cite{ogbdt} present an online gradient-boosting decision tree model to integrate a classifier operating on individual patches and generate segmentation masks of the tracked object.
However, all these methods are designed based on hand-craft features, which are not robust enough for complex object variations,
and they are not aware of the internal relations between the non-rigid tracking and salient object detection.

Visual tracking is essentially a selective attention procedure based on human visual systems; however, this saliency characteristic is often ignored in designing a tracking system.
As a basic pre-processing procedure in computer vision, salient object detection has shown great success for object re-targeting ~\cite{retargeting,re-targeting}, scene classification~\cite{scene} and semantic segmentation~\cite{grabcut}.
Significant progress has been made~\cite{ca,sr,cs,cd,object,thumbnailing}, however, several gaps still exist in applying the saliency detection algorithm to solve the tracking problem.
Saliency detection usually operates on the holistic image, and loses local specificity and scale consideration.
Conversely, visual tracking requires to focus on a specific object rather than the entire scene in a cluttered environment.
Several attempts~\cite{saltrack,connection,bit,dsm} have been performed to connect visual tracking and saliency detection; however, their generated saliency maps usually focus on enhancing the contrast of the center of the object and the local background.
These methods are suitable for the bounding box-based object tracking, but can not produce the segmentation-based outputs for non-rigid object tracking.
In this paper, we believe that \textbf{the goals of saliency detection and non-rigid object tracking are quite similar, i.e., producing pixel-wise outputs that distinguish the objects of interest from its surrounding background}.

\begin{figure*}
\begin{center}
\includegraphics[width=0.98\linewidth, height=8.2cm]{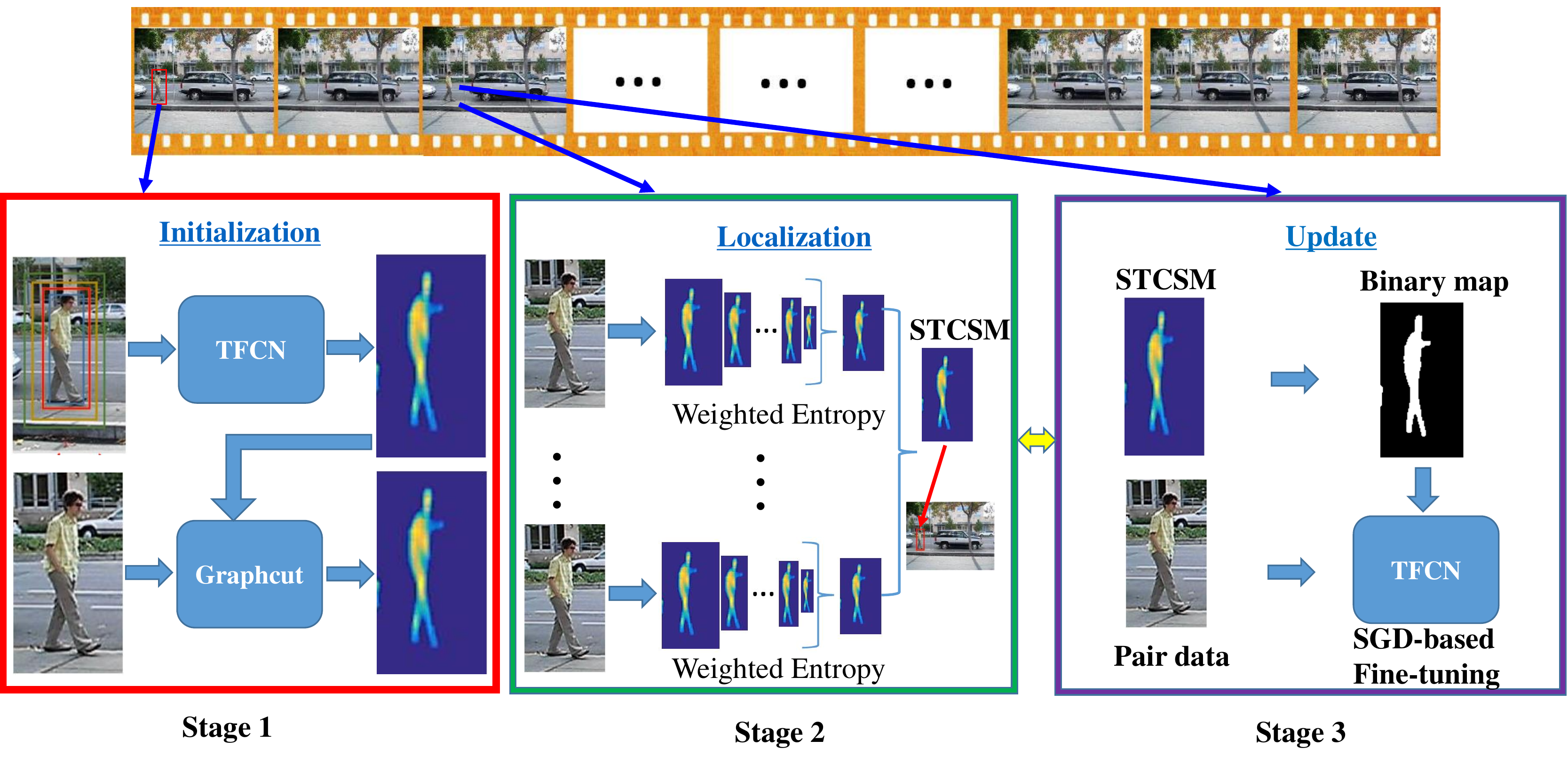}
\vspace{-4mm}
\caption{The overall framework of our proposed tracking approach. In the first stage, we use our proposed multi-scale multi-region image representation methods and the pre-trained TFCN to predict the local saliency maps in the start frame. The Graphcut~\cite{grabcut} is employed in the start frame to increase the robustness of saliency region predictions. Then we use the weighted entropy to fuse the local saliency maps, resulting in a discriminative saliency map for the target localization. During tracking, we use the STCSM model to incorporate spatial-temporal information into the location inference. After the localization, we perform thresholding on the STCSM and update the TFCN with the binary saliency map by SGD methods.}
\label{fig:framework}
\vspace{-6mm}
\end{center}
\end{figure*}

Inspired by the above-mentioned discussions, this paper proposes a novel non-rigid object tracking method based on
spatial-temporal consistent discriminative saliency detection.
The proposed method can extract the accurate regions of the tracked target as tracking output, which achieves better description of the non-rigid objects while reduces background pollution to the target model.
More specifically, we first develop a tailored fully convolutional neural network (TFCN), which is pre-trained on a well-constructed saliency detection dataset to predict the saliency map for a given image region.
Then, the proposed TFCN model utilizes local image regions with various scales and spatial configurations
as inputs, resulting in multiple local saliency maps.
Based on the weighted entropy method in~\cite{we}, these local saliency maps are effectively fused to produce a final discriminative saliency map for online tracking.
In addition, a structural output target-background classifier is built on the accumulated discriminative saliency maps, which effectively utilize the spatial-temporal information to generate pixel-wise outputs for depicting the state of the tracked object.
Finally, we extract the regions of interest (ROIs) and fine-tune the TFCN to obtain the local saliency map in the next frame. Fig.~\ref{fig:framework} illustrates the critical stages of the proposed tracking method.

In summary, the contributions of this work are as follows:
\begin{itemize}
\item An efficient TFCN is developed to produce the local saliency prior for a given image region, which not only provides the pixel-wise outputs but also integrates the semantic information of targets.
\item A multi-scale multi-region mechanism is presented to generate multiple local saliency maps and then fuse them through a weighted entropy method. This mechanism can produce a final discriminative saliency map for distinguishing the objects of interest from its surrounding background and facilitating the tracking process.
\item The saliency detection and visual tracking are integrated into a unified online learning framework based on only one deep convolutional neural network (CNN). The entire framework is jointly learned to optimize the task of simultaneous saliency detection and visual tracking.
\item The final outputs of the proposed tracker are pixel-wise saliency maps with the structural property and computational scalability, which is more suitable for the non-rigid object tracking problem.
\item Extensive experiments on public saliency detection and visual tracking datasets show that our algorithm achieves considerably impressive results in both research fields.
\end{itemize}


The rest of this paper is organized as follows.
In Section II, we give an overview of visual object tracking, salient object detection and their relationships in the perspective of deep learning.
Then we introduce the proposed spatial-temporal consistent saliency detection model in Section III, and present the non-rigid object tracking algorithm in Section IV.
In Section V, we evaluate and analyze the proposed method by extensive experiments and comparisons with other methods.
Finally, we provide conclusions in Section VI.
%
%
\section{Related Work}
Recently, deep leaning methods, especially deep CNNs can powerfully extract multi-level feature representations from raw images, leading to impressive performance for both visual tracking and saliency detection.
In this section, we briefly introduce several related works and discuss the relations between these two topics in the view of deep learning.
A complete survey of these methods is beyond the scope of this paper and we refer the readers to recent survey papers~\cite{borji2014salient,borji2015salient,otbs} for more details.
\subsection{Deep Learning for Visual Object Tracking}
In the visual tracking field, many practices indicate that the feature extractor plays the most important role in a tracker~\cite{wang2015understanding}.
Thus, the recent state-of-art trackers have taken advantages of the power of deep learned features.
For instance, Wang et al.~\cite{sade} offline train a stacked denoising autoencoder (SDAE) to learn generic image features that are robust
against visual variations. Then they transfer and fine-tune the trained model for the online tracking process.
Li et al.~\cite{deeptrack} exploit a CNN-based model for learning features concerning the tracked objects for visual tracking, which performs better than several representative trackers.
In addition, Wang et al.~\cite{fcnt} observe that the higher layers of pre-trained deep CNNs comprise abundant semantic information, whereas the lower layers include image details and considerable discriminative cues.
Thus, they extract feature maps of $conv4$-$3$ and $conv5$-$3$ layers in the VGG-16 model~\cite{vgg}, and build two shallow CNNs to capture
high-level and low-level information respectively.
These two CNN models are combined to generate confidence maps for localizing the tracked object.
Motivated by the same observation, Ma et al.~\cite{hcf} simply introduce deep features extracted from different convolutional layers (i.e. $conv3$-$4$, $conv4$-$4$ and $conv5$-$4$) of VGG-19~\cite{vgg} into the correlation filter framework and infer the location of the target using a coarse-to-fine mechanism.
Nam et al.~\cite{mdnet} pre-train a deep CNN using a large set of videos to obtain a generic target representation. When tracking a target in a new sequence, they construct a new network by combining the shared layers in the pre-trained CNN with a new binary classification layer.
To resolve the low-speed problem of deep learning-based tracker, David et al.~\cite{fct} proposed a simple feed-forward network to off-line learn a generic relationship between object motion and appearance. Then without online fine-tuning, they use the pre-trained model to track novel objects that do not appear in the training set.
Zhu et al.~\cite{ebt} employ an object proposal network that generates a small yet refined set of bounding box candidates
to mitigate numerous complications, such as object appearance changes, size and shape deformations.
Although deep learning-based tracker significantly improve the tracking accuracy, we note that almost all of them are designed based on the bounding box strategy.
Therefore, these trackers can not provide pixel-level tracking results, which limits their performance in the non-rigid object tracking problem.
Experimental evidences will be given in Section V.
\subsection{Deep Learning for Salient Object Detection}
Salient object detection aims to identify the most conspicuous objects or regions in an image. Since the revolution of deep learning in computer vision, salient object detection has made a great progress.
For instance, Wang et al.~\cite{legs} firstly propose two deep neural networks to integrate local estimation and global search for generating accurate saliency maps.
Zhao et al.~\cite{mc} consider both global and local contexts of images, and model the saliency detection procedure in a multi-context deep CNN framework, which overcomes the problem that salient objects may appear in a low-contrast background.
Motivated by the relationship of saliency detection and semantic segmentation, Li et al.~\cite{ds} propose a multi-task deep saliency method based on a FCN model with collaborative feature learning.
Subsequently, Li et al.~\cite{mdf} construct a saliency model based on multi-scale features of multiple deep CNNs, and improve the performance on several public benchmarks.
Lee \emph{et al.}~\cite{eld} propose to encode low-level distance map and high-level sematic features of deep CNNs for salient object detection.
Liu et al.~\cite{dhs} propose a deep hierarchical network for detecting salient objects. The network first makes a coarse global prediction. Then a novel hierarchical recurrent convolutional neural network (HRCNN) is adopted to refine the details of saliency maps step by step. The whole architecture works in a global to local and coarse to fine manner.
Wang et al.~\cite{rfcn} also develop a deep recurrent FCN to incorporate the coarse predictions as saliency priors, and stage-wisely refine the generated saliency maps.
Zhang et al.~~\cite{amulet} present a generic aggregating multi-level convolutional feature framework for salient object detection.
All of these methods demonstrate the effectiveness of deep CNNs in predicting saliency maps, however, they merely work on the static images and do not pay attention to the scale variations of objects, which is important and necessary in visual object tracking.
Therefore, it's harmful to the tracking performance by simply making use of these saliency detection methods.
Meanwhile, appropriately introducing dynamic information and solving the scale change in saliency detection are critical for saliency detection-based visual tracking.
\subsection{Relations between Visual Tracking and Saliency Detection}
Though visual tracking and saliency detection are always separately studied for different applications in computer vision,
research in psychology indicates that selective visual attention process or saliency detection is crucial to visual tracking~\cite{cp}.
Several attempts~\cite{saltrack,connection,bit,dsm} have been made to reveal the relations between visual tracking and saliency detection.
Mahadevan et al.~\cite{saltrack} firstly connect center-surround saliency detection and visual tracking, and present a biologically-inspired tracker.
They utilize numerous low-level visual features to boost the overall saliency map for bounding box-based target localization.
Borji et al.~\cite{ac} combine particle filters and bottom-up salient regions for adapting the object representation in complex scene context, resulting in higher tracking results than the basic approach.
Liu et al.~\cite{sa} present a novel visual attention shift tracking algorithm.
They first extracts a pool of salient objects or regions that have good localization properties from a salient
map.
Then, by the learned knowledge from historical data on the fly, the attentional selection process generates a
sequence of shifting between those objects and implements a detection of the target in them one by one.
Recently, Hong et al.~\cite{dsm} use a pre-trained R-CNN~\cite{rcnn} and feature back-projecting methods to generate the target-specific saliency map, then locate the tracked object on it.
Though the results are impressive, the saliency maps obtained by existing methods are all with respect to the center of the tracked object, which facilitates the bounding box-based trackers.
It is unsuitable in generating segmentation-based outputs for non-rigid object tracking.

This paper provides new insights and attempts for integrating saliency detection and visual tracking, and developing an effective non-rigid object tracker.
First, the tasks of saliency detection and non-rigid object tracking are quite similar, i.e., producing pixel-wise outputs that distinguish the objects of interest from its surrounding background.
Thus, conducting non-rigid object tracking using local saliency maps is reasonable.
Second, the proposed saliency detection method is developed based on a FCN, which can exploit the powerful deep features, introduce the rich semantic information, and facilitate the generation of segmentation based outputs.
Third, beside spatial information, motion information can also predict the location of the object.
If motion information is ignored or motion is inaccurately modeled, then tracking may fail.
To deal with this issue, we introduce the accumulated spatial-temporal saliency map, which can quickly captures the interesting object during tracking.
\section{Multi-Scale Local Region Saliency Model}
In this section, we start by describing the architectures of the famous fully convolutional networks (FCNs)~\cite{fcn} and the proposed TFCN network. Then we give the details of how to generate the discriminative saliency map based on our TFCN, which is competent in dealing with tracking problems.
\subsection{FCN Architectures}
The incipient FCN architecture~\cite{fcn} is an end-to-end, pixel-to-pixel learning model, which can produce a pixel-wise prediction and has been widely used for dense labeling tasks. The model differs from traditional CNN model because it essentially converts all fully-connected layers into convolution operators and use transposed convolutions for upsampling feature maps.
Specifically, the output of a convolutional operator is calculated by
\begin{equation}
    \textbf{Y}^{j}= f(\sum_{i}\textbf{X}^{i}*\textbf{W}^{i,j}+b^{j}),
  \label{equ:equ1}
\end{equation}
where the operator * represents 2-D convolution and $f(\cdot)$ is an element-wise non-linear activation function, e.g. ReLu ($f(x)=max(0,x)$). $\textbf{X}^{i}$ is the $i$-th input feature map and $\textbf{Y}^{j}$ is the $j$-th output feature map. $\textbf{W}^{i,j}$ is a filter of size $k\times k$ and $b^{j}$ is the corresponding bias term.
The transposed convolutions perform the transformation in the opposite direction of a normal convolution. In the FCN, transposed convolutions are used to project feature maps to a higher-dimensional space.

As shown in Fig.~\ref{fig:FCNs}(a), the FCN model first perform several layers of convolution and pooling on the image or feature maps to extract multi-scale feature representations of the image. Then the back-end layers perform several transposed convolutions that increases the resolution-reduced feature maps to the image size. Finally, the prediction is achieved through applying the pixel-wise classification with a Softmax function.
In~\cite{fcn}, the authors introduce several skip-connections, which add high-level prediction layers to intermediary layers to generate prediction results at multiple resolutions. The skip-connections significantly improve the semantic segmentation performance.
\subsection{Tailored FCN (TFCN)}
Our proposed TFCN is largely inspired by the FCN-8s~\cite{fcn} semantic segmentation model due to the two common characteristics between saliency detection and semantic segmentation.
First, the goals of saliency detection and semantic segmentation are relatively close. The goal of saliency detection is to extract the salient region from the background, whereas semantic segmentation is to distinguish different objects from the background.
Second, both tasks produce pixel-level outputs. Each pixel of the input image requires to be categorized into two or multiple classes. Thus, the pre-trained FCN-8s model can be utilized to provide prior information on generic objects.
The original FCN-8s model introduces two skip connections and adds high-level prediction layers to intermediary layers to generate prediction results at the resolution of image size.

\begin{figure*}
\begin{center}
\begin{tabular}{@{}c@{}c@{}c}
\includegraphics[width=0.275\linewidth,height=5.6cm]{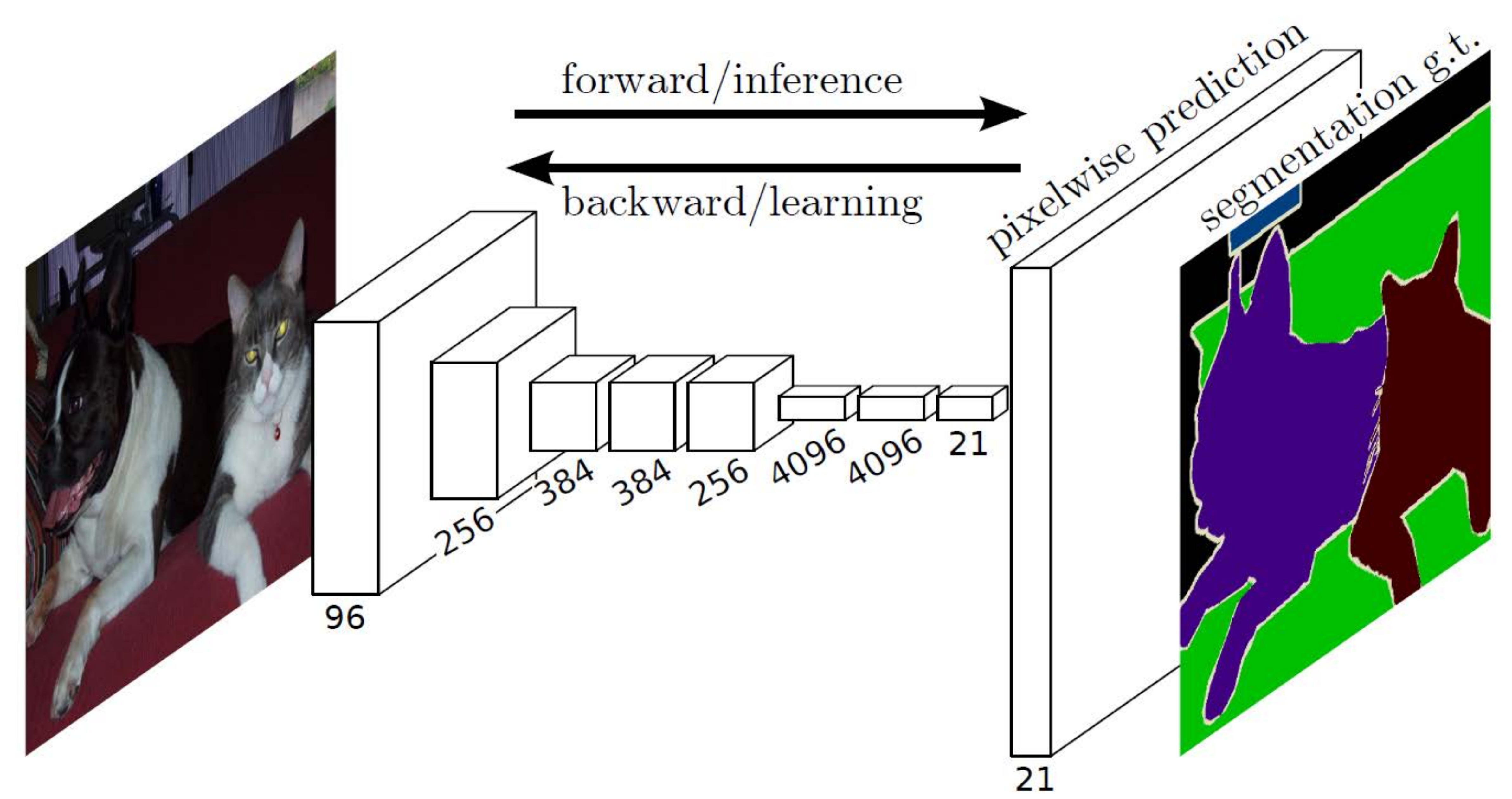} \ &
\includegraphics[width=0.30\linewidth,height=5.6cm]{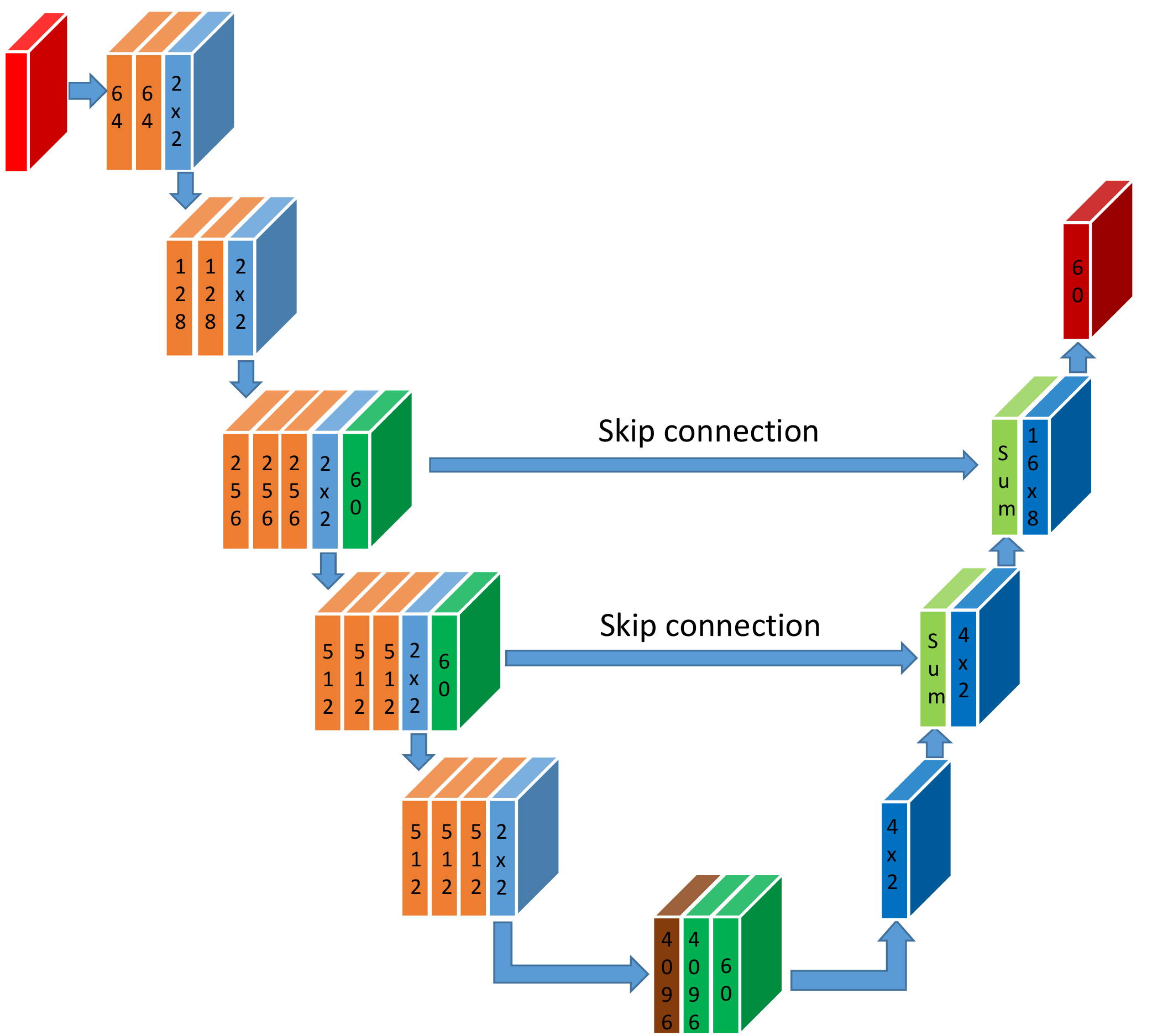} \ &
\includegraphics[width=0.425\linewidth,height=5.6cm]{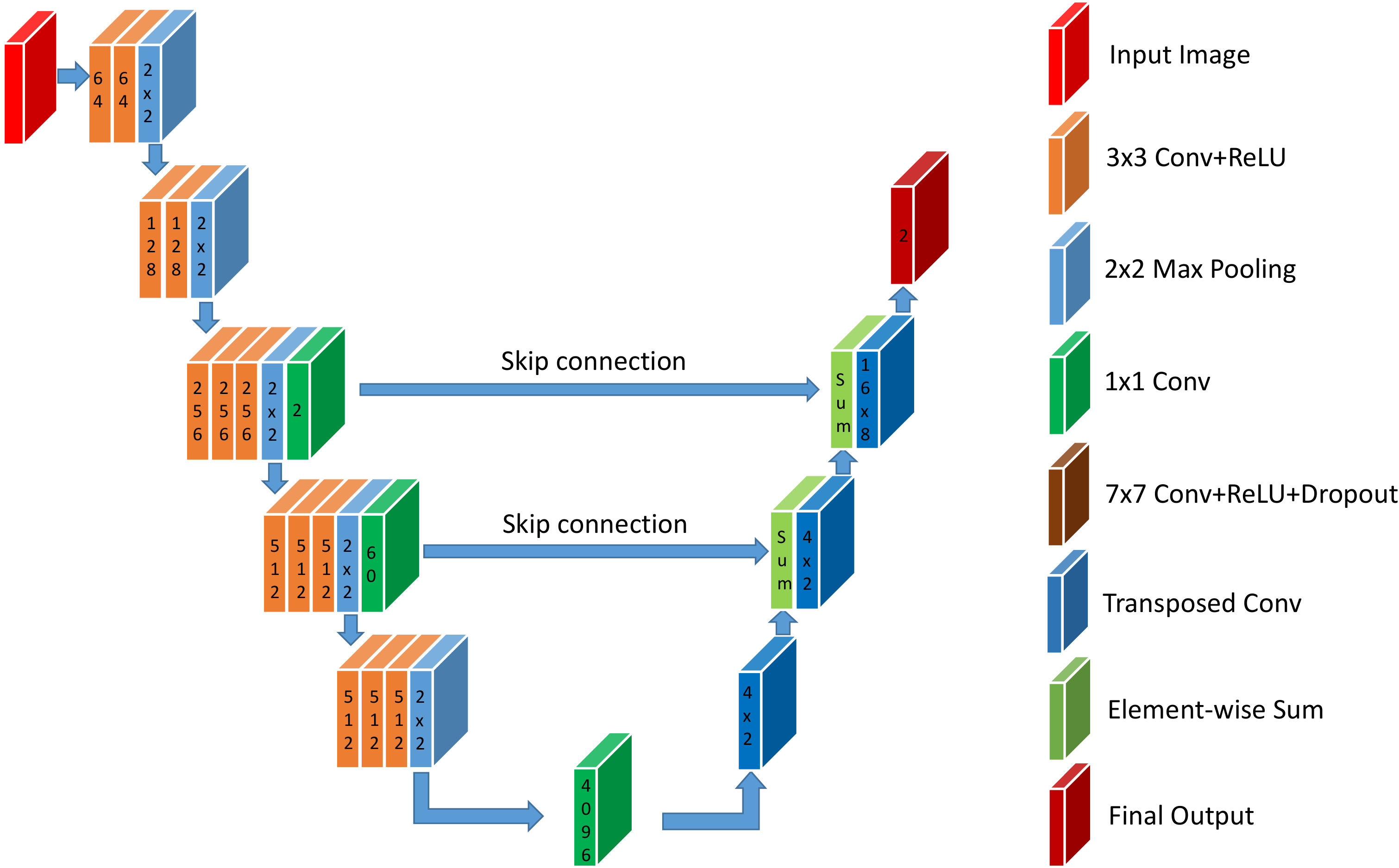} \\
(a) FCN &  (b) FCN-8s &  (c) TFCN\\
\end{tabular}
\vspace{-2mm}
\caption{Detailed comparisons of the FCN-8s~\cite{fcn} and our proposed TFCN model.
(a) Fully convolutional network (FCN). (b) The state of the art FCN-8s sematic segmentation model. (c) The proposed TFCN model for region saliency detection.
Conv stands for convolution. The number inside each layer indicates the amount of convolutional kernels.}
\label{fig:FCNs}
\vspace{-6mm}
\end{center}
\end{figure*}

We introduce scale considerations and modify the terminal structure of the original FCN-8s model to adapt to the saliency detection task and accelerate the training process.
The major modifications include: (1) changing the filter size from $7\times7$ to $1\times1$ in the $fc6$ layer to enlarge resolutions of feature maps and keep abundant details; (2) discarding the $drop6$, $fc7$, $relu7$ and $drop7$ layers because of their insignificant contribution for our tasks, and connecting the $fc6$ and $score59$ layers directly; (3) setting the $num\_out$ as 2 in the $upsample$-$fused$-$16$, $score$-$pool3$, and $upsample$ layers because the TFCN model is expected to predict the scores for two classes (salient foreground or general back-ground); and (4) keeping the $num\_out$ value unchanged in $score59$, $upscore2$, and $score$-$pool4$ layers for integrating moderate semantic information.
To be more precise, the Fig.~\ref{fig:FCNs}(b) and Fig.~\ref{fig:FCNs}(c) illustrate the detailed differences of the original FCN-8s and the proposed TFCN, respectively.
%
\subsection{Local Region Saliency Maps and Their Fusion}
We use the proposed TFCN to generate local saliency maps which can be used for visual tracking. The overall procedure comprises pre-training the TFCN, extracting scale dependent regions, and fusing discriminative saliency maps.
\subsubsection{Pre-training TFCN}
The TFCN model is derived from FCN-8s designed for the semantic segmentation task.
Thus, the direct application on salient object detection may lead to the negative transfer~\cite{transfer}.
To deal with this issue, we first pre-train the TFCN model based on a well-collected saliency dataset (described in Section V) before conducting local saliency detection.
Formally, given the salient object detection training dataset $ S=\{(X_n,Y_n)\}^{N}_{n=1}$ with $N$ training pairs, where $X_n =
\{x^n_j,j = 1,...,T\}$ and $Y_n = \{y^n_j,j = 1,...,T\}$ are the input image and the binary ground-truth image with $T$ pixels, respectively.
$y^n_j = 1$ denotes the foreground pixel and $y^n_j = 0$ denotes background pixel.
For notional simplicity, we subsequently drop the subscript $n$ and consider each image independently.
We denote $\textbf{W}$ as the parameters of the TFCN.
For the pre-training, the loss function can be expressed as
\begin{equation}
  \label{equ:equ2}
\begin{aligned}
  \mathcal{L}_f(\textbf{W})= - \beta \sum_{j\in Y_{+}} \text{log~Pr}(y_{j}=1|X;\textbf{W})\\
  -(1-\beta)\sum_{j\in Y_{-}} \text{log~Pr}(y_{j}=0|X;\textbf{W}),
\end{aligned}
\end{equation}
where $Y_{+}$ and $Y_{-}$ denote the foreground and background label sets, respectively.
The loss weight $\beta = |Y_{+}|/(|Y_{+}|+|Y_{-}|)$, and $|Y_{+}|$ and $|Y_{-}|$ denote the foreground and background pixel number, respectively.
Pr$(y_j =1|X;\textbf{W})\in [0,1]$ is the confidence score that measures how likely the pixel belong to the foreground.
The ground truth of each image in the saliency dataset is a 0-1 binary map, which perfectly matches the channel output of the TFCN model.
For the saliency inference, outputs of the last convolutional layer are utilized to distinguish saliency foreground from the general background.

Before putting the training images into the TFCN, each image is subtracted with the ImageNet mean~\cite{imagenet} and resized into the same size (500$\times$500).
For the correspondence, we also resize the 0-1 binary maps to the same size. Our pre-training uses stochastic gradient descent (SGD) with a momentum, learning rate decay schedule.
The detail settings of training parameters appear in the experiment section.
\subsubsection{Extracting Scale-dependent Regions}
\begin{figure}[t]
\footnotesize
\begin{tabular}{@{}c@{}c}
\includegraphics[width=0.43\linewidth]{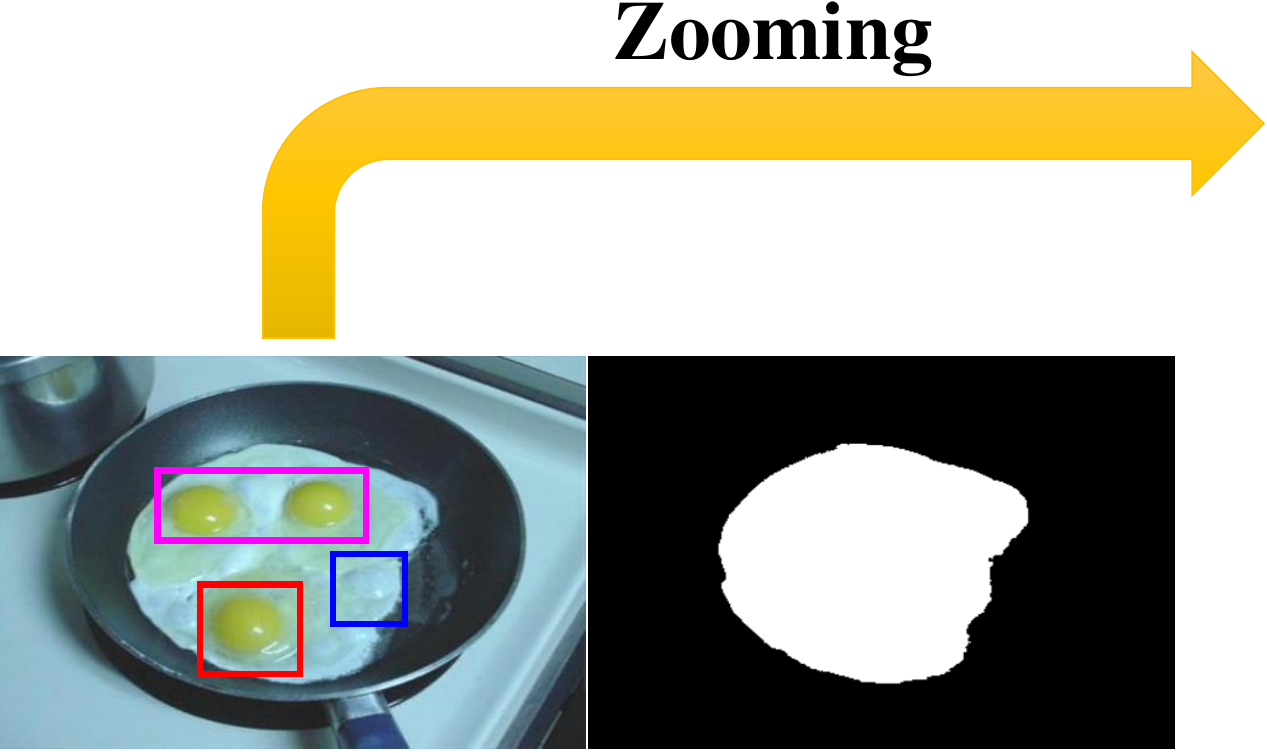} \ &
\includegraphics[width=0.56\linewidth]{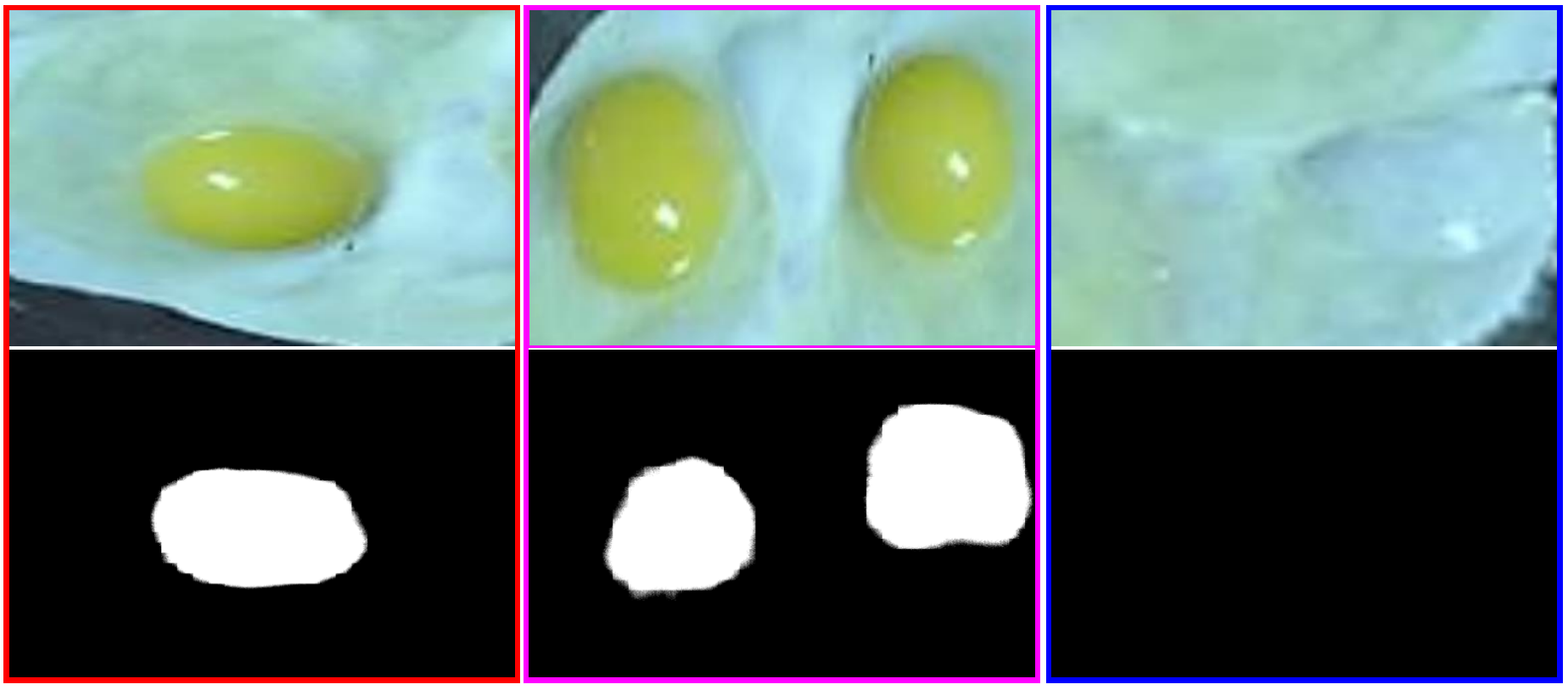} \\
(a) &  (b)\\
\end{tabular}
\vspace{-2mm}
\caption{Illustration of the effect of different received visual fields.
(a) Original image with the salient region. (b) Top: Three selected
typical regions, which are almost included by the overall salient region.
Bottom: Corresponding saliency maps.}
\label{fig:receivedfields}
\end{figure}

The human visual system suggests that the size of the received visual fields affects the fixating mechanisms significantly~\cite{rf}.
Fig.~\ref{fig:receivedfields}(a) shows a natural image with structural and hierarchical characteristics.
Human visual system focuses on the fried sunny-side up eggs and makes these regions salient as shown in the right.
If we zoom in a specific region (shown in Fig.~\ref{fig:receivedfields}(b)), we may select the area of the egg yolks as the most salient region or obtain an inconspicuous activation.
This sensibility of visual perception motivates us to present a novel image representation method and develop a saliency detection model based on the selected regions with different spatial layouts and scale variations.

\begin{figure}[t]
\begin{center}
   \includegraphics[width=\linewidth,height=5cm]{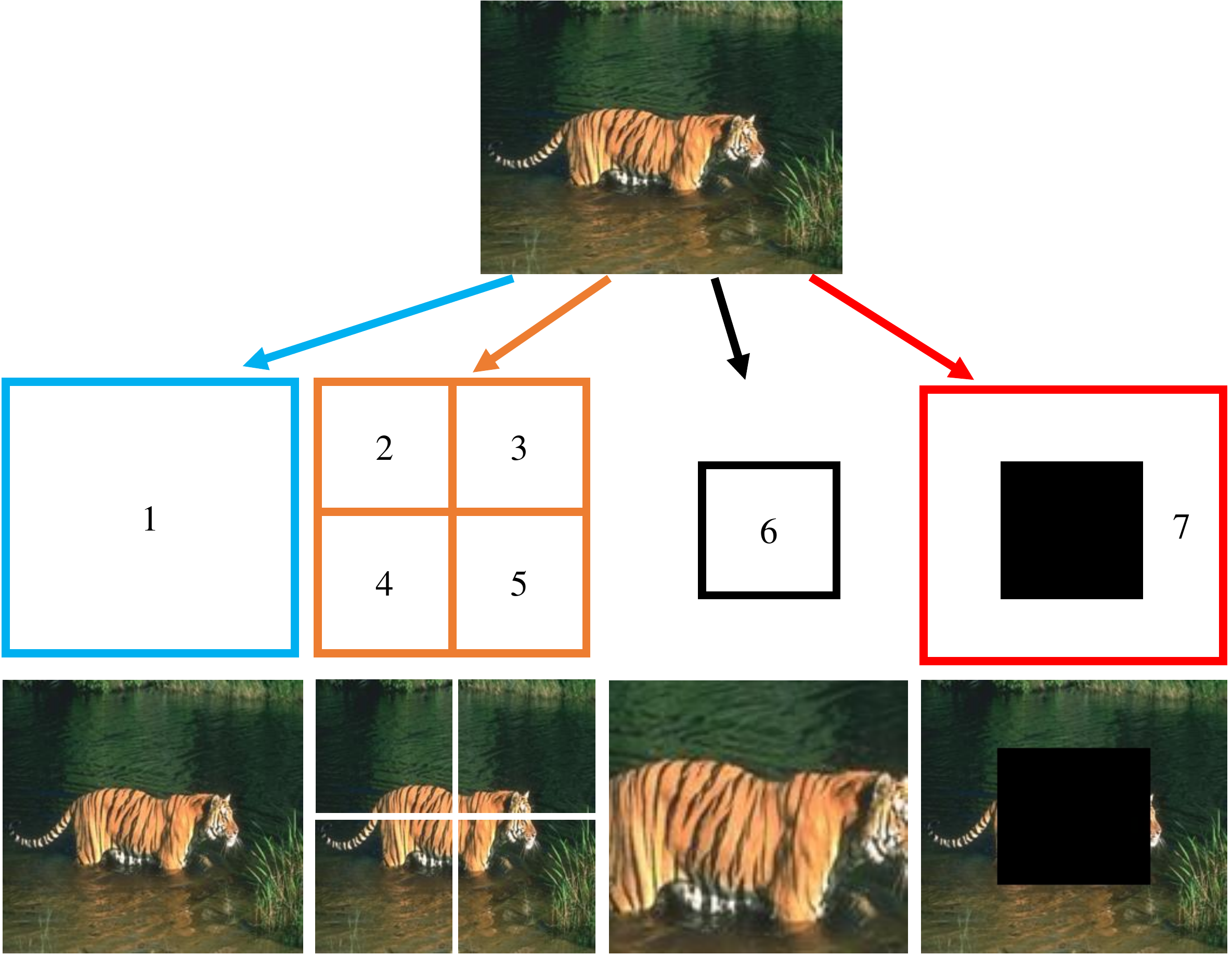}
\end{center}
\vspace{-2mm}
\caption{Illustration of the multi-region representation.}
\label{fig:parts}
\vspace{-4mm}
\end{figure}

More specifically, we exploit a multi-region target representation scheme shown in Fig.~\ref{fig:parts}.
We divide each image region into seven parts and calculate seven saliency maps over these regions.
The detailed configuration is as follows. (1) The first saliency map is obtained from the entire image region; (2) The next four saliency maps are calculated on the four equal parts to introduce spatial information. (3) The last two saliency maps are extracted from the inside and outside areas to highlight the scale support.
In addition, we introduce a multiple scale mechanism into each part to enhance the diversity of the region representation.
$N$ scales are sampled to generate multiple regions with different scales for a given part of the centered region $(l_{x}, l_{y})$ with size $(w_{0}, h_{0})$. In this study, the chosen sizes are $(n\times w, n\times h)$, where $w=\frac{3}{4}w_{0},h=\frac{3}{4}h_{0}$, and $n = 1,2, ..., N$.

\subsubsection{Fusing Saliency Maps}
\begin{figure}
\centering
\footnotesize
\begin{tabular}{@{}c@{}c@{}c@{}c@{}c@{}c@{}c}
\includegraphics[width=0.14\linewidth,height=1cm]{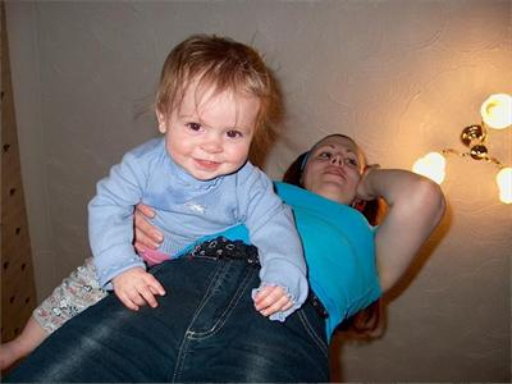}\ &
\includegraphics[width=0.14\linewidth,height=1cm]{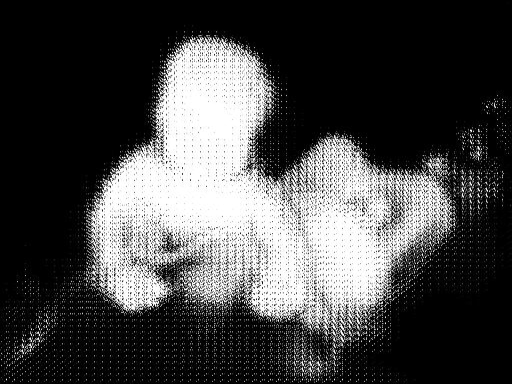}\ &
\includegraphics[width=0.14\linewidth,height=1cm]{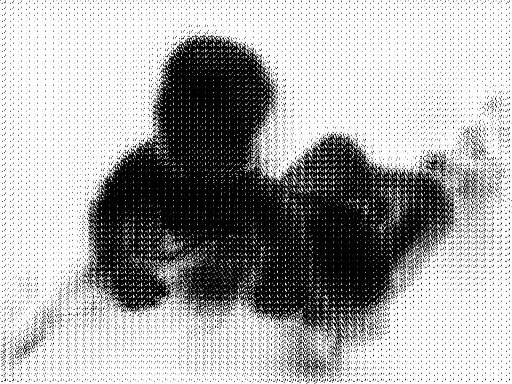}\ &
\includegraphics[width=0.14\linewidth,height=1cm]{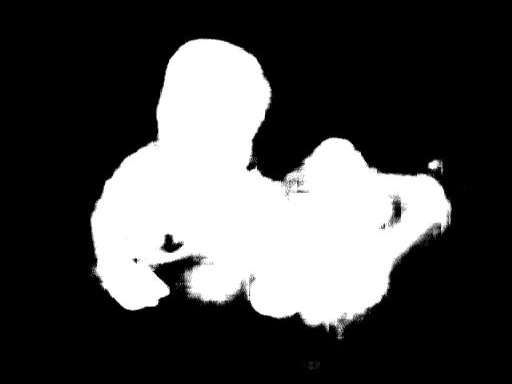}\ &
\includegraphics[width=0.14\linewidth,height=1cm]{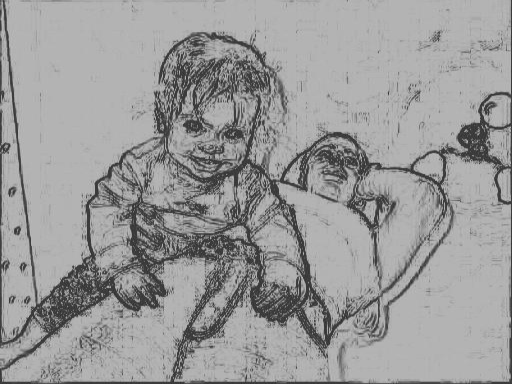}\ &
\includegraphics[width=0.14\linewidth,height=1cm]{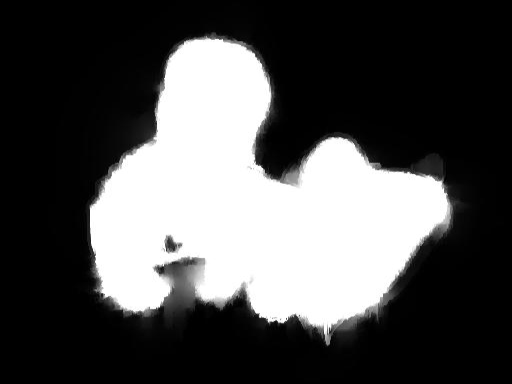}\ &
\includegraphics[width=0.14\linewidth,height=1cm]{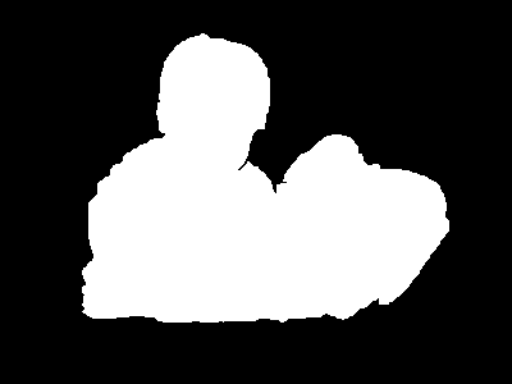}\\
(a) & (b) & (c) & (d) &(e) &(f) &(g) \\
\end{tabular}
\vspace{-2mm}
\caption{Illustration of saliency maps. From left to right: (a) Original image;
(b) Exaction map; (c) Inhibition map; (d) Saliency map; (e) Texture map generated
by fuzzy logical filters; (f) Final saliency map after domain transform; and (g) ground truth. }
\label{fig:fusessaliency}
\vspace{-6mm}
\end{figure}

Based on the proposed multi-scale multi-region scheme, we can obtain $M\times N$ saliency maps ($\textbf{S}_{m,n}, m = 1, ..., M, n = 1, ..., N$) in total, where $M$ denotes the number of regions for describing spatial layouts and $N$ is the number of sampled scales. Each saliency map $\textbf{S}_{m,n}$ is calculated based on the TFCN model, i.e.,
\begin{equation}
\textbf{S}_{m,n}  = padding(\textbf{M}^{E}_{m,n}-\textbf{M}^{I}_{m,n}),
  \label{equ:equ3}
\end{equation}
where $\textbf{M}^{E}_{m,n}$ and $\textbf{M}^{I}_{m,n}$ are the exaction map and the inhibition map obtained by the TFCN outputs, respectively. $padding(.)$ guarantees that different saliency maps are of equal sizes.
We note that $\textbf{M}^{E}$ and $\textbf{M}^{I}$ have reciprocal properties and the proposed strategy is able to eliminate several noises introduced by transposed convolution operations, shown in Fig.~\ref{fig:fusessaliency}(b-d).
Then, we adopt an additive rule to integrate the information of different regions for each scale,
\begin{equation}
\textbf{S}_{n}  = max(\sum_{m=1}^{M}\textbf{S}_{m,n},0),
  \label{equ:equ4}
\end{equation}
where the $max(.)$ operator avoids model degradation.

Finally, we exploit a weighted strategy to combine multiple saliency maps with different scales effectively,
\begin{equation}
\textbf{S}  = \sum_{n}w_{n}\textbf{S}_{n},
  \label{equ:equ5}
\end{equation}
where $\textbf{S}$ is the fused saliency map, $w_{n}$ is the weight of the $n$-th scale ($w_{n}\ge 0$, $\sum\limits_{n}w_{n}=1$). Generally, a more important saliency map has a large weight assigned to it. Thus,
we utilize the weighted entropy to measure the discriminative power of the fused map. Let $\textbf{w} = [w_{1}, w_{2}, .., w_{n}]^{T}$, the weighted entropy is defined as,
\begin{equation}
H(\textbf{w})  = -\Gamma(\textbf{w})\sum_{i}s_{i}^{\alpha+1}\ln s_{i},
  \label{equ:equ6}
\end{equation}
where $\Gamma(\textbf{w}) = 1/\sum_{i}s_{i}^{\alpha}$ denotes a normalization term, $\alpha$ denotes a constant and $s_{i}$ is a function of $\textbf{w}$, here we choose the function (\ref{equ:equ5}). To reduce computational complexity, we set $\alpha = 1$ and obtain $\Gamma(\textbf{w}) = 1/\sum_{i}s_{i}^{\alpha}=1$.

According to~\cite{wes}, a small weighted entropy represents that the saliency map is highly different from others, indicating more discrimination.
The optimal weight vector $\textbf{w}$ can be obtained by minimizing the objective function (\ref{equ:equ6}), which can be effectively solved based on the iterative gradient descent method in~\cite{we}.
After obtaining the fused saliency map $\textbf{S}$, we utilize the domain transform technique to enhance its spatial consistency.
Specifically, a high-quality edge preserving filter~\cite{dt} is performed on $\textbf{S}$ with the texture map generated by fuzzy logical filters. As shown in Fig.~\ref{fig:fusessaliency}(d-f), several holes or disconnected regions can be filled after domain transform.
\section{Non-Rigid Online Tracking Model}
In this section, we present the proposed non-rigid object tracking method in detail.
We develop our online tracking method based on the resulting saliency map in Section III.
We will show how to initialize the state of target objects with the discriminative saliency map and how to utilize spatial-temporal saliency information for object tracking.
The overall framework of the proposed method is shown in Fig.~\ref{fig:framework}. The critical components and discussions are presented as follows.
\subsection{Tracker Initialization}
Given the center location $(x_1,y_1)$ and specified region $R_1$ of the target object in the first frame, we first crop an image patch $P_1$ centered at the target location with the 1.5 times target size ($W_1\times H_1$), calculated on the specified region.
Subsequently, image patch $P_1$ is put into the pre-trained TFCN model to generate a saliency map $\textbf{S}_{1}$ through feed-forward propagation.
To improve the robustness of saliency map, Grabcut~\cite{grabcut} is conducted to obtain a foreground mask $\textbf{M}_{1}$ using the map $\textbf{S}_{1}$ as a prior.
Finally, the TFCN model is fine-tuned based on the intersection region of $\textbf{S}_{1}$ and $\textbf{M}_{1}$ maps in the first frame, which provides more accurate information of the target.
Note that we only use the Grabcut in the first frame.
\subsection{Target Localization}
In visual tracking, the shape and deformation information of the tracked object in previous frames can be utilized to predict the new state in the current frame because of the assumption of spatial-temporal consistency.
After cropping the same region in the $t$-th frame, we determine the state of the tracked object based a spatial-temporal consistent saliency map (STCSM) model, which is defined as
\begin{equation}
\textbf{S}^{STC}_{t}  = \textbf{S}_{t}+\sum_{k=t-\tau-1}^{t-1}\beta(k)\textbf{S}^{STC}_{k}, \beta(k)= \frac{1}{c^{k}},
  \label{equ:equ7}
\end{equation}
where $\textbf{S}_{t}$ denotes the saliency map in the current frame, which can be obtained by the saliency detection method presented in Section III.C.
$\textbf{S}^{STC}_{t}$ is the STCSM model up to the $t$-th frame, $\tau$ is the accumulated time interval and $\beta(k)$ corresponds to the weights of previous STCSM models with a decay factor $c$.
We note that $\beta(k)$ imposes higher weights for recent frames and lower weights for previous frames.
In addition, thanks to the availability of the TFCN, $\textbf{S}_t$ can be used in the generation of the accumulated saliency map before determining the location of the target object.
This is very different from most of existing trackers which only take previous information into account and directly detect the location on current frame.
The proposed accumulated saliency maps are also pixel-wise maps, having the structural property and computational scalability.
The structural property maintains more discriminative information between backgrounds and the target object.

For target localization, we can directly treat the saliency region obtained by the STCSM method as the tracking result in the current frame.
The obtained saliency region provides not only accurate tracking states but also detailed segmentation masks concentrating the tracked object (Fig.~\ref{fig:comparison}(d)).
This strategy gives more accurate tracking states which is very appropriate for non-rigid object tracking.
In addition, we can also exploit a compact rectangle including the overall saliency region and consider the center of this rectangle as the location of the tracked object.
The latter strategy facilitates the fair comparisons between our method and many bounding box-based trackers.
\subsection{Online Update}
To update the proposed tracker for online adaptation, we first convert the STCSM into a binary maps using a thresholding operator and treat this binary map as ground truth in the current frame.
Subsequently, we fine-tune the TFCN model from the $upsample$-$fused$-$16$ layer to the $loss$ layer with the SGD algorithm to enhance the adaptiveness of the tracker effectively.
Because we have one labeled image pair in each frame, fine-tuning the TFCN only with this image pair tends to overfitting.
Thus, we employ data augmentation by mirror reflection and rotation techniques. Simultaneously we also use the tracked regions of recent 20 frames for the fine-tuning.
The learning parameters can be found in Section V.B.
The overall process of our tracking system is summarized as Algorithm 1.
\vspace{-4mm}
\begin{table}[htbp]
\begin{tabular}{p{0.95\linewidth}}
    \toprule
    \textbf{Algorithm 1:} Our non-rigid object tracking approach\\
    \midrule
    \textbf{Input: } Frames ${I_{t}}$, $t\in [1,T]$, initial location $(x_1,y_1)$ and region $R_{1}$.\\
    \textbf{Output: } Object Mask $O_{t}$ and compact bounding box ${B_{t}}$, $t\in[2,T]$.\\
    \textbf{1: }  \textbf{for} each $t =2,...,T$ \textbf{do}\\
    \textbf{2: }  \quad Extract patch $P_t$ based on $R_{t-1}$ by image cropping.\\
    \textbf{3: }  \quad Fuse saliency maps by\\
    \textbf{4: }   \qquad1)\space Feed-forward the TFCN with $P_t$ using method in Section III.C.\\
    \textbf{5: }   \qquad2)\space Minimize the objective in Eq.~\eqref{equ:equ6} to seek the weights $\textbf{w}$.\\
    \textbf{6: }   \qquad3)\space Obtain fused saliency map $S_t$ by Eq.~\eqref{equ:equ5}.\\
    \textbf{7: }   \qquad4)\space Perform domain transform on $S_t$.\\
    \textbf{8: }  \quad Compute $S_{t}^{STC}$ by Eq.~\eqref{equ:equ7}.\\
    \textbf{9: }  \quad Obtain $O_t$ by thresholding $S_{t}^{STC}$.\\
    \textbf{10: }  \quad Obtain ${B_{t}}$, center location $(x_t,y_t)$ and $R_{t}$ through $O_t$.\\
    \textbf{11: } \quad Augment labeled image pairs.\\
    \textbf{12: }  \quad Fine-tune the TFCN with augmented labeled image pairs.\\
    \textbf{13: }  \textbf{end}\\
    \bottomrule
\end{tabular}
\label{alg:Alg1}
\vspace{-6mm}
\end{table}
\subsection{Differences with Existing Works}
\begin{figure}
\footnotesize
\begin{center}
\begin{tabular}{@{}c@{}c@{}c@{}c}
\vspace{-0.5mm}
\includegraphics[width=0.245\linewidth, height=0.18\linewidth]{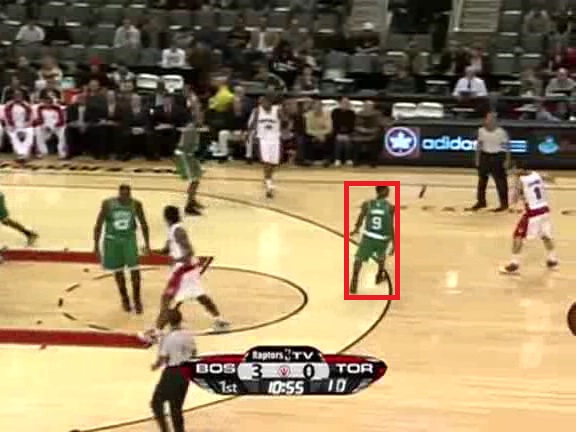}\ &
\includegraphics[width=0.245\linewidth, height=0.18\linewidth]{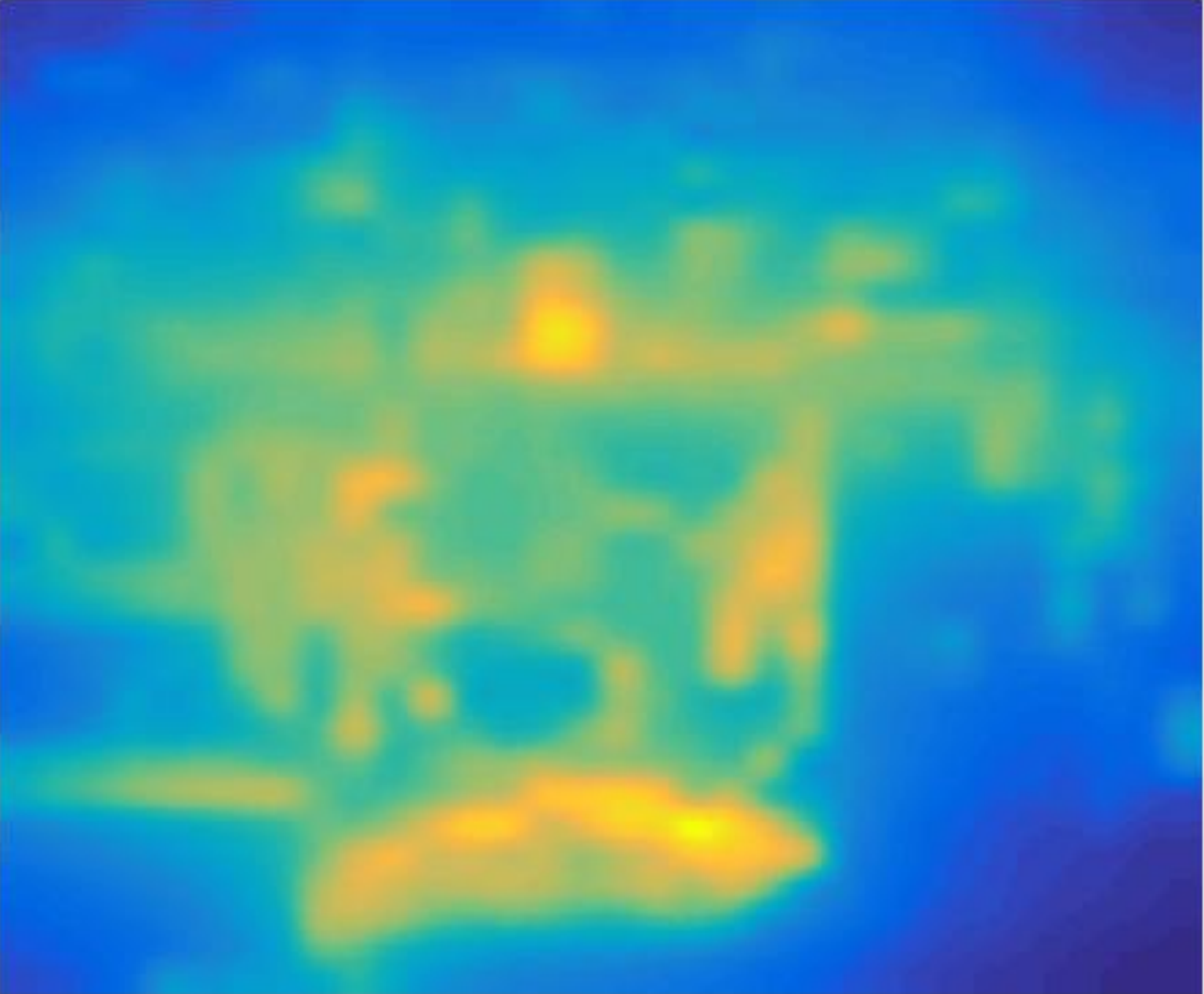}\ &
\includegraphics[width=0.245\linewidth, height=0.18\linewidth]{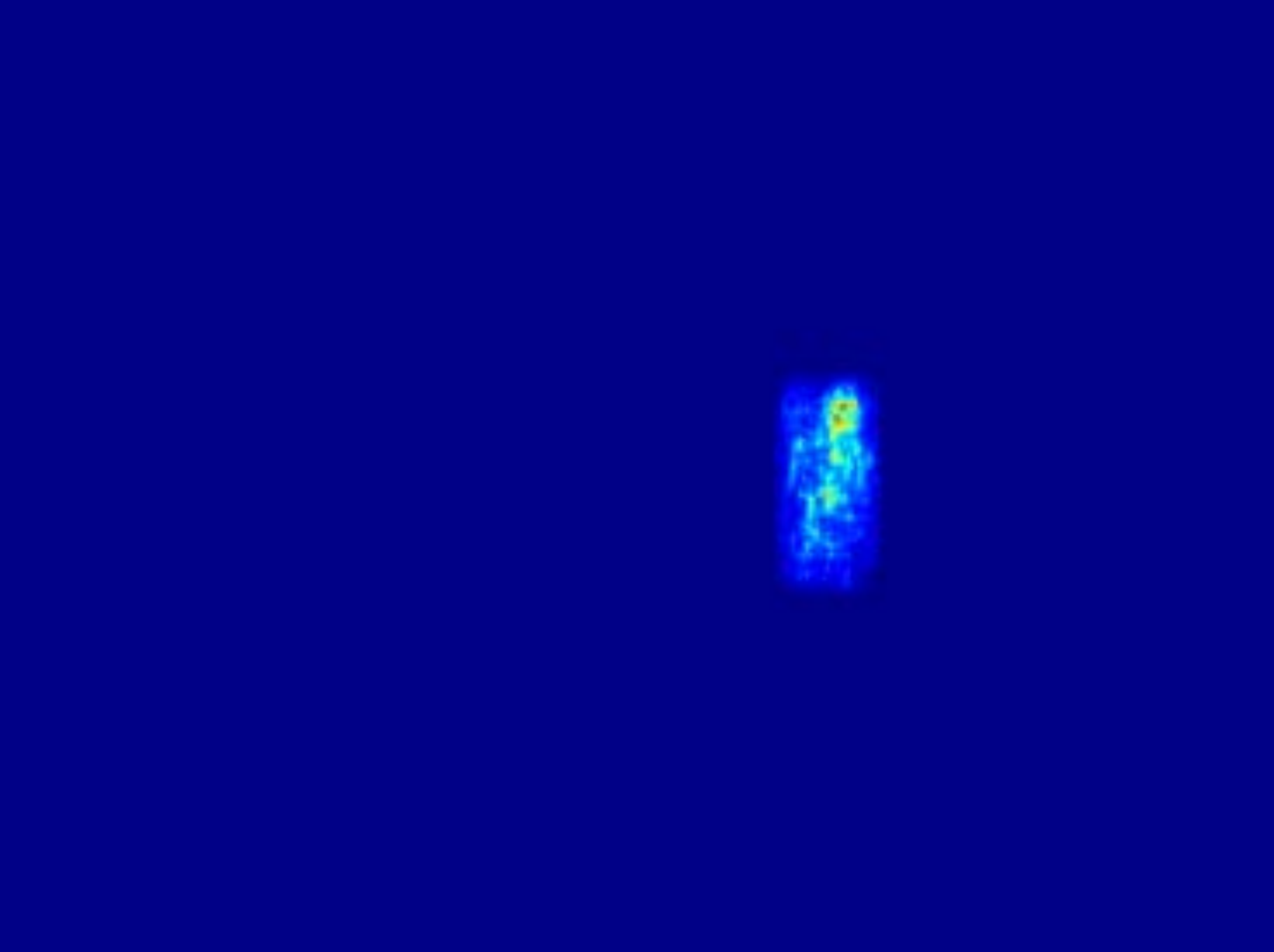}\ &
\includegraphics[width=0.245\linewidth, height=0.18\linewidth]{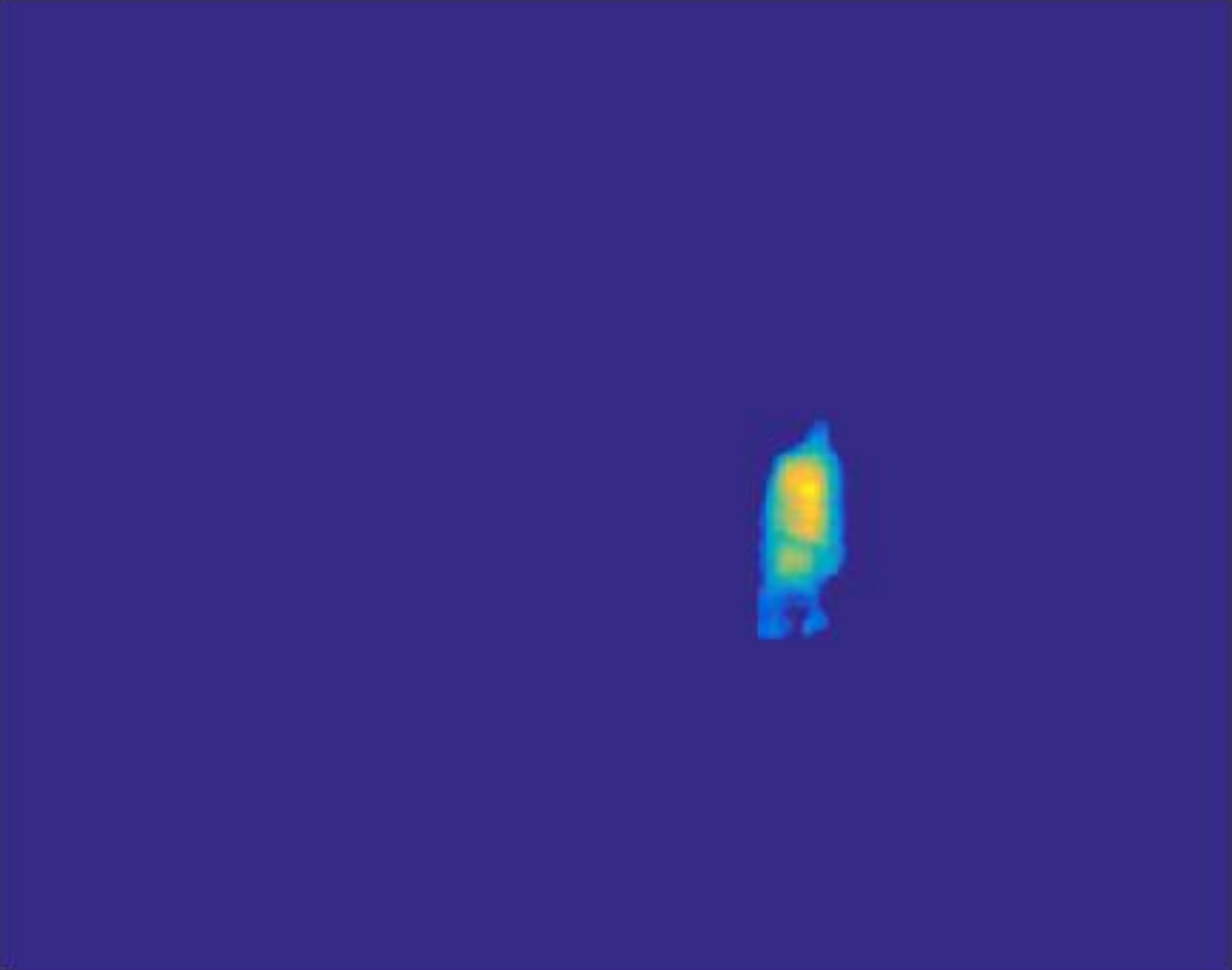}\\
\vspace{-0.5mm}
\includegraphics[width=0.245\linewidth, height=0.18\linewidth]{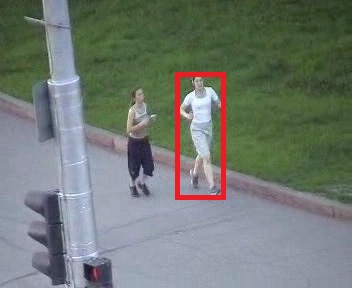}\ &
\includegraphics[width=0.245\linewidth, height=0.18\linewidth]{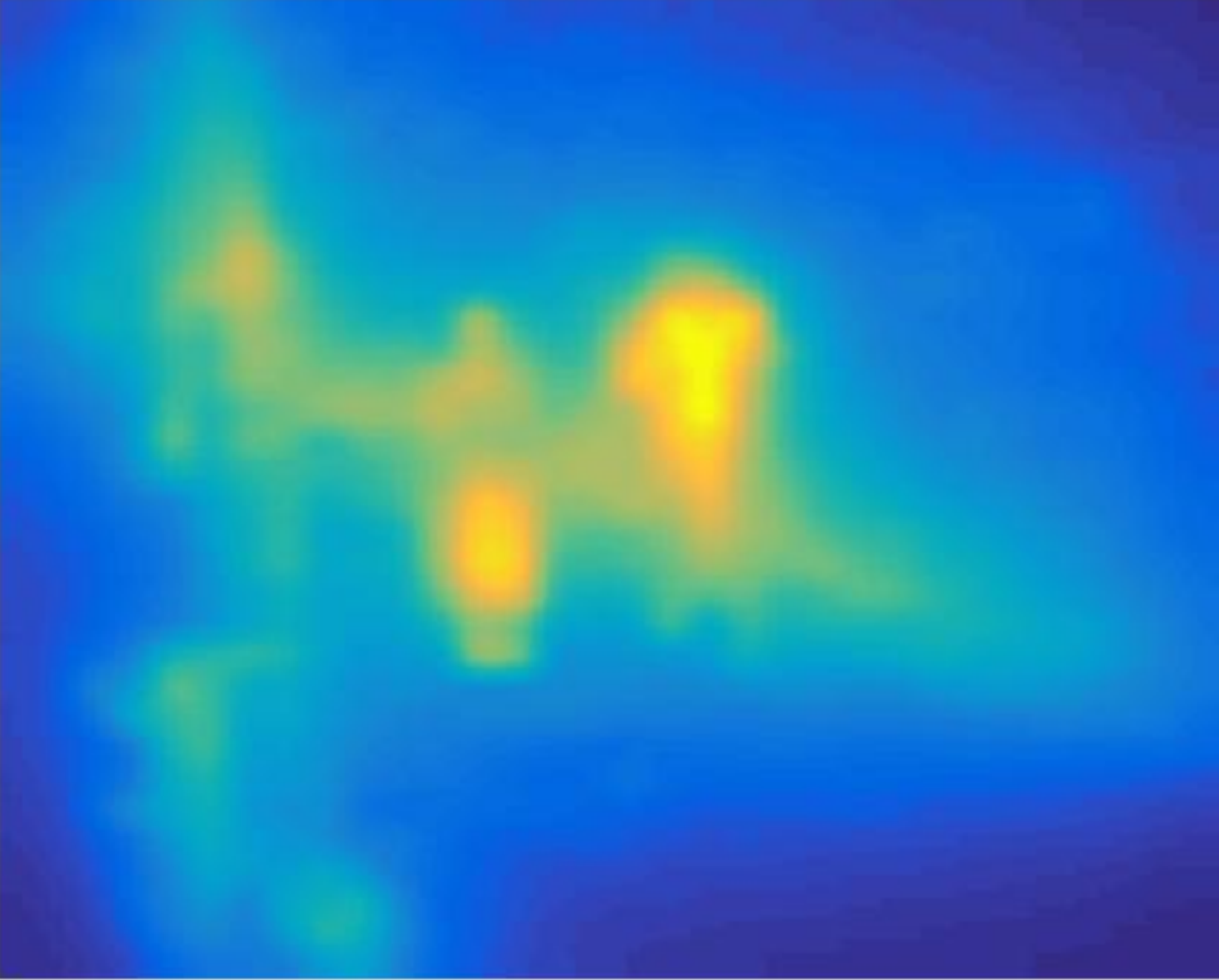}\ &
\includegraphics[width=0.245\linewidth, height=0.18\linewidth]{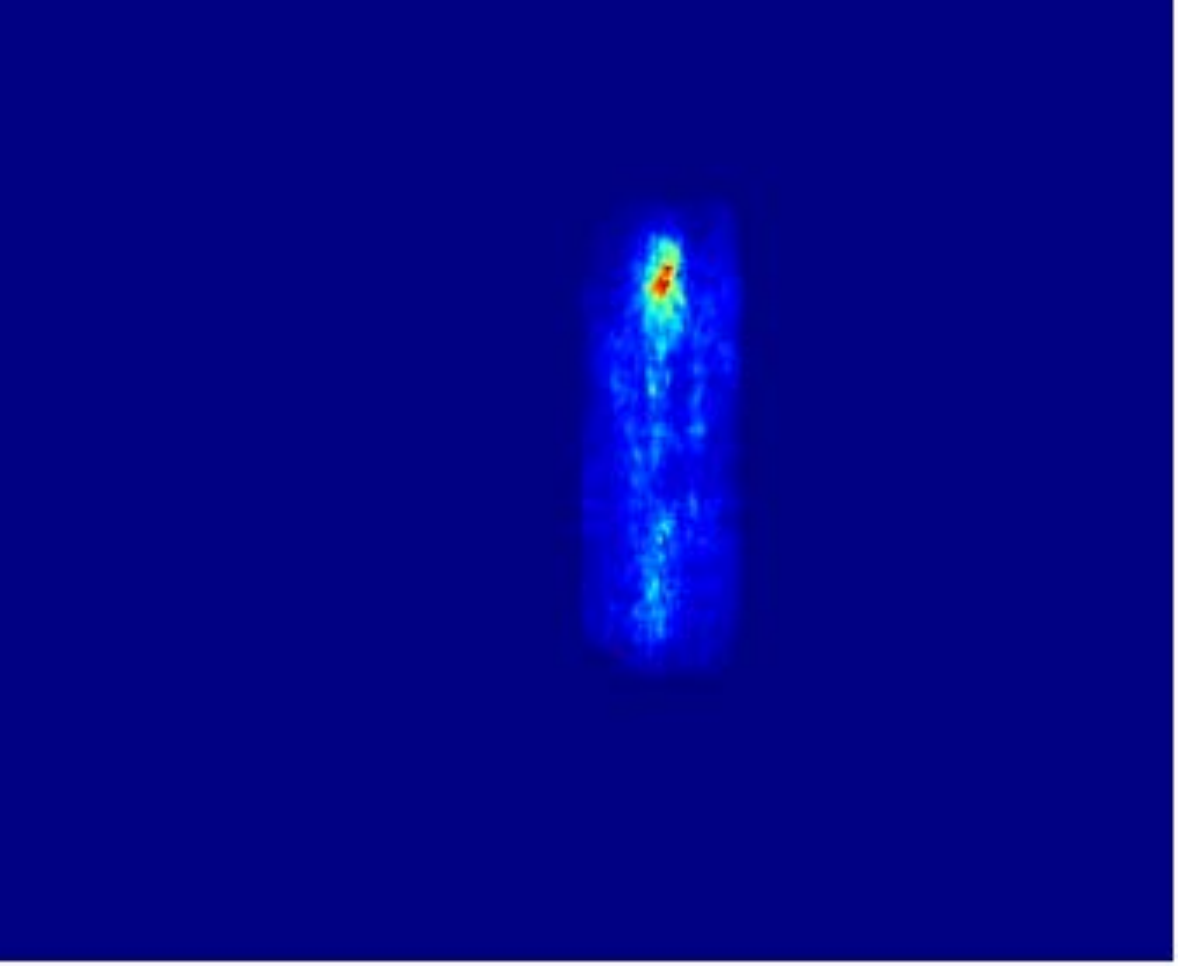}\ &
\includegraphics[width=0.245\linewidth, height=0.18\linewidth]{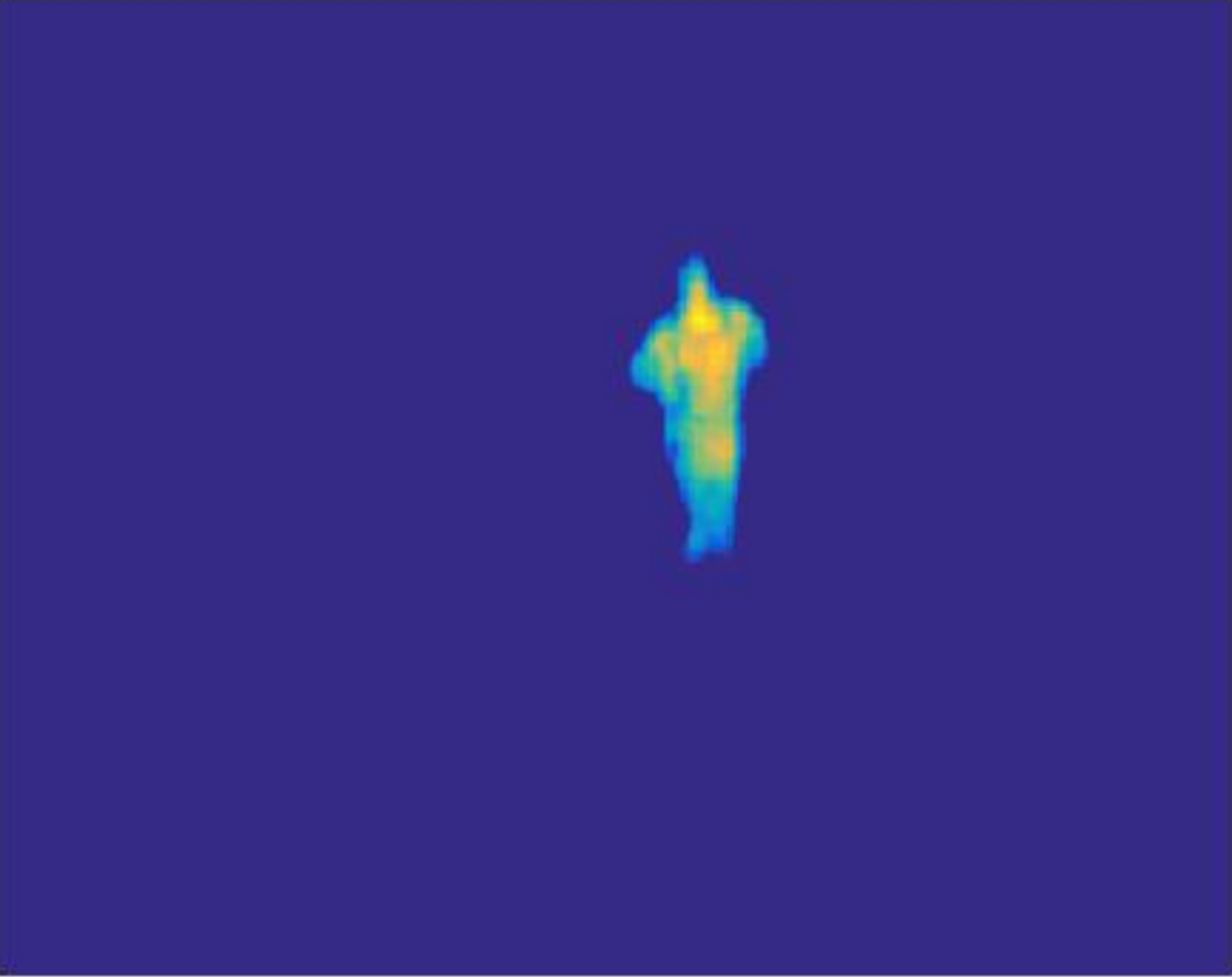}\\
\vspace{-0.5mm}
\includegraphics[width=0.245\linewidth, height=0.18\linewidth]{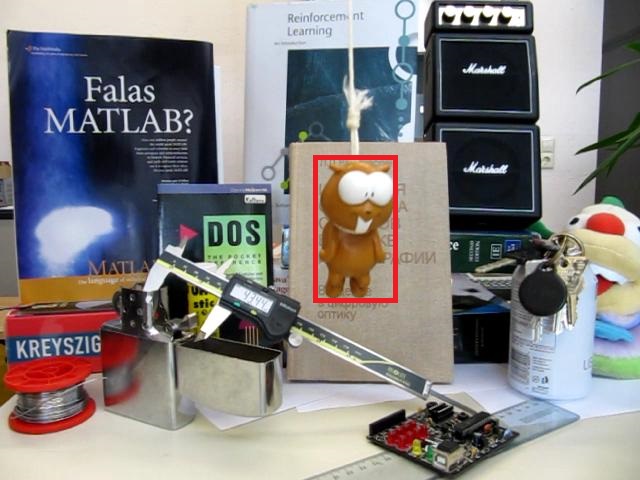}\ &
\includegraphics[width=0.245\linewidth, height=0.18\linewidth]{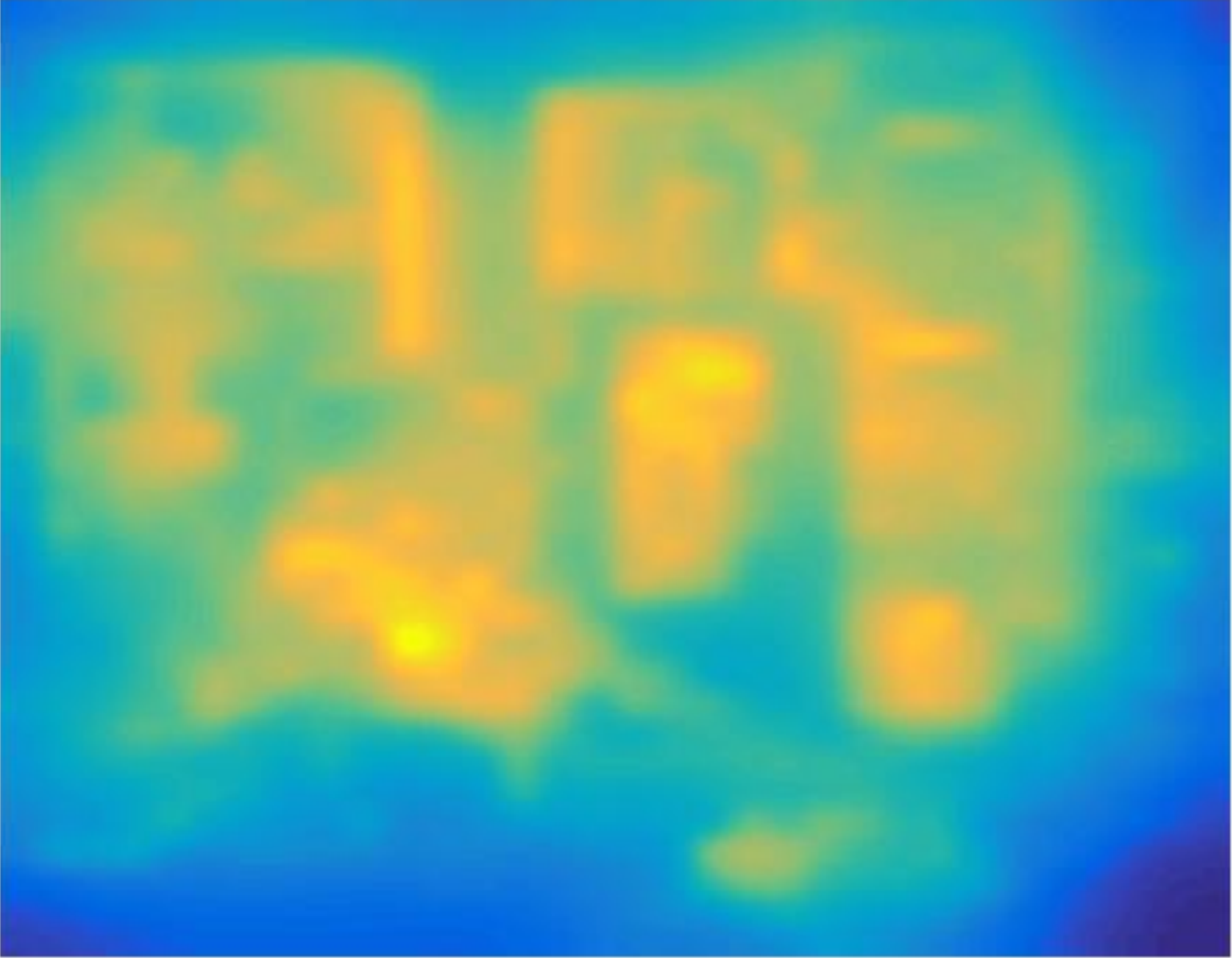}\ &
\includegraphics[width=0.245\linewidth, height=0.18\linewidth]{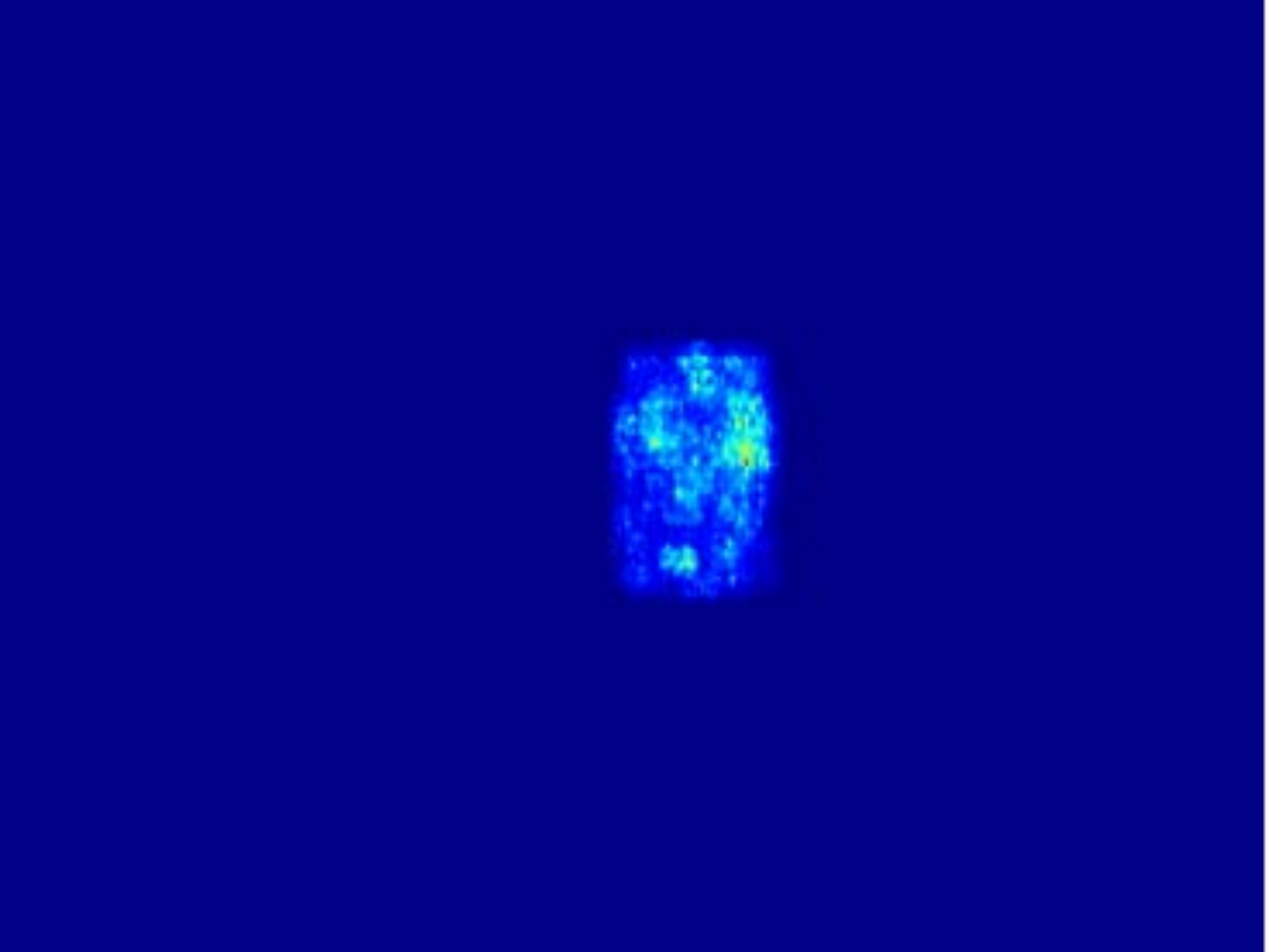}\ &
\includegraphics[width=0.245\linewidth, height=0.18\linewidth]{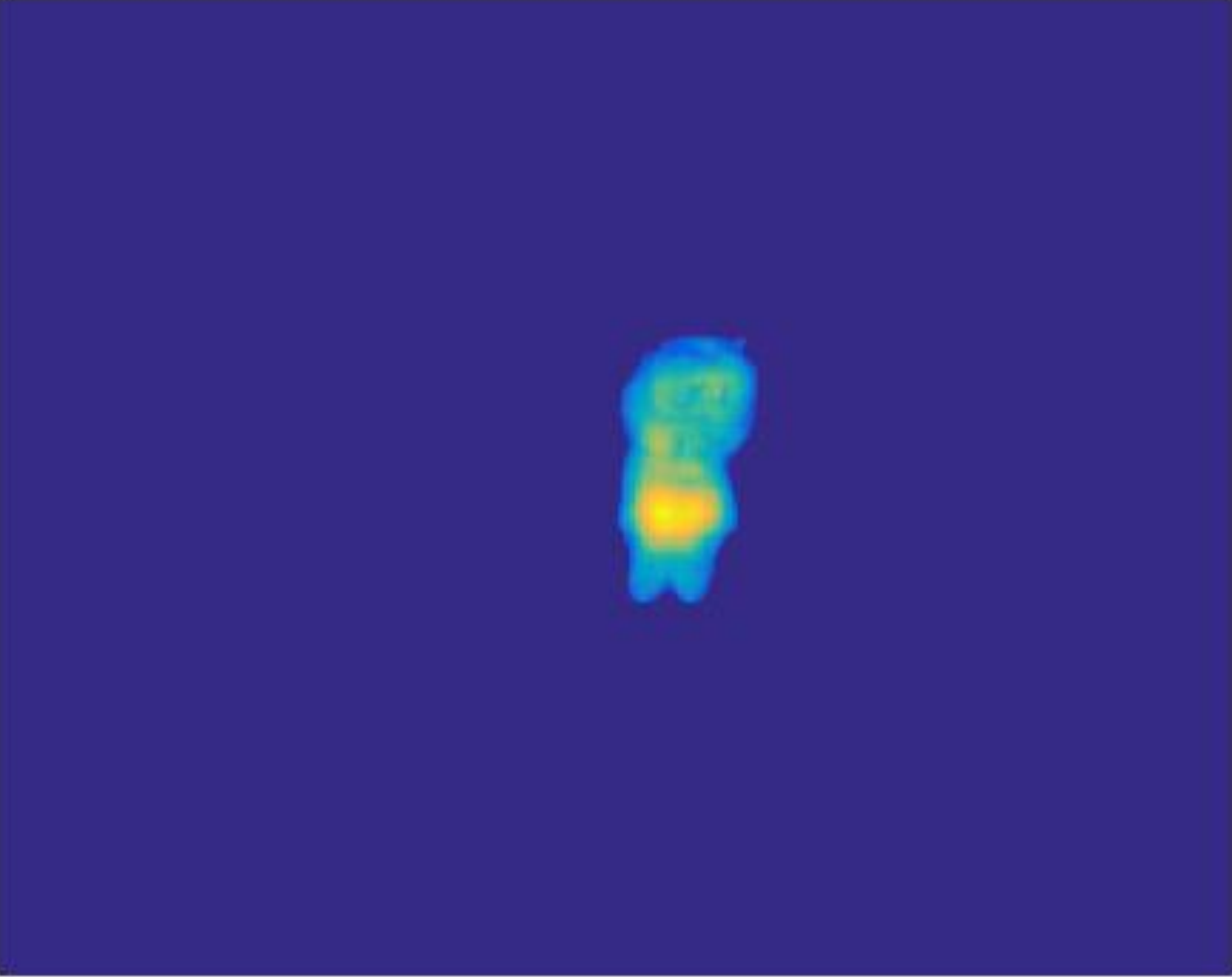}\\
\vspace{-0.5mm}
\includegraphics[width=0.245\linewidth, height=0.18\linewidth]{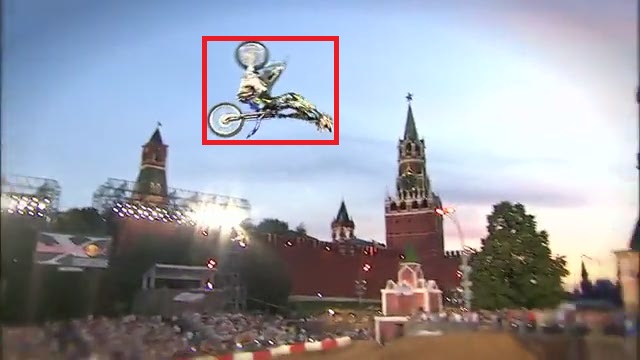}\ &
\includegraphics[width=0.245\linewidth, height=0.18\linewidth]{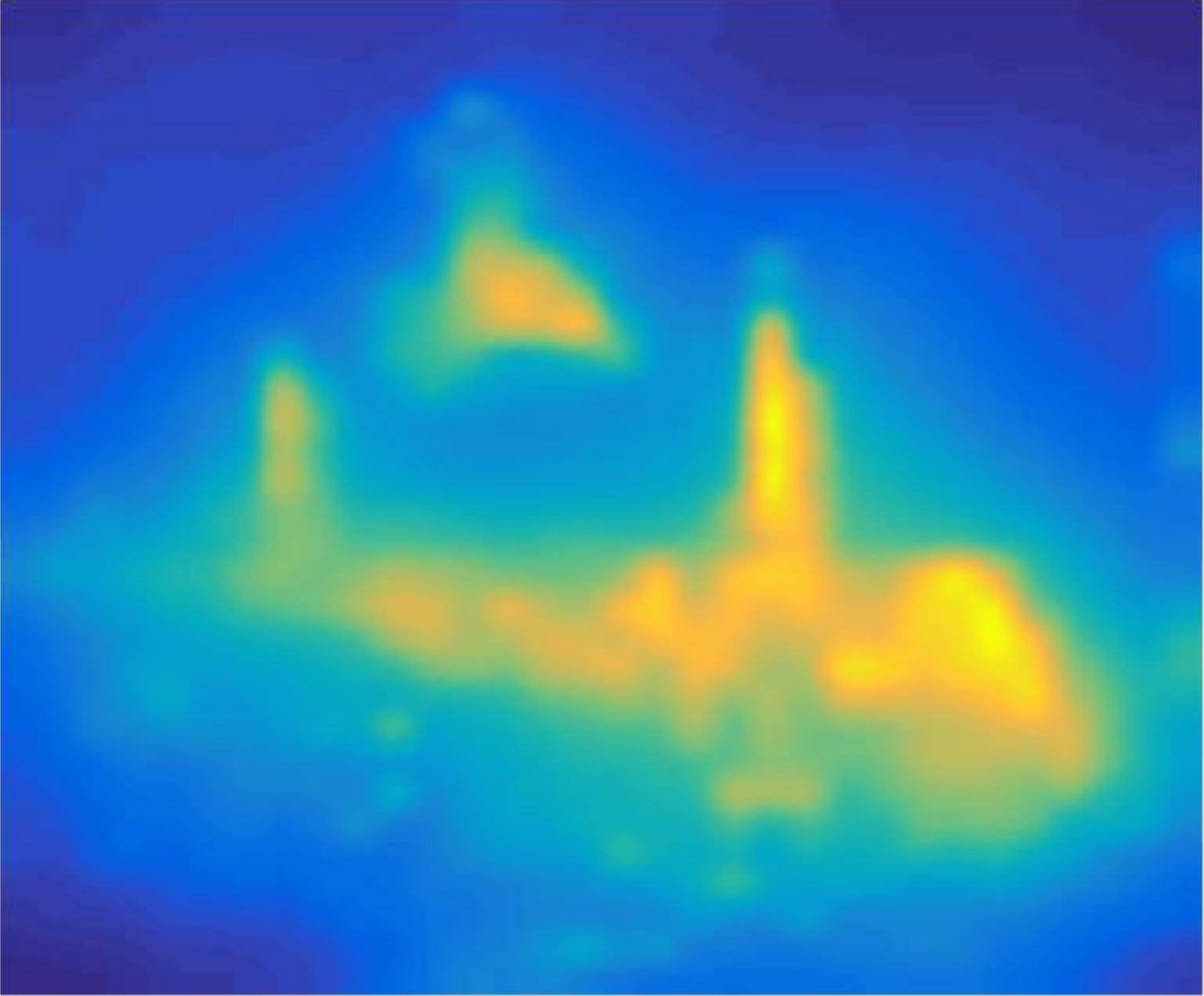}\ &
\includegraphics[width=0.245\linewidth, height=0.18\linewidth]{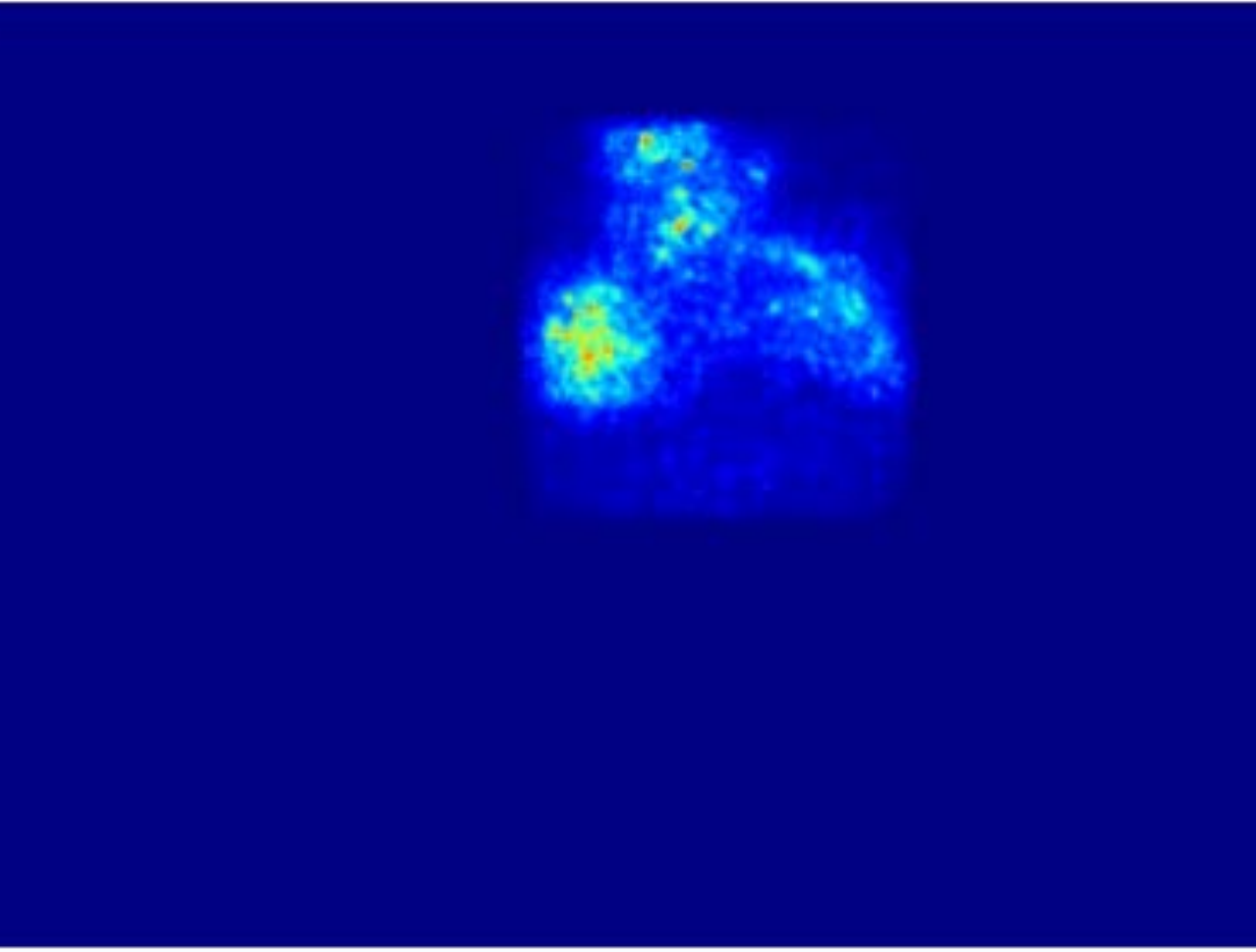}\ &
\includegraphics[width=0.245\linewidth, height=0.18\linewidth]{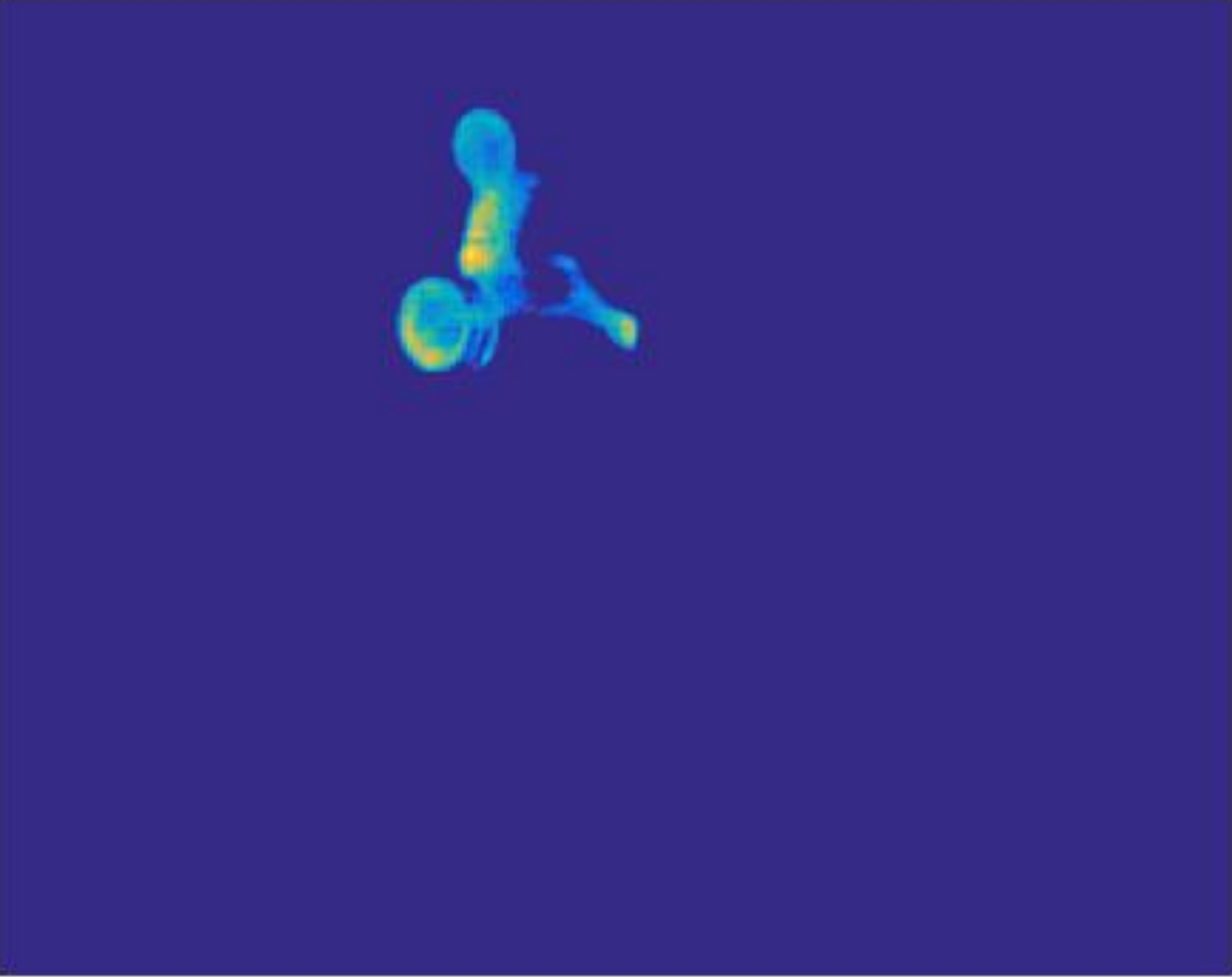}\\
\vspace{-0.5mm}
(a) & (b) &(c) &(d)\\
\end{tabular}
\end{center}
\vspace{-2mm}
\caption{Saliency maps of different methods. From left to right: (a) input frames; (b) video saliency
detection~\cite{videosal}; (c) target-specific saliency map~\cite{dsm};
and (d) our method.}
\label{fig:comparison}
\vspace{-6mm}
\end{figure}

First, our method significantly differs from the video saliency detection algorithms in three aspects: (1) Our aim is to track a single object of interest, whereas that of video saliency detection is to capture all salient objects in a scene; (2) The proposed method focuses on simultaneous local saliency detection and visual tracking, whereas video saliency detection always emphasize obtaining motion saliency maps; (3) The video sequences for visual tracking are usually more complicated, which makes the video
saliency method incapable of capturing the tracked objects. The representative results of video saliency detection are shown in Fig.~\ref{fig:comparison}(b).

Second, our method has several obvious advantages compared with the recent deep saliency-based tracker~\cite{dsm}.
In~\cite{dsm}, saliency maps are derived from features of the fixed fully connected layers, which are often very noisy (Fig.~\ref{fig:comparison}(c)).
However, the proposed saliency model is directly learned from raw images in an end-to-end mechanism.
In addition, the introduction of scale estimation and temporal consistency makes the obtained saliency maps dense
and edge-preserving, which further facilitates non-rigid object tracking.
Fig.~\ref{fig:comparison} illustrates that our method achieves better visual effects and location accuracy compared with other methods.
\section{Experiments}
In this section, we show the experimental results of our method.
First, we describe the datasets for saliency detection and visual tracking.
Then, we give the implementation details of our method. Third, we test and compare our proposed method with other state-of-the-art methods. Both quantitative and qualitative analysis are presented to show the effectiveness of our method.
\subsection{Datasets}
We construct a new large-scale saliency detection dataset by combining the THUS~\cite{thus} and object extraction (OE)~\cite{oe} datasets to pre-train the TFCN model.
The simple mirror reflection and rotation techniques ($0^{\circ}, 90^{\circ}, 180^{\circ}, 270^{\circ}$) are used for data augmentation, resulting in a total of 161,464 training images.
We adopt the recent ECSSD~\cite{hs}, PASCAL-S~\cite{pascal-s} and DUT-OMRON~\cite{dut-omron} datasets to evaluate our method and other deep learning-based algorithms for the saliency detection task.
Other datasets like JuddDB~\cite{sa} and SED~\cite{sed} are frequently used to evaluate saliency detection methods; however, we focus on the single salient object detection in this paper, which is considerably related to the online tracking problem.
The recent tracking dataset presented in~\cite{ogbdt} is adopted to highlight the tracking process of non-rigid and articulated objects
for the tracking task.
In addition, the popular OTB-50~\cite{otb} dataset is used to evaluate the generalization ability of our tracker.
\subsection{Implementation Details}
We implement our approaches based on the MATLAB R2014b platform with the Caffe toolbox~\cite{caffe}.
The maximum edge of ROIs are restricted to be 256, considering the tracked targets may be relatively small or extremely
large.
Other critical parameters are set as follows: the number of rectangular regions $N = 6$, the accumulated time interval $\tau = 2$, and the decay factor $c = 1.1$.
For the TFCN pre-training, we offline train the TFCN using the stochastic gradient descent (SGD) with a momentum 0.9, weight decay 0.0005, and mini-batch 8.
We set the base learning rate to 1e-8 and decrease the learning rate by 0.1 when training loss reaches a plateau.
The training process of our TFCN converges after 100k iterations.
During online visual tracking, we set the max iteration to 100, batch size 1, and learning rate 1e-12, and keep other parameters fixed as in saliency detection.
All above parameters are fixed during our experiments.
We run our approach in a quad-core PC machine with an i7-4790 CPU (with 16G memory) and a NVIDIA Titan X GPU (with 12G memory).
The pre-training process of our TFCNN model takes almost 6 hours.
The proposed simultaneous saliency detection and tracking algorithm runs at approximately 7 fps.
\subsection{Experimental Results on Saliency Detection}
We compare the proposed saliency detection algorithm with three state-of-the-art ones including two deep CNN-based algorithms (LEGS~\cite{legs} and MDF~\cite{mdf}) and the DRFI~\cite{drfi} method\footnote{In traditional methods, DRFI performs best in most of salient object detection benchmarks~\cite{borji2015salient}.}, which bases on the integration of multiple hand-craft features.
For fair comparison, we utilize either the implementations with recommended parameter settings or the saliency maps provided by the authors.
To evaluate all compared methods, we use two most common metrics, i.e., F-measure and precision-recall (PR) curve.
The F-measure value is defined as,
\begin{equation}
  F_{\beta} =\frac{(1+\beta^2)\times Precision\times Recall}{\beta^2\times Precision \times Recall},
  \label{equ:equ8}
\end{equation}
where $\beta^2$ is set as 0.3 to weigh precision more than recall as in [46] [41] [5] [47].
We report the performance when each saliency map is binarized with an image-dependent threshold.
This adaptive threshold is determined to be twice the mean saliency of the image:
\begin{equation}
T_{a} = \frac{2}{W\times H}\sum_{x=1}^{W}\sum_{y=1}^{H}S(x,y),
  \label{equ:equ9}
\end{equation}
where $W$ and $H$ are width and height of an image, $S(x,y)$ is the saliency value of the pixel at
$(x,y)$. We report the average precision, recall, F-measure and AUC over each dataset.
We note that a higher F-measure value means that the corresponding algorithm can capture more valid salient regions.
Thus, achieving a high F-measure is very helpful for online tracking.

\begin{figure}
\centering
\begin{tabular}{@{}c@{}c}
\includegraphics[width=0.5\linewidth, height=0.4\linewidth]{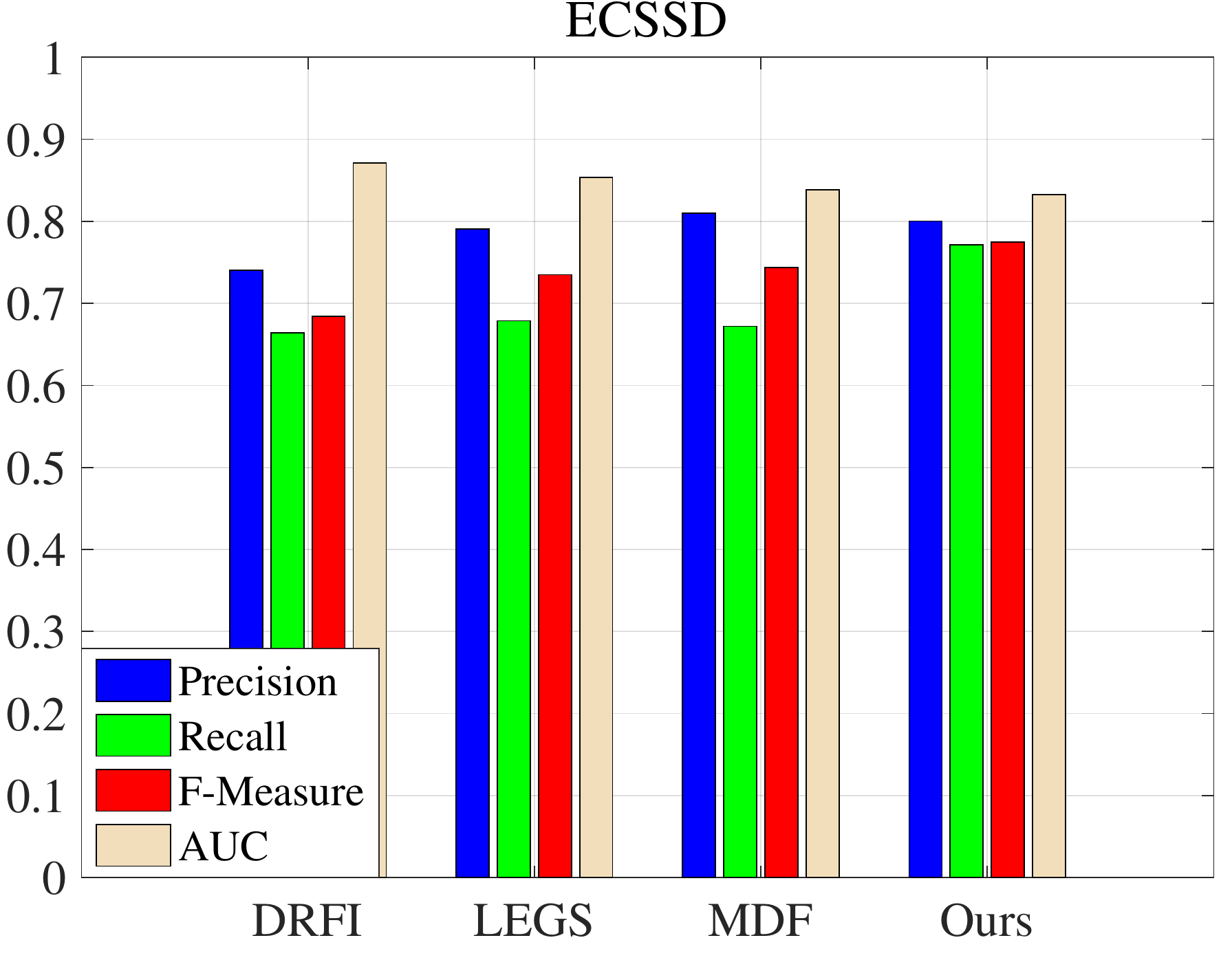}\ &
\includegraphics[width=0.5\linewidth, height=0.4\linewidth]{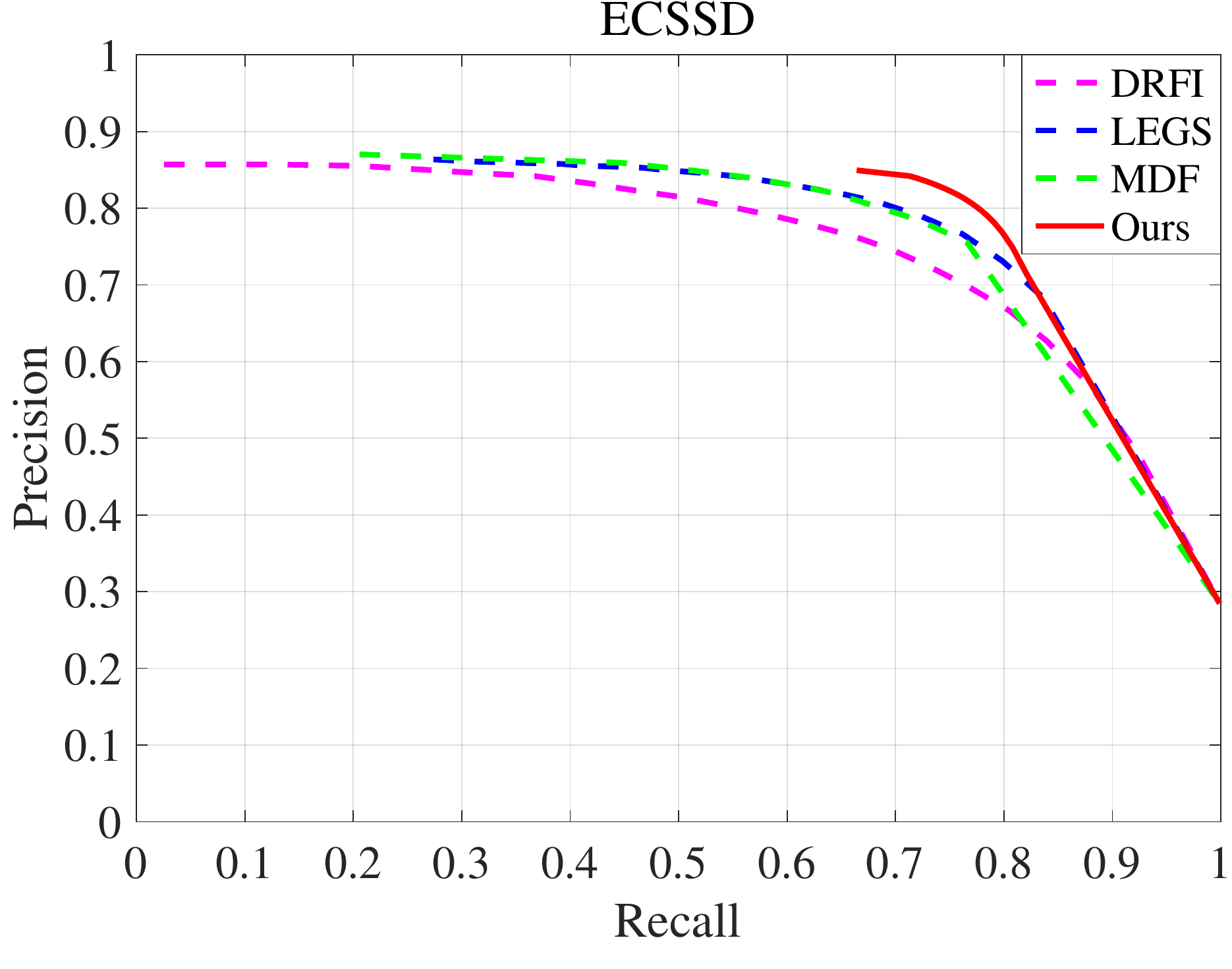}\ \\
\includegraphics[width=0.5\linewidth, height=0.4\linewidth]{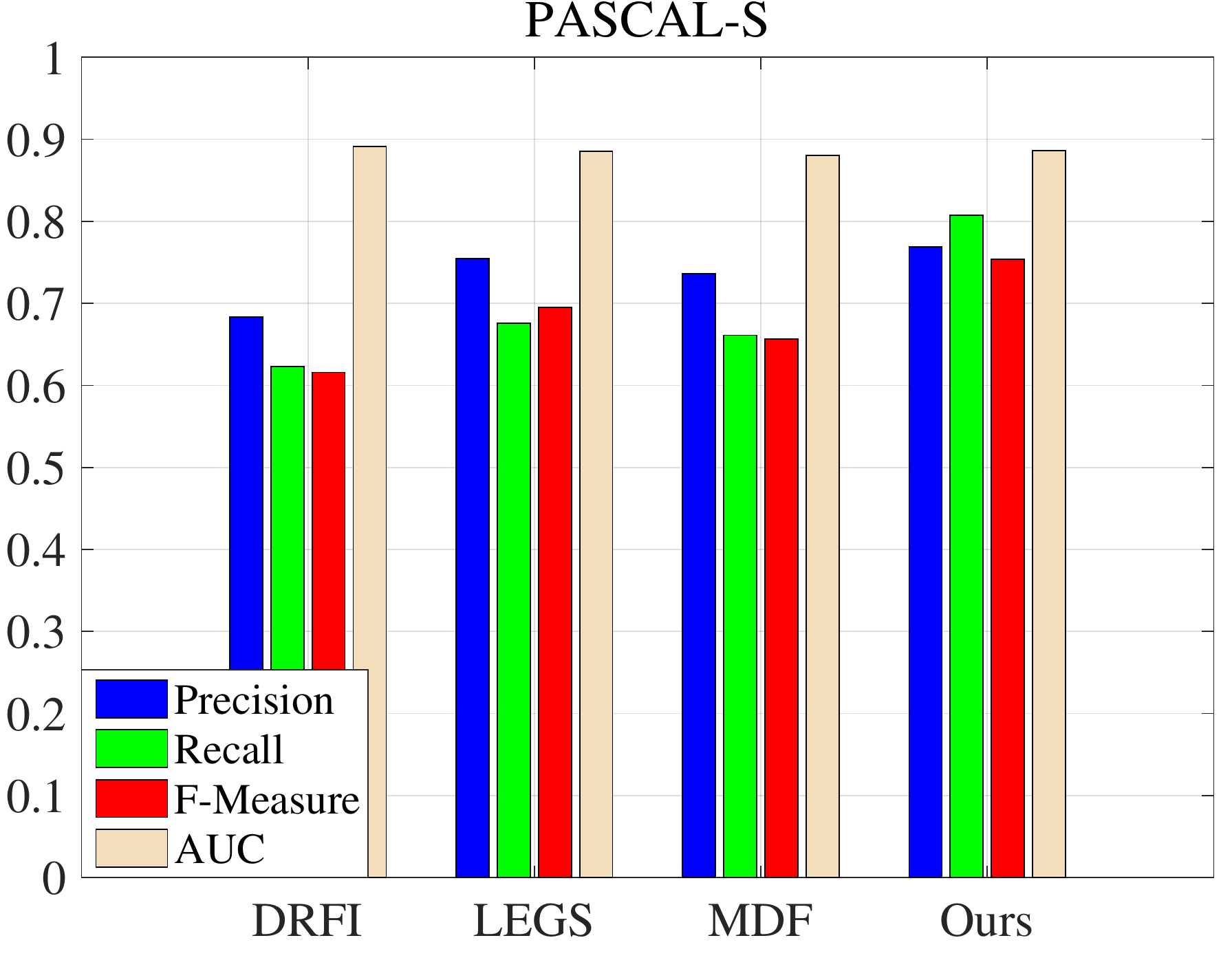}\ &
\includegraphics[width=0.5\linewidth, height=0.4\linewidth]{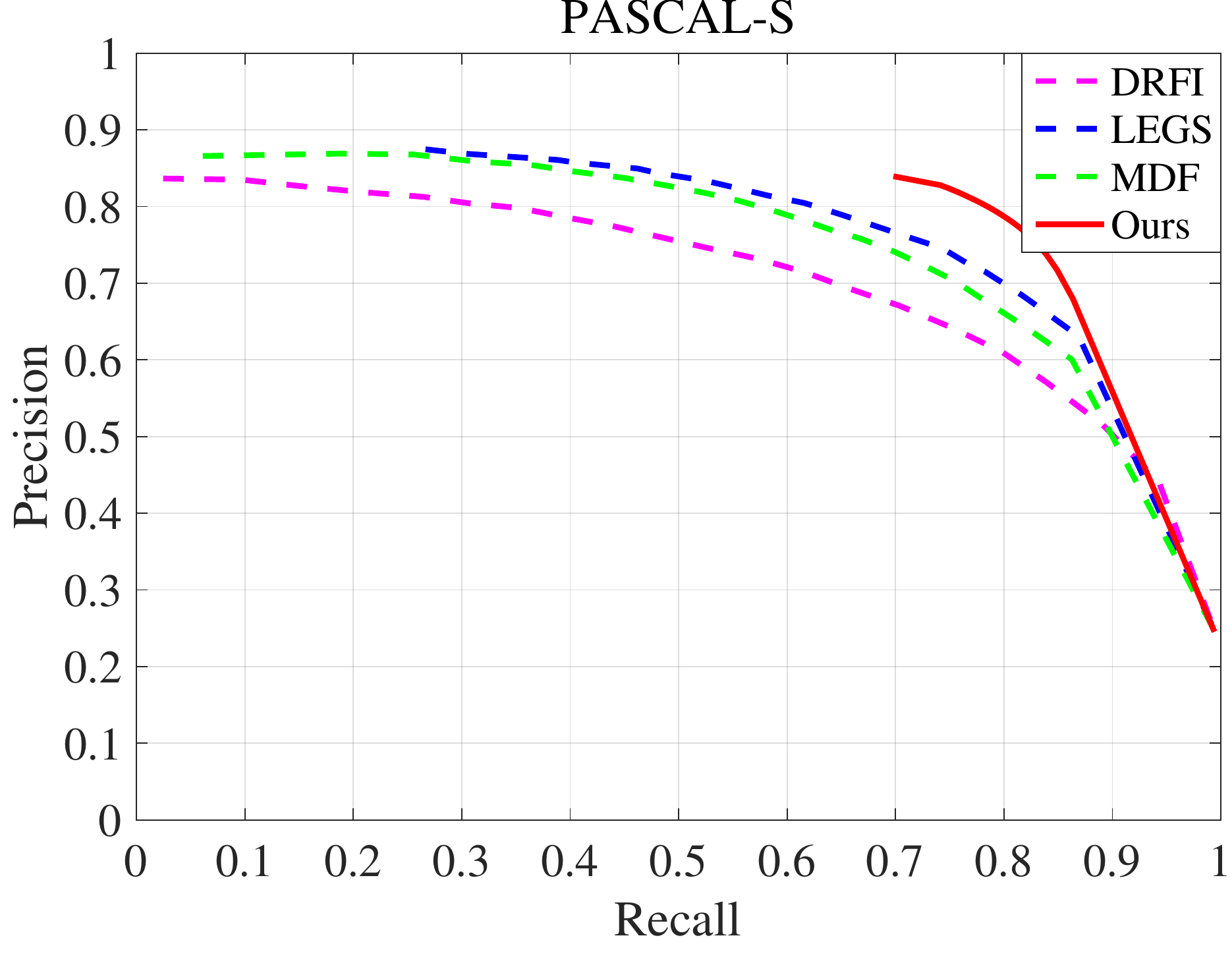}\ \\
\includegraphics[width=0.5\linewidth, height=0.4\linewidth]{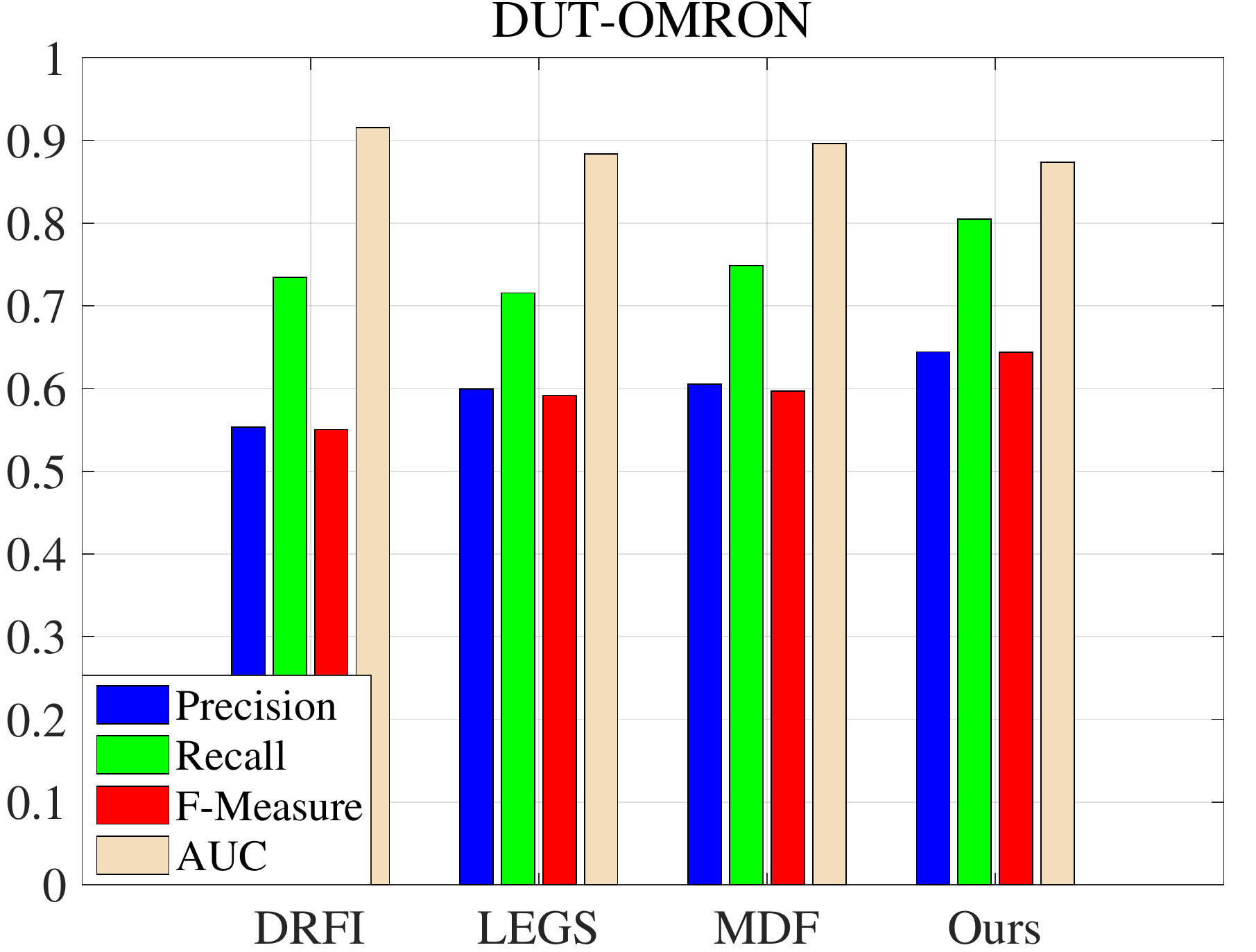}\ &
\includegraphics[width=0.5\linewidth, height=0.4\linewidth]{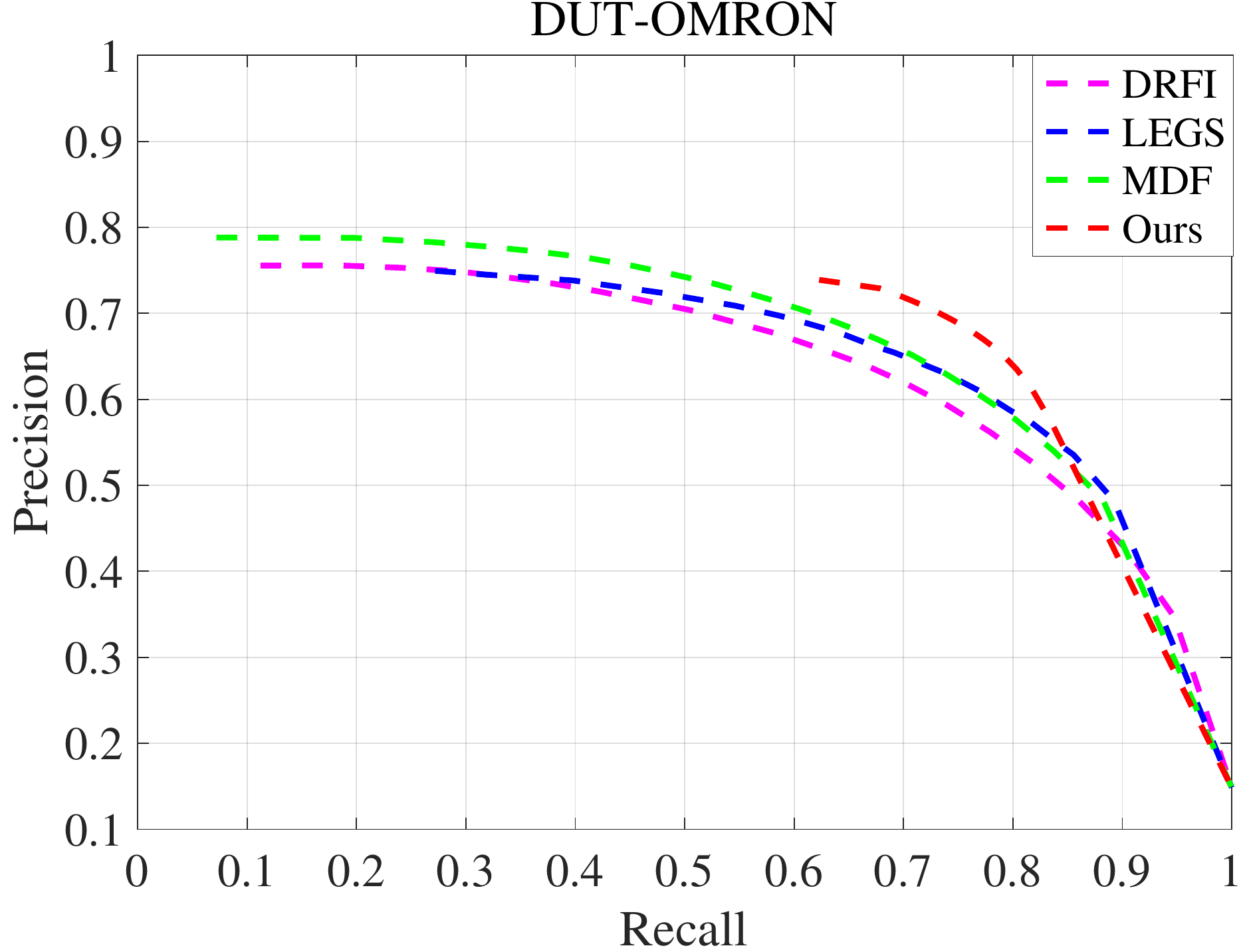}\ \\
{\footnotesize (a)} & {\footnotesize (b)}\ \\
\end{tabular}
\vspace{-2mm}
\caption{Quantitative results of all compared methods in terms of (a) Bar graph
and (b) P-R curve metrics. Our method consistently outperforms other compared methods.
\label{fig:PRF}}
\end{figure}

\begin{figure}
\centering
\footnotesize
\begin{tabular}{@{}c@{}c@{}c@{}c@{}c@{}c}
\vspace{-0.5mm}
\includegraphics[width=0.16\linewidth]{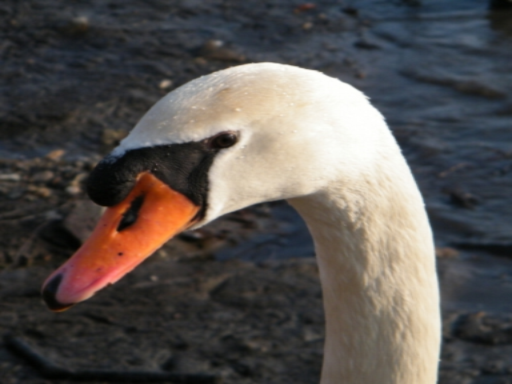} 				\ &
\includegraphics[width=0.16\linewidth]{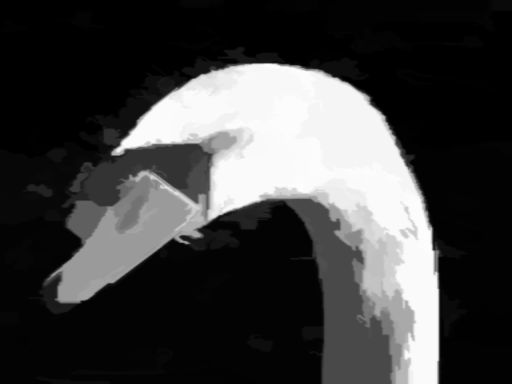} 	\ &
\includegraphics[width=0.16\linewidth]{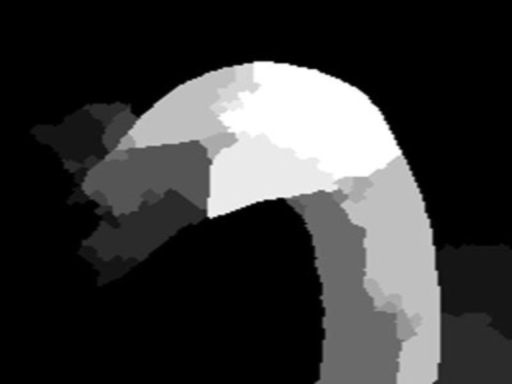} 	\ &
\includegraphics[width=0.16\linewidth]{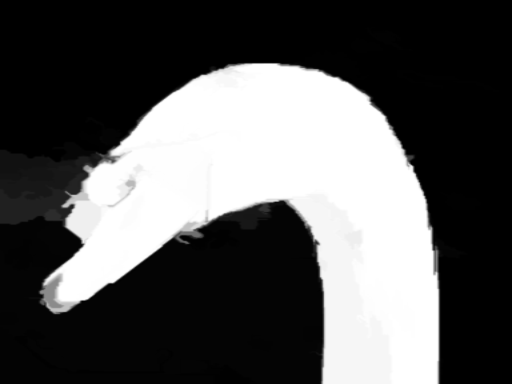} 	\ &
\includegraphics[width=0.16\linewidth]{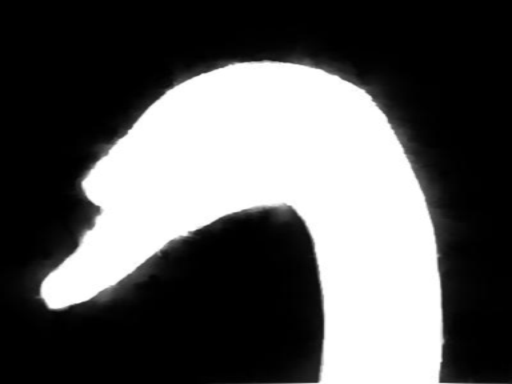} 		\ &
\includegraphics[width=0.16\linewidth]{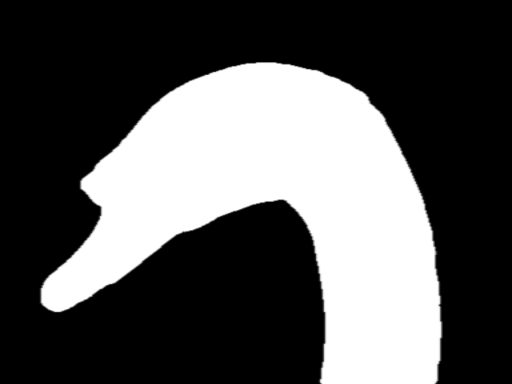} 		\\
\vspace{-0.5mm}
\includegraphics[width=0.16\linewidth]{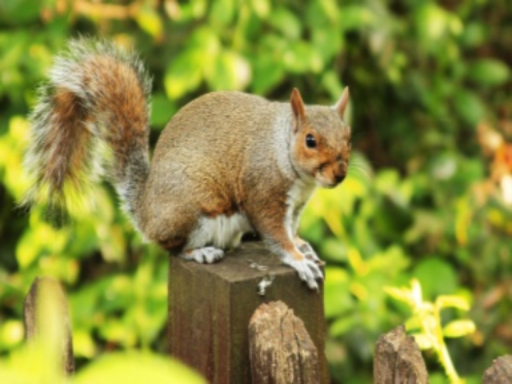} 				\ &
\includegraphics[width=0.16\linewidth]{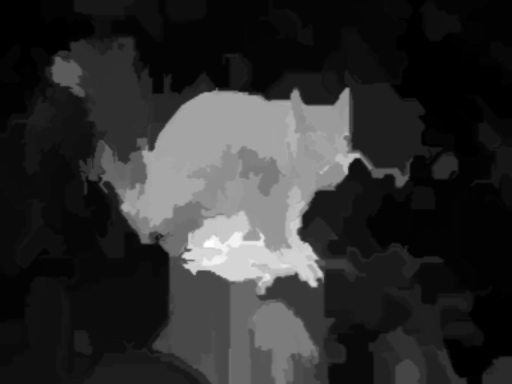} 	\ &
\includegraphics[width=0.16\linewidth]{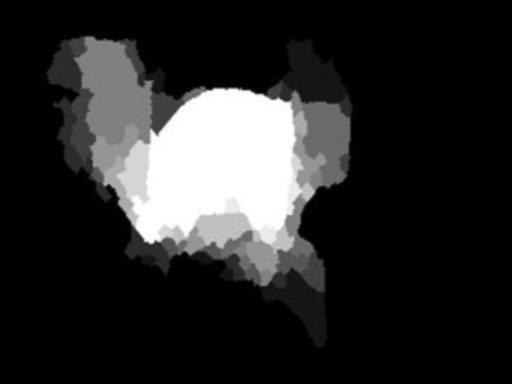} 	\ &
\includegraphics[width=0.16\linewidth]{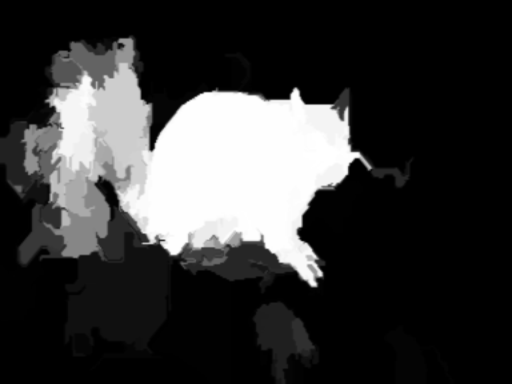} 	\ &
\includegraphics[width=0.16\linewidth]{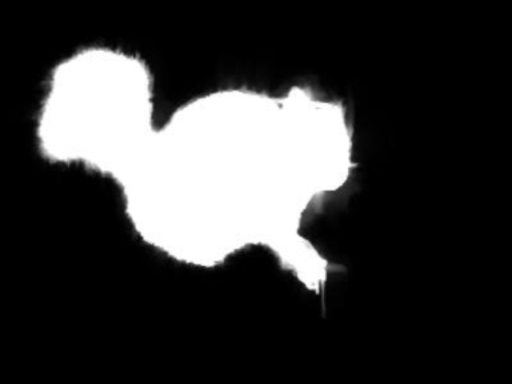} 		\ &
\includegraphics[width=0.16\linewidth]{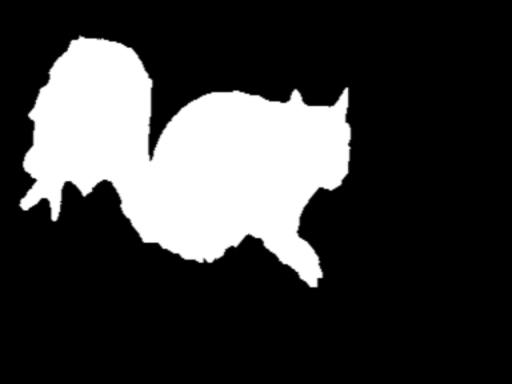} 		\\
\vspace{-0.5mm}
\includegraphics[width=0.16\linewidth]{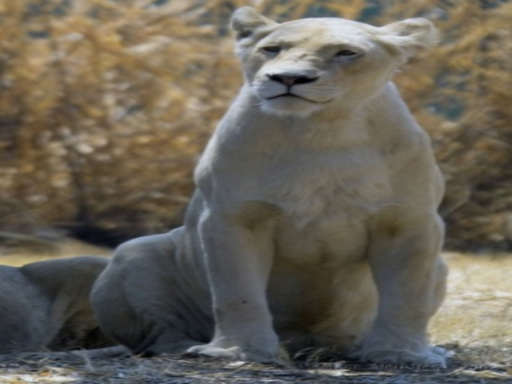} 				\ &
\includegraphics[width=0.16\linewidth]{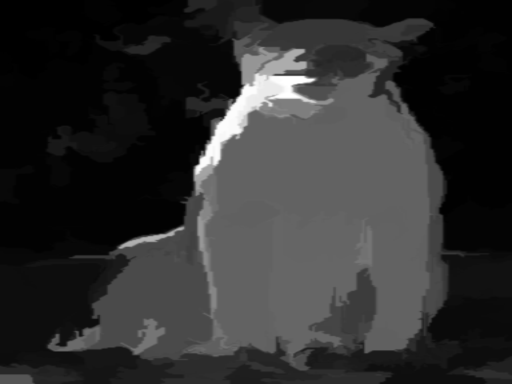} 	\ &
\includegraphics[width=0.16\linewidth]{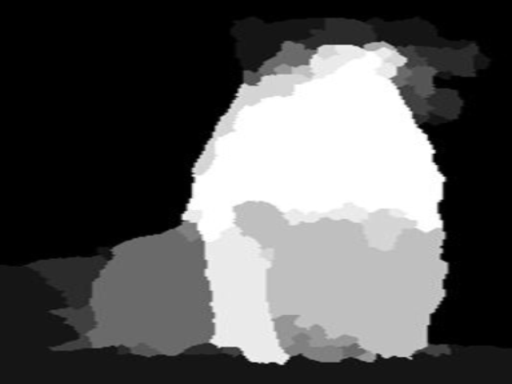} 	\ &
\includegraphics[width=0.16\linewidth]{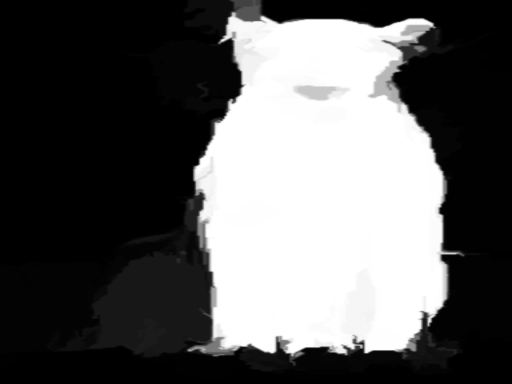} 	\ &
\includegraphics[width=0.16\linewidth]{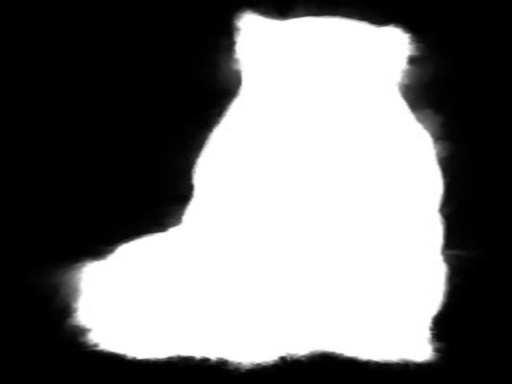} 		\ &
\includegraphics[width=0.16\linewidth]{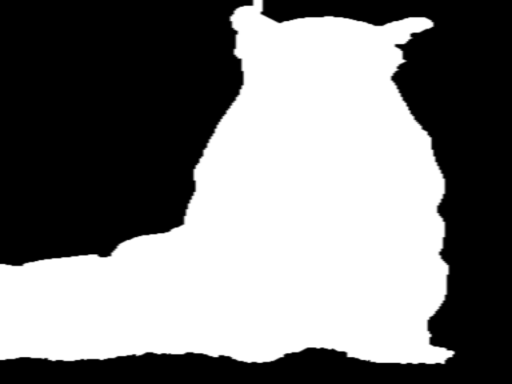} 		\\
\vspace{-0.5mm}
\includegraphics[width=0.16\linewidth]{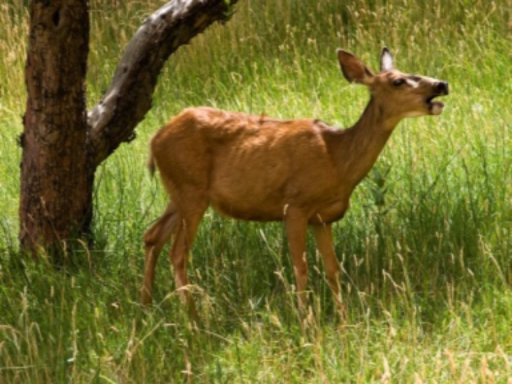} 				\ &
\includegraphics[width=0.16\linewidth]{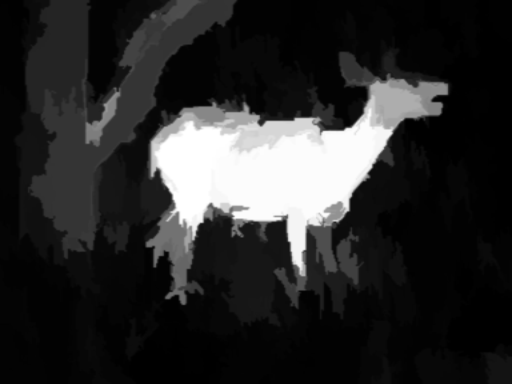} 	\ &
\includegraphics[width=0.16\linewidth]{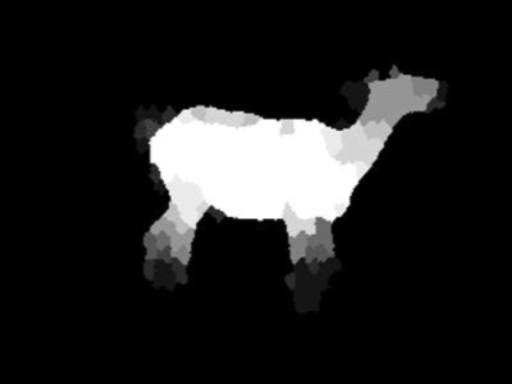} 	\ &
\includegraphics[width=0.16\linewidth]{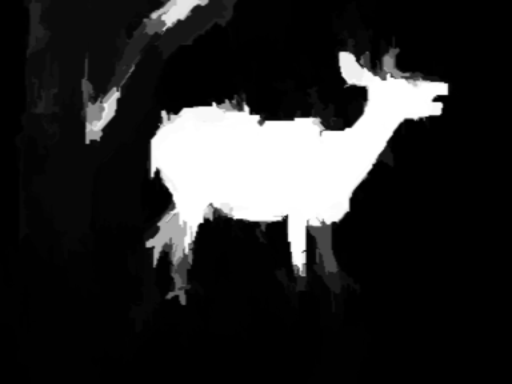} 	\ &
\includegraphics[width=0.16\linewidth]{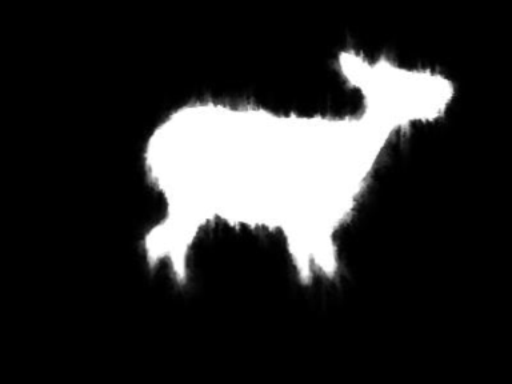} 		\ &
\includegraphics[width=0.16\linewidth]{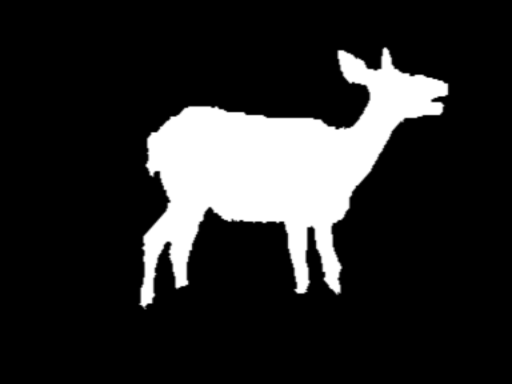} 		\\
\vspace{-0.5mm}
(a) & (b) & (c) &(d) &(e) &(f)\\
\end{tabular}
\vspace{-2mm}
\caption{Representative results of the compared methods for saliency detection.
From left to right: (a) input images; (b) DRFI~\cite{drfi}; (c) LEGS~\cite{legs}; (d) MDF~\cite{mdf}; (e) our method; and (f) ground truth. Our method consistently produces saliency maps closest to the ground truth.
\label{fig:saliencyexamples}}
\vspace{-4mm}
\end{figure}

Fig.~\ref{fig:PRF}(a) illustrates the precision, recall and F-measure values for all three datasets, from which we can see that the proposed method achieves better performance than other competing ones.
In addition, Fig.~\ref{fig:PRF}(b) demonstrates the P-R curves of different algorithms, in which the average P-R values are calculated using a series of binary maps that take different thresholds on the original saliency map.
As can be seen from the figures, our method performs better than other state-of-the-art ones, even when the recall threshold is high.
Several representative saliency maps are shown in Fig.~\ref{fig:saliencyexamples}, which clearly shows that the results obtained by our method are significantly closer to the ground truth.
The predicted saliency maps can convincingly identify the salient objects and provide the accurate saliency regions.
\setlength{\tabcolsep}{1.2pt}
\begin{table*}
\small
\begin{center}
\caption{Quantitative results on the non-rigid object tracking dataset~\cite{ogbdt}.
All numbers are presented in terms of percentage (\%). For each row, the best and second-best
results are shown in red and blue fonts respectively. }
\label{table:bboverlap}
\doublerulesep=1pt
\begin{tabular}{|c|c|c|c|c|c|c|c|c|c|c|c|c|c|c|c|c|c|c|c|c|c|c|c|}
\hline
\multicolumn{4}{|c}{}&\multicolumn{12}{|c}{(a) Bounding box overlap ratio}&\multicolumn{8}{|c|}{(b) Segmentation overlap ratio}\\
\hline
\multicolumn{4}{|c|}{}
&HT&SPT&PT&OGBT&Struck&SCM&MEEM&MUSTer&DSMT&FCNT&HCFT&Ours&HT&SPT&PT&OGBT&Ours$_{1}$&Ours$_{2}$&Ours$_{3}$&Ours\\
\hline
\multicolumn{4}{|c|}{\scriptsize{Cliff-dive1}}
&61.0&66.5&29.9&\textcolor[rgb]{0,0,1}{75.9}&62.4&61.8&33.3&60.9&67.3&62.3&68.1&\textcolor[rgb]{1,0,0}{77.2}&64.2&54.6&60.1& 67.6&64.5&\textcolor[rgb]{0,0,1}{69.3}&56.4&\textcolor[rgb]{1,0,0}{71.4}\\
\hline
\multicolumn{4}{|c|}{\scriptsize{Cliff-dive2}}
&\textcolor[rgb]{0,0,1}{52.0}&30.3&13.4&49.3&34.0&30.0&27.6&14.8&37.6&36.2&40.7&\textcolor[rgb]{1,0,0}{54.7}&\textcolor[rgb]{0,0,1}{49.4}&41.8&16.0& 36.7&43.1&45.6&21.&\textcolor[rgb]{1,0,0}{50.8}\\
\hline
\multicolumn{4}{|c|}{\scriptsize{Diving}}
&7.9&35.2&12.3&\textcolor[rgb]{1,0,0}{50.6}&33.6&15.1&12.6&20.7&41.3&24.3&34.5&\textcolor[rgb]{0,0,1}{44.8}&6.7&21.2&25.5&
44.1&33.2&37.0&28.5&\textcolor[rgb]{1,0,0}{42.4}\\
\hline
\multicolumn{4}{|c|}{\scriptsize{Gymnastics}}
&10.4&42.6&26.0&70.4&53.7&13.9&17.6&47.7&\textcolor[rgb]{0,0,1}{72.1}&56.6&71.2&\textcolor[rgb]{1,0,0}{74.6}&9.2&10.6&52.0& \textcolor[rgb]{0,0,1}{69.8}&53.2&67.2&38.5&\textcolor[rgb]{1,0,0}{72.4}\\
\hline
\multicolumn{4}{|c|}{\scriptsize{High-jump}}
&39.1&5.3&0.6&\textcolor[rgb]{1,0,0}{51.5}&15.4&8.4&21.1&8.3&43.6&44.2&48.9&\textcolor[rgb]{0,0,1}{49.4}&40.4&\textcolor[rgb]{1,0,0}{52.1}&0.9&42.8&
32.5&42.6&24.5&\textcolor[rgb]{0,0,1}{46.6}\\
\hline
\multicolumn{4}{|c|}{\scriptsize{Motocross1}}
&55.1&11.2&3.6&62.4&29.2&10.8&9.1&22.1&\textcolor[rgb]{0,0,1}{68.8}&67.6&68.5&\textcolor[rgb]{1,0,0}{70.5}&52.2&8.9&1.4&53.1&
54.2&\textcolor[rgb]{0,0,1}{57.4}&41.3&\textcolor[rgb]{1,0,0}{59.2}\\
\hline
\multicolumn{4}{|c|}{\scriptsize{Motocross2}}
&62.0&46.1&21.3&\textcolor[rgb]{0,0,1}{72.1}&70.6&41.7&60.4&68.1&71.3&67.9&70.7&\textcolor[rgb]{1,0,0}{74.9}&53.0&37.1&39.7& 64.5&56.8&\textcolor[rgb]{0,0,1}{66.5}&48.2&\textcolor[rgb]{1,0,0}{69.7}\\
\hline
\multicolumn{4}{|c|}{\scriptsize{Mtn-bike}}
&48.6&53.0&12.9&61.2&66.2&71.4&64.8&69.5&69.4&68.5&\textcolor[rgb]{1,0,0}{72.1}&\textcolor[rgb]{0,0,1}{71.0}&53.4&43.0&32.1& 54.9&48.2&\textcolor[rgb]{0,0,1}{55.3}&36.5&\textcolor[rgb]{1,0,0}{57.4}\\
\hline
\multicolumn{4}{|c|}{\scriptsize{Skiing}}
&50.2&27.7&21.3&39.0&3.2&7.4&29.5&3.5&32.6&\textcolor[rgb]{1,0,0}{54.1}&52.6&\textcolor[rgb]{0,0,1}{53.2}&41.0&37.3&43.0& 32.1&38.6&\textcolor[rgb]{0,0,1}{45.7}&31.4&\textcolor[rgb]{1,0,0}{47.5}\\
\hline
\multicolumn{4}{|c|}{\scriptsize{Transformer}}
&63.4&55.8&13.9&\textcolor[rgb]{0,0,1}{86.6}&57.7&55.6&51.4&59.6&82.4&72.8&84.1&\textcolor[rgb]{1,0,0}{89.4}&45.0&2.8&5.5& 74.0&54.3&\textcolor[rgb]{0,0,1}{74.3}&45.1&\textcolor[rgb]{1,0,0}{76.8}\\
\hline
\multicolumn{4}{|c|}{\scriptsize{Volleyball}}
&27.7&26.8&15.2&46.2&36.2&13.2&31.1&19.0&38.5&45.3&\textcolor[rgb]{1,0,0}{49.8}&\textcolor[rgb]{0,0,1}{49.2}&31.1&6.5&25.1& 41.1&25.1&\textcolor[rgb]{0,0,1}{43.8}&33.4&\textcolor[rgb]{1,0,0}{45.2}\\
\hline
\multicolumn{4}{|c|}{\scriptsize{\textbf{Average}}}
&43.4&36.4&15.5&\textcolor[rgb]{0,0,1}{60.5}&42.0&29.9&32.6&35.8&56.8&54.5&60.2&\textcolor[rgb]{1,0,0}{64.5}&40.5&28.7&27.4
&52.8&45.8&\textcolor[rgb]{0,0,1}{55.0}&36.8&\textcolor[rgb]{1,0,0}{58.1}\\
\hline
\end{tabular}
\end{center}
\vspace{-6mm}
\end{table*}
\setlength{\tabcolsep}{1.2pt}
\subsection{Experimental Results on Visual Tracking}
\subsubsection{Non-rigid Object Tracking}
The most recent dataset of non-rigid object tracking~\cite{ogbdt} includes 11 challenging image sequences with pixel-wise annotations in each frame.
The bounding box annotations are generated by computing the tightest rectangular boxes containing all target pixels.
Based on this dataset, we compare the proposed method with four public segmentation-based algorithms (HT~\cite{hough}, SPT~\cite{Superpixel}, PT~\cite{Pixeltrack} and OGBT~\cite{ogbdt}) and seven state-of-the-art bounding box-based trackers (Struck~\cite{struck}, SCM~\cite{scm}, MEEM~\cite{meem}, MUSTer~\cite{muster}, DSMT~\cite{dsm}, FCNT~\cite{fcnt}, and HCFT~\cite{hcf}). These trackers based on hard-crafted~\cite{struck,scm,meem,muster} or deep~\cite{dsm,fcnt,hcf} features achieve top performance on recent large-scale tracking benchmarks~\cite{otb}.

\begin{figure*}
\centering
\footnotesize
\begin{tabular}{@{}c@{}c@{}c@{}c@{}c@{}c@{}c}
\includegraphics[width=0.14\linewidth, height=0.099\linewidth]{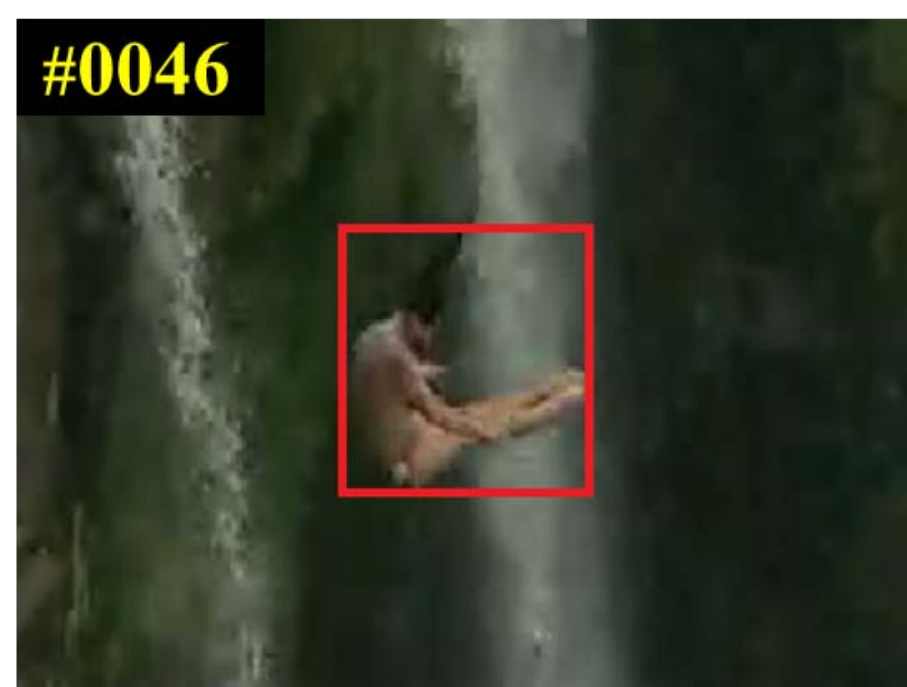} \ &
\includegraphics[width=0.14\linewidth, height=0.099\linewidth]{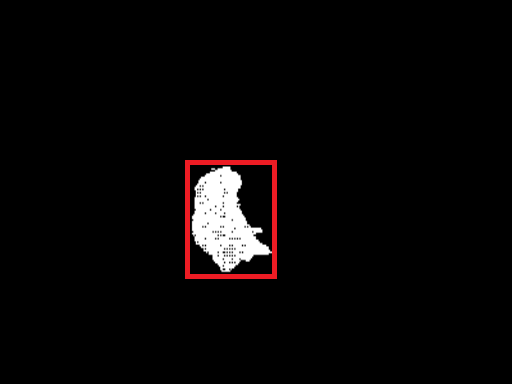} \ &
\includegraphics[width=0.14\linewidth, height=0.099\linewidth]{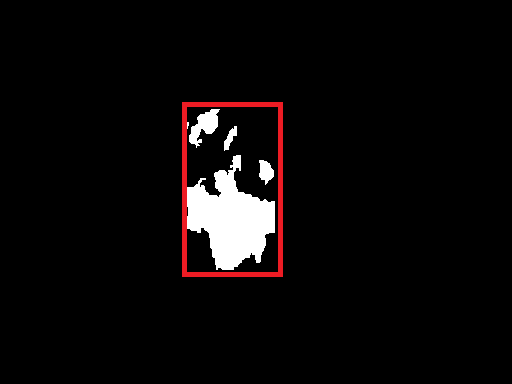} \ &
\includegraphics[width=0.14\linewidth, height=0.099\linewidth]{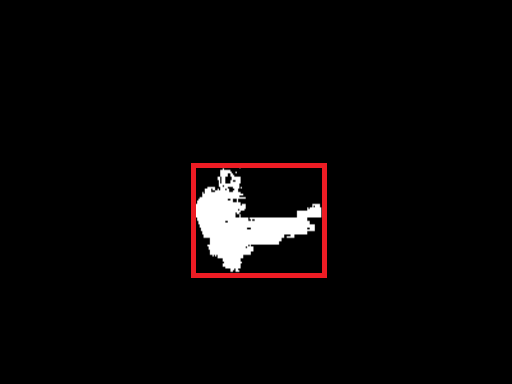} \ &
\includegraphics[width=0.14\linewidth, height=0.099\linewidth]{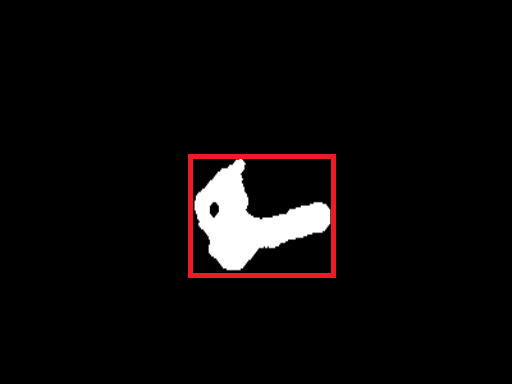} \ &
\includegraphics[width=0.14\linewidth, height=0.099\linewidth]{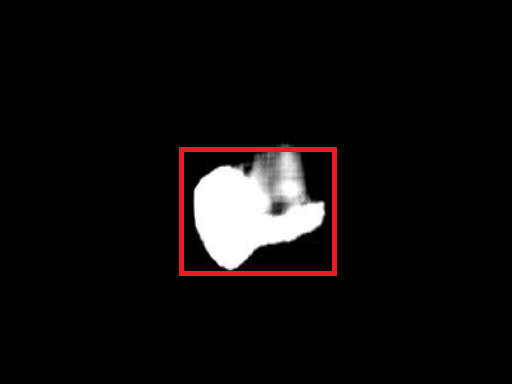} \ &
\includegraphics[width=0.14\linewidth, height=0.099\linewidth]{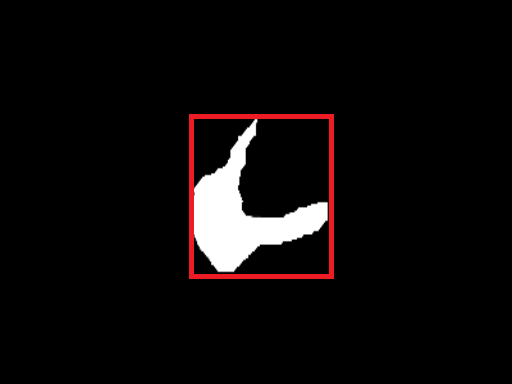} \ \\
\includegraphics[width=0.14\linewidth, height=0.099\linewidth]{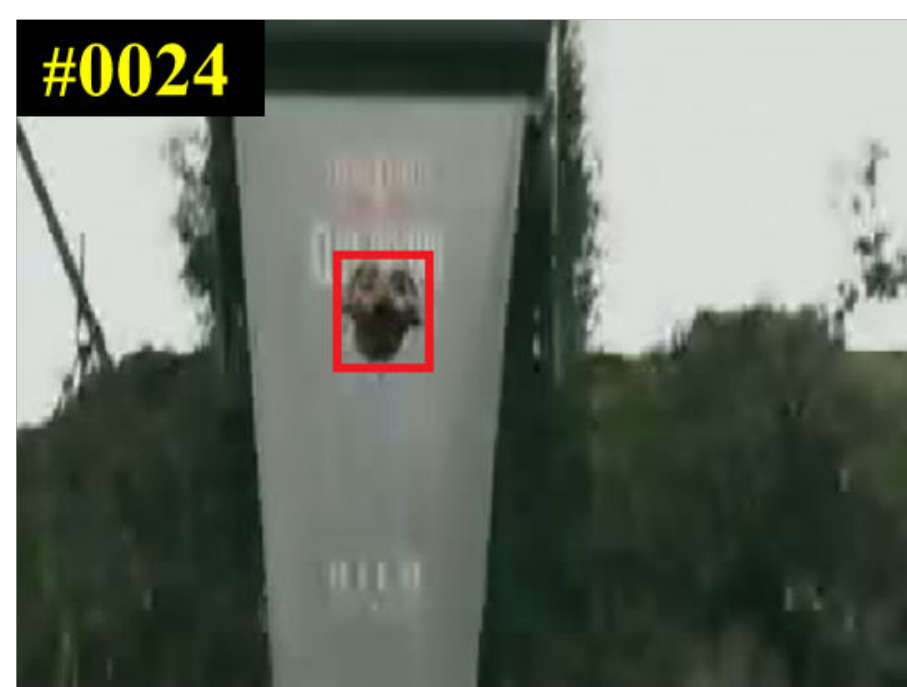} \ &
\includegraphics[width=0.14\linewidth, height=0.099\linewidth]{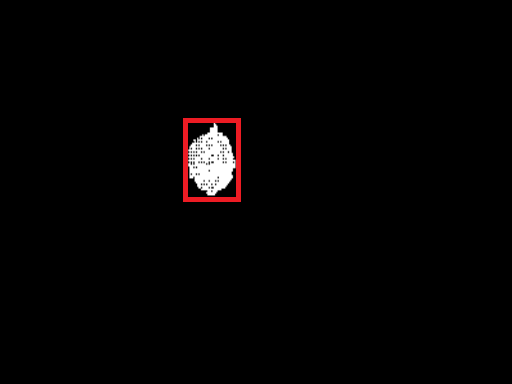} \ &
\includegraphics[width=0.14\linewidth, height=0.099\linewidth]{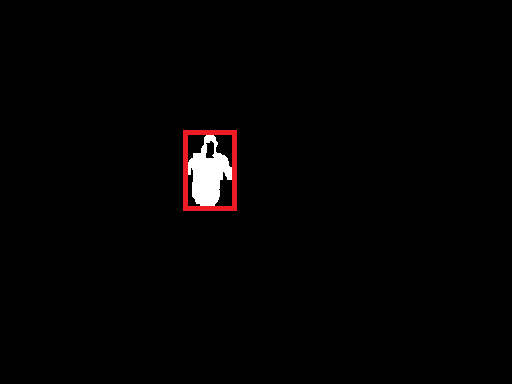} \ &
\includegraphics[width=0.14\linewidth, height=0.099\linewidth]{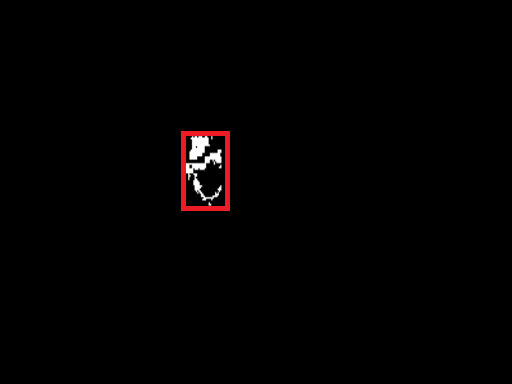} \ &
\includegraphics[width=0.14\linewidth, height=0.099\linewidth]{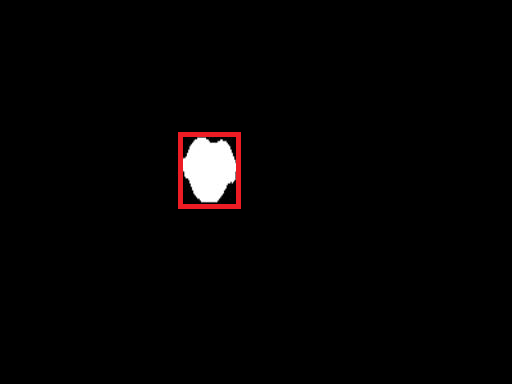} \ &
\includegraphics[width=0.14\linewidth, height=0.099\linewidth]{ogbdt0024.png} \ &
\includegraphics[width=0.14\linewidth, height=0.099\linewidth]{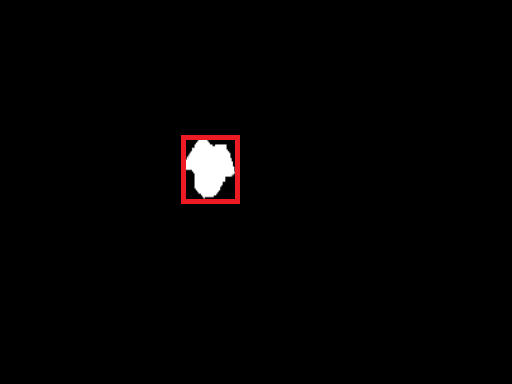} \ \\
\includegraphics[width=0.14\linewidth, height=0.099\linewidth]{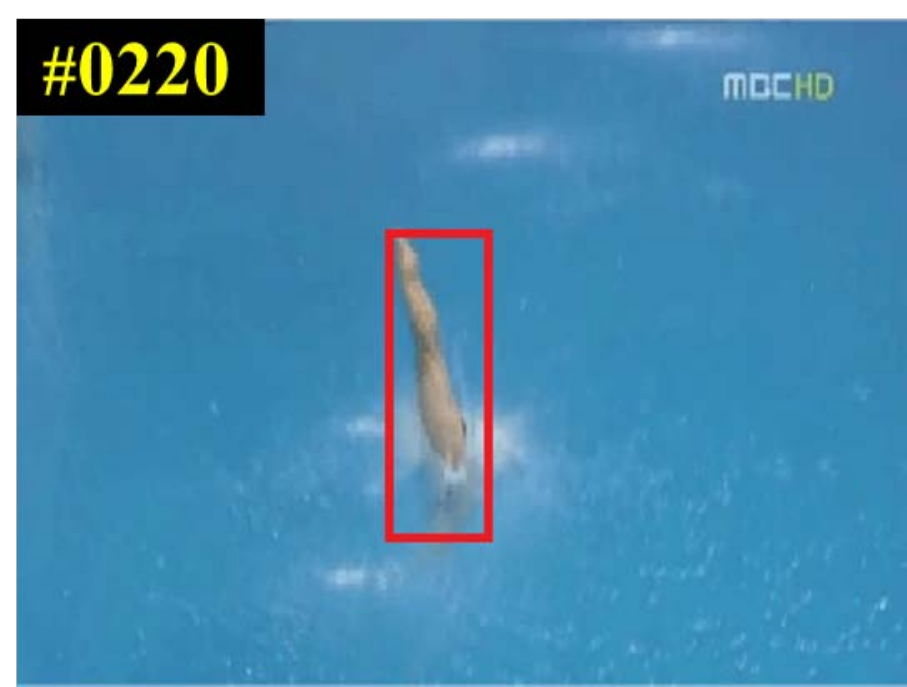} \ &
\includegraphics[width=0.14\linewidth, height=0.099\linewidth]{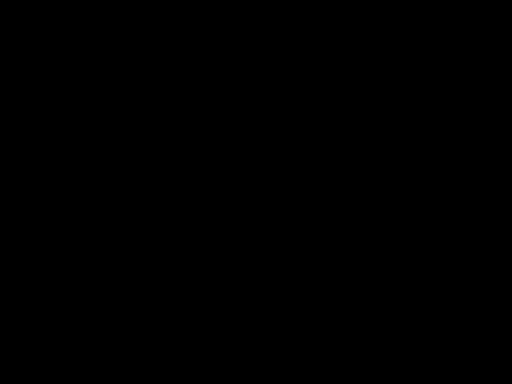} \ &
\includegraphics[width=0.14\linewidth, height=0.099\linewidth]{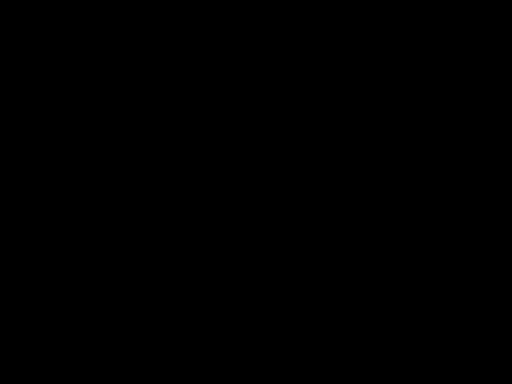} \ &
\includegraphics[width=0.14\linewidth, height=0.099\linewidth]{PT0220.png} \ &
\includegraphics[width=0.14\linewidth, height=0.099\linewidth]{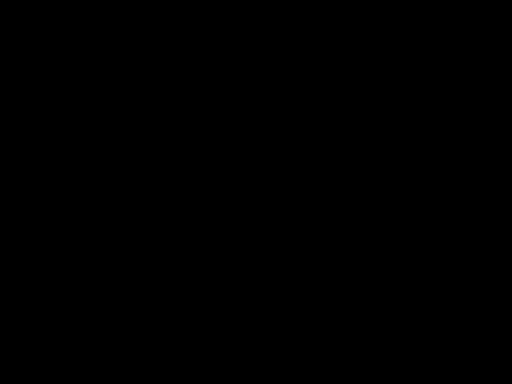} \ &
\includegraphics[width=0.14\linewidth, height=0.099\linewidth]{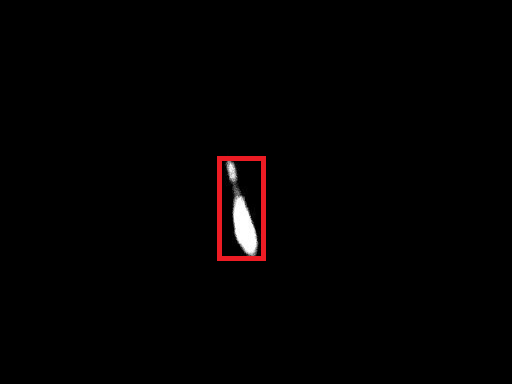} \ &
\includegraphics[width=0.14\linewidth, height=0.099\linewidth]{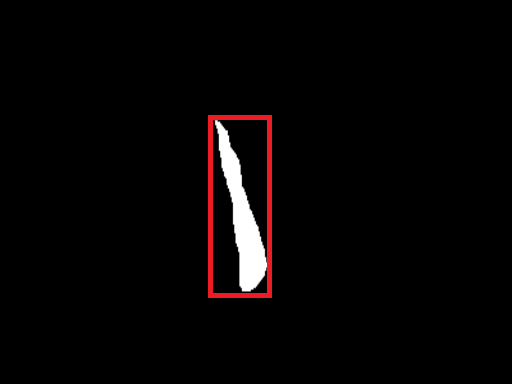} \ \\
\includegraphics[width=0.14\linewidth, height=0.099\linewidth]{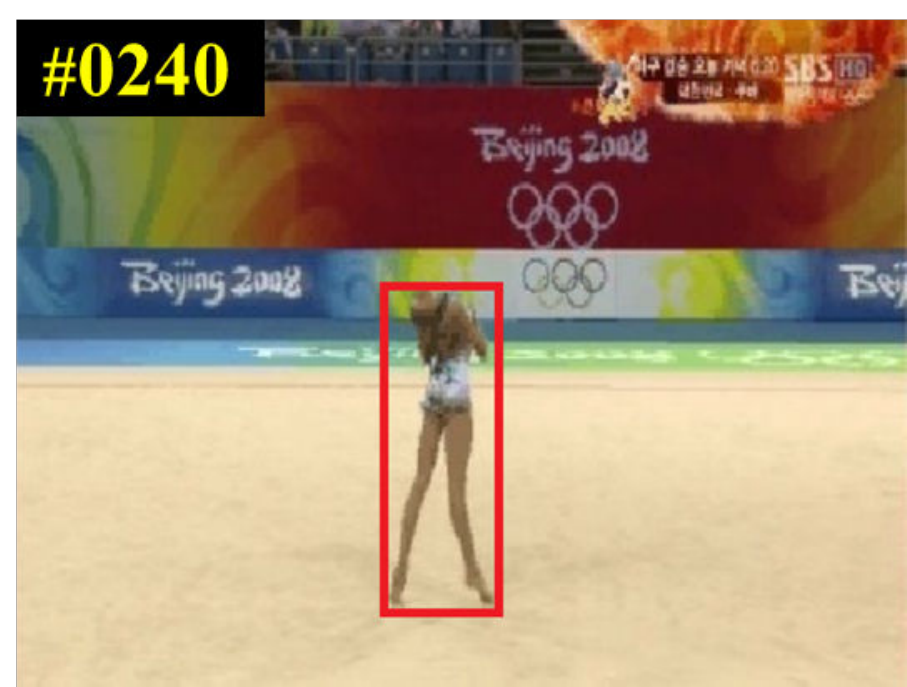} \ &
\includegraphics[width=0.14\linewidth, height=0.099\linewidth]{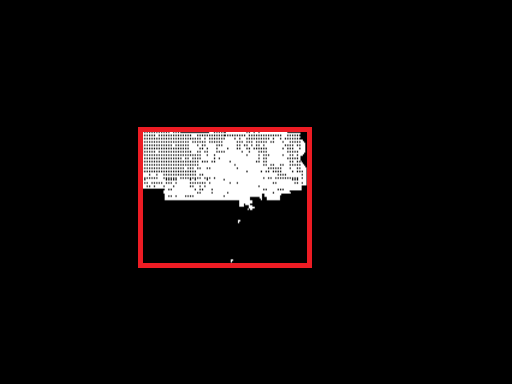} \ &
\includegraphics[width=0.14\linewidth, height=0.099\linewidth]{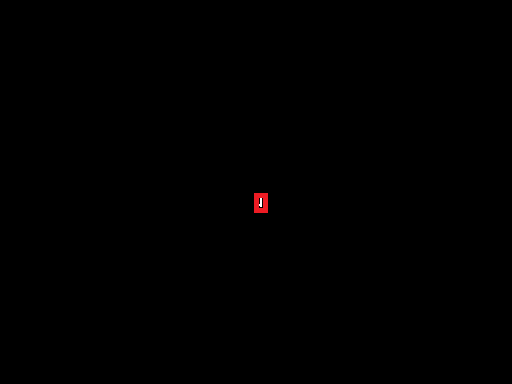} \ &
\includegraphics[width=0.14\linewidth, height=0.099\linewidth]{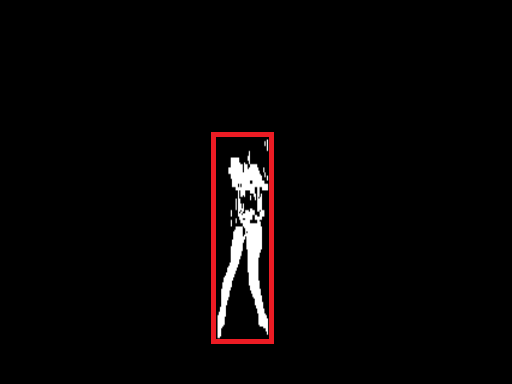} \ &
\includegraphics[width=0.14\linewidth, height=0.099\linewidth]{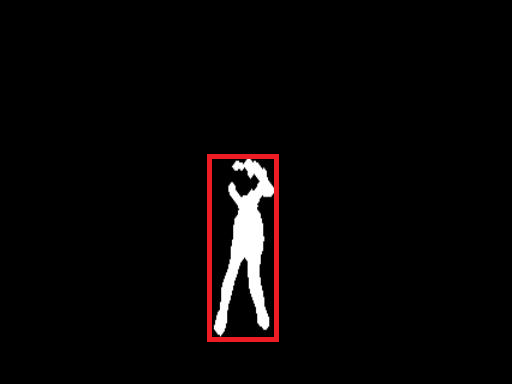} \ &
\includegraphics[width=0.14\linewidth, height=0.099\linewidth]{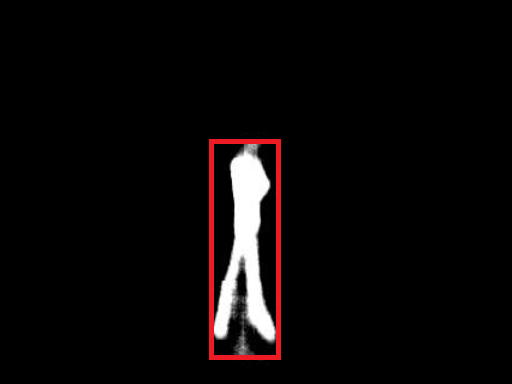} \ &
\includegraphics[width=0.14\linewidth, height=0.099\linewidth]{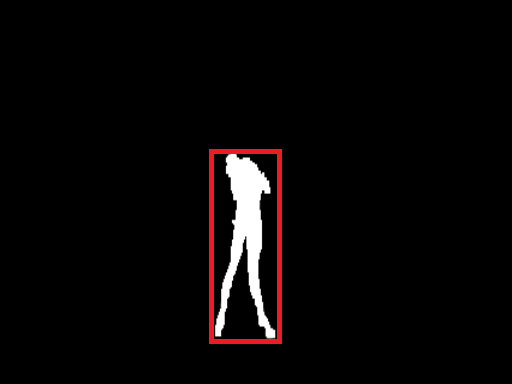} \ \\
\includegraphics[width=0.14\linewidth, height=0.099\linewidth]{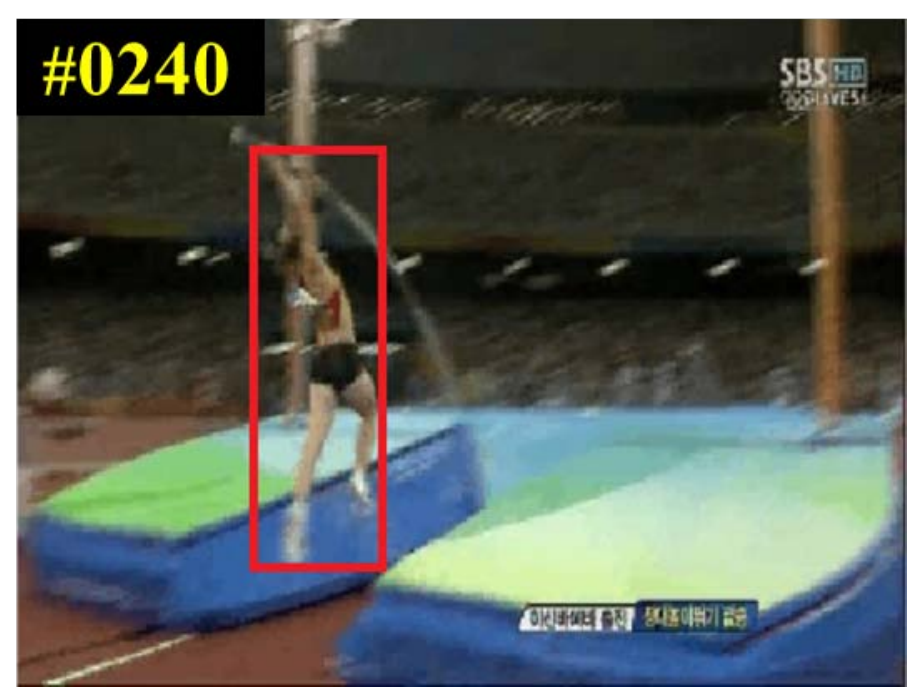} \ &
\includegraphics[width=0.14\linewidth, height=0.099\linewidth]{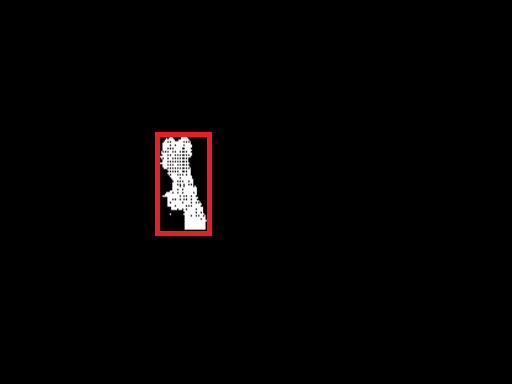} \ &
\includegraphics[width=0.14\linewidth, height=0.099\linewidth]{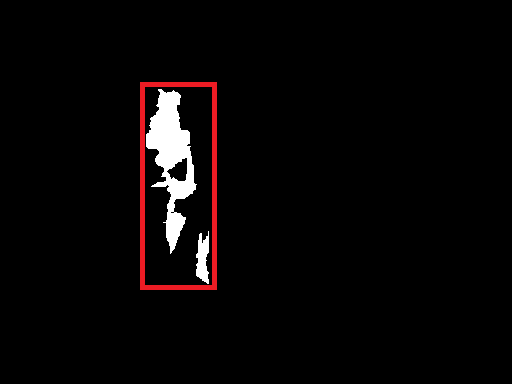} \ &
\includegraphics[width=0.14\linewidth, height=0.099\linewidth]{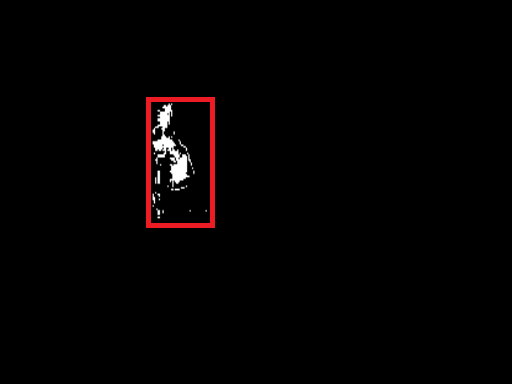} \ &
\includegraphics[width=0.14\linewidth, height=0.099\linewidth]{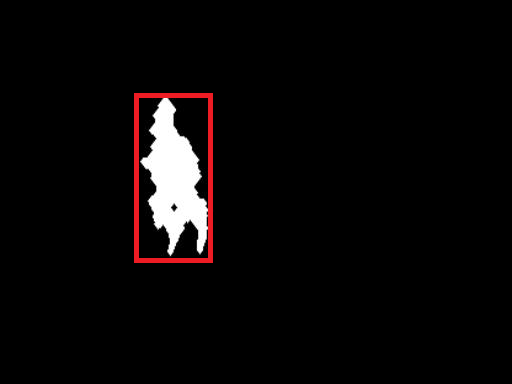} \ &
\includegraphics[width=0.14\linewidth, height=0.099\linewidth]{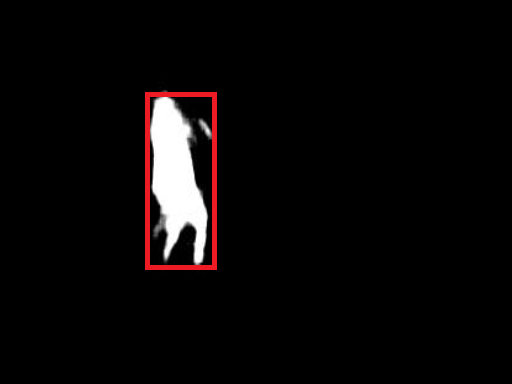} \ &
\includegraphics[width=0.14\linewidth, height=0.099\linewidth]{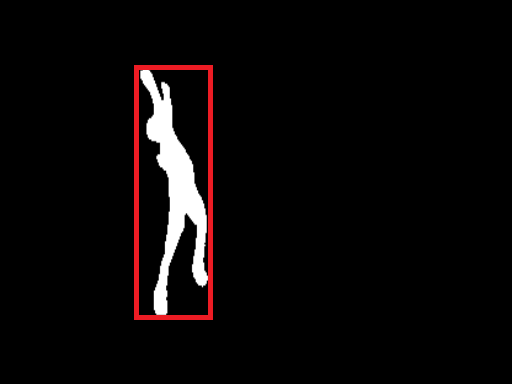} \ \\
\includegraphics[width=0.14\linewidth, height=0.099\linewidth]{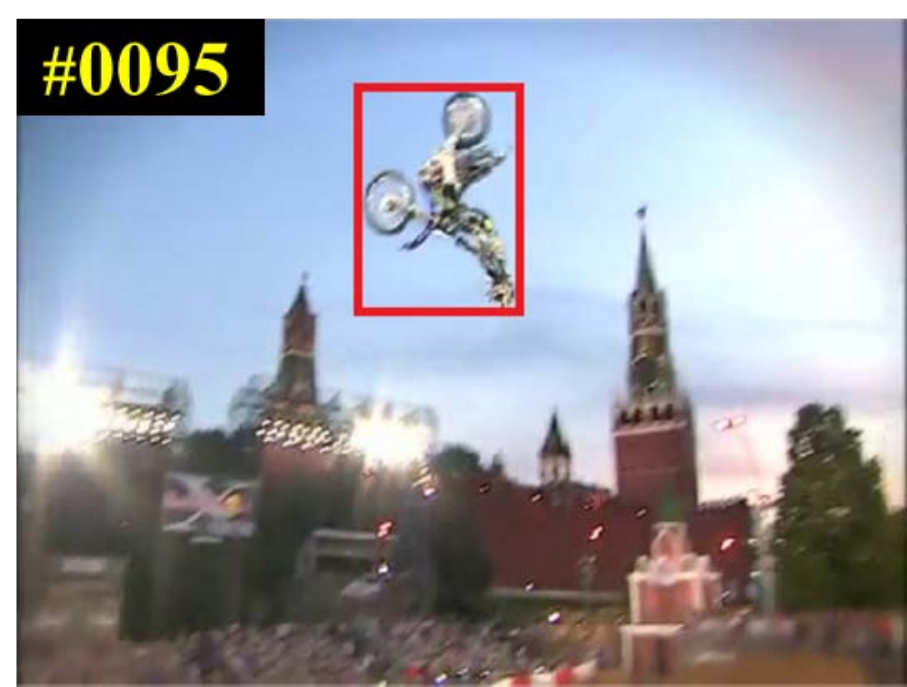} \ &
\includegraphics[width=0.14\linewidth, height=0.099\linewidth]{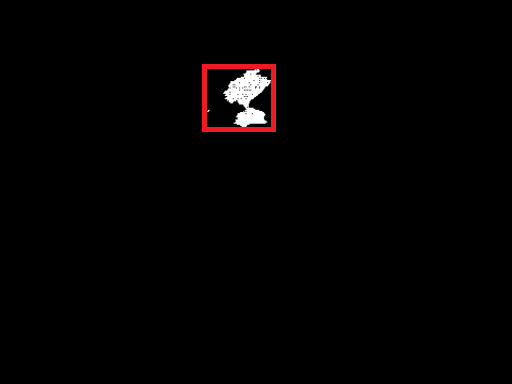} \ &
\includegraphics[width=0.14\linewidth, height=0.099\linewidth]{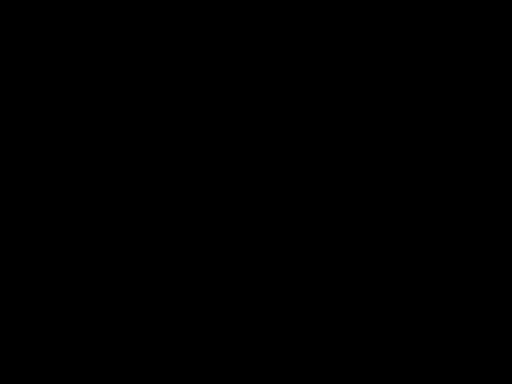} \ &
\includegraphics[width=0.14\linewidth, height=0.099\linewidth]{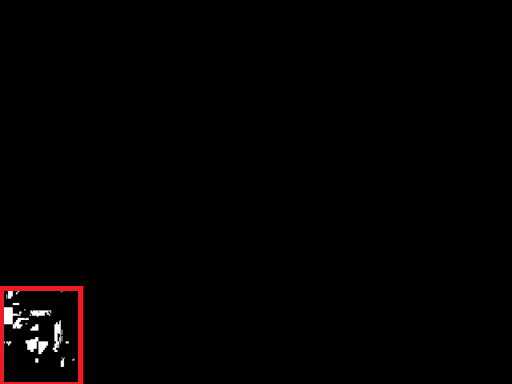} \ &
\includegraphics[width=0.14\linewidth, height=0.099\linewidth]{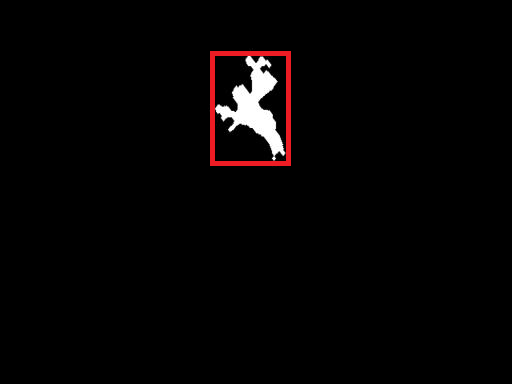} \ &
\includegraphics[width=0.14\linewidth, height=0.099\linewidth]{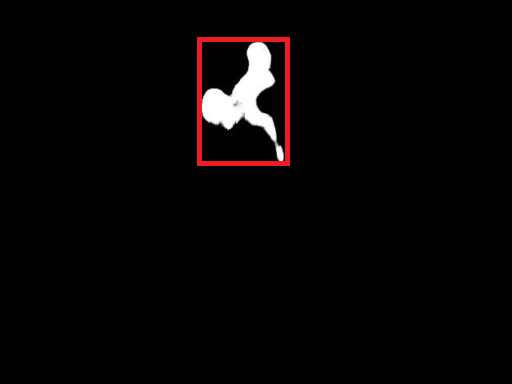} \ &
\includegraphics[width=0.14\linewidth, height=0.099\linewidth]{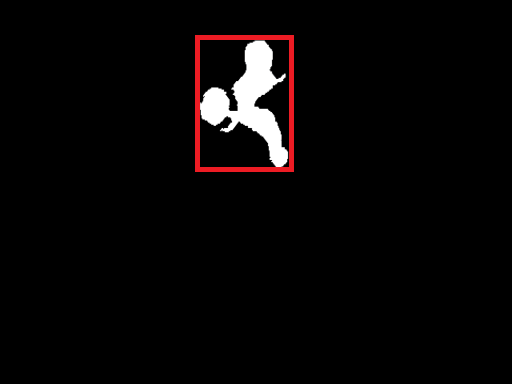} \ \\
\includegraphics[width=0.14\linewidth, height=0.099\linewidth]{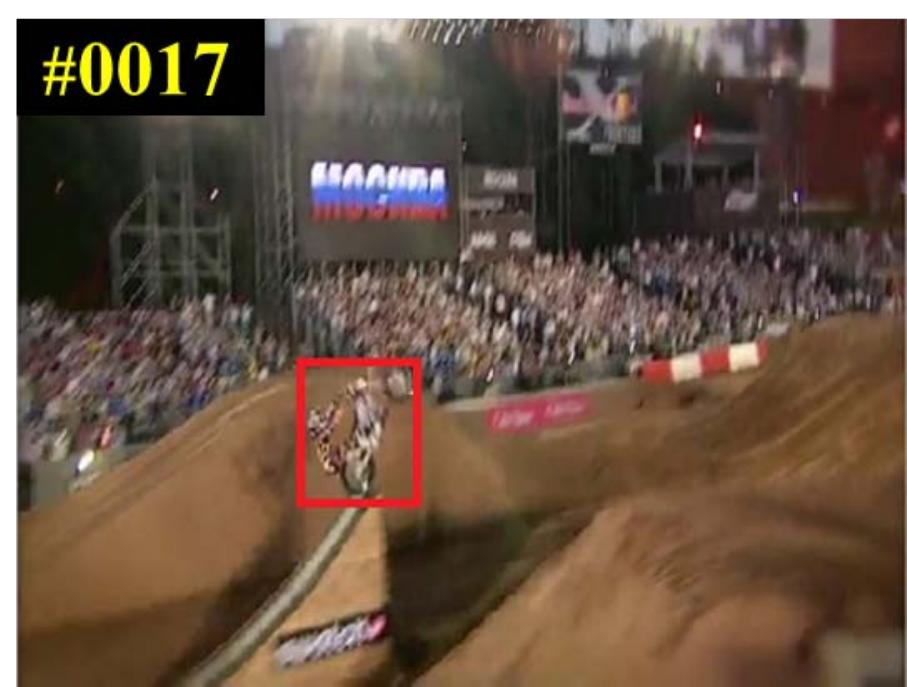} \ &
\includegraphics[width=0.14\linewidth, height=0.099\linewidth]{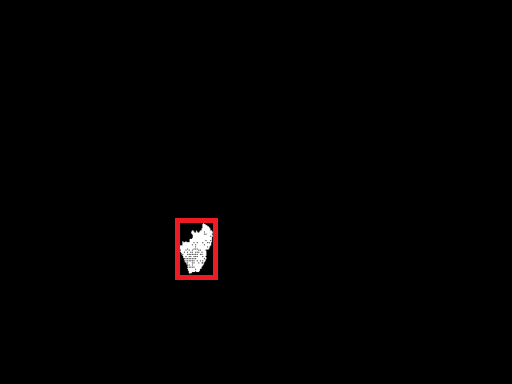} \ &
\includegraphics[width=0.14\linewidth, height=0.099\linewidth]{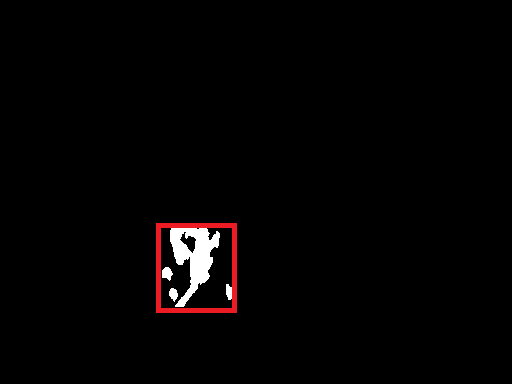} \ &
\includegraphics[width=0.14\linewidth, height=0.099\linewidth]{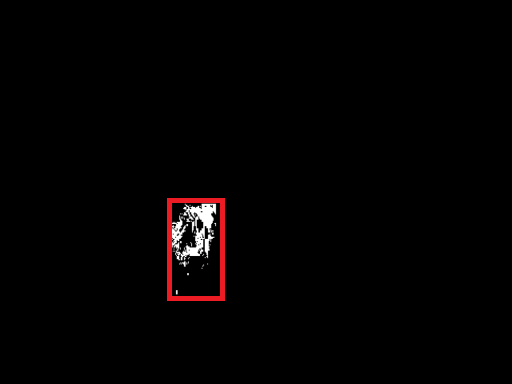} \ &
\includegraphics[width=0.14\linewidth, height=0.099\linewidth]{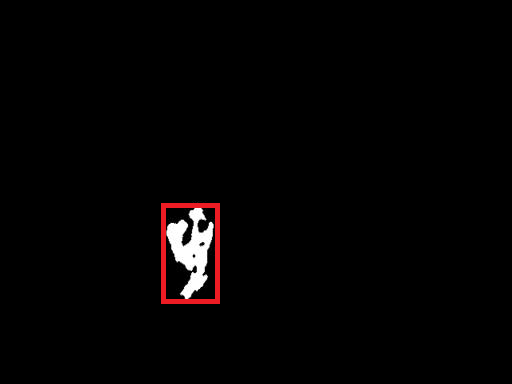} \ &
\includegraphics[width=0.14\linewidth, height=0.099\linewidth]{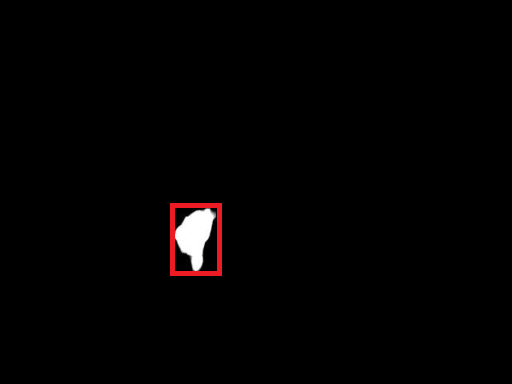} \ &
\includegraphics[width=0.14\linewidth, height=0.099\linewidth]{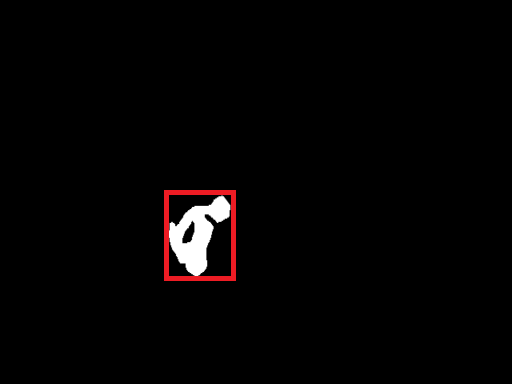} \ \\
\includegraphics[width=0.14\linewidth, height=0.099\linewidth]{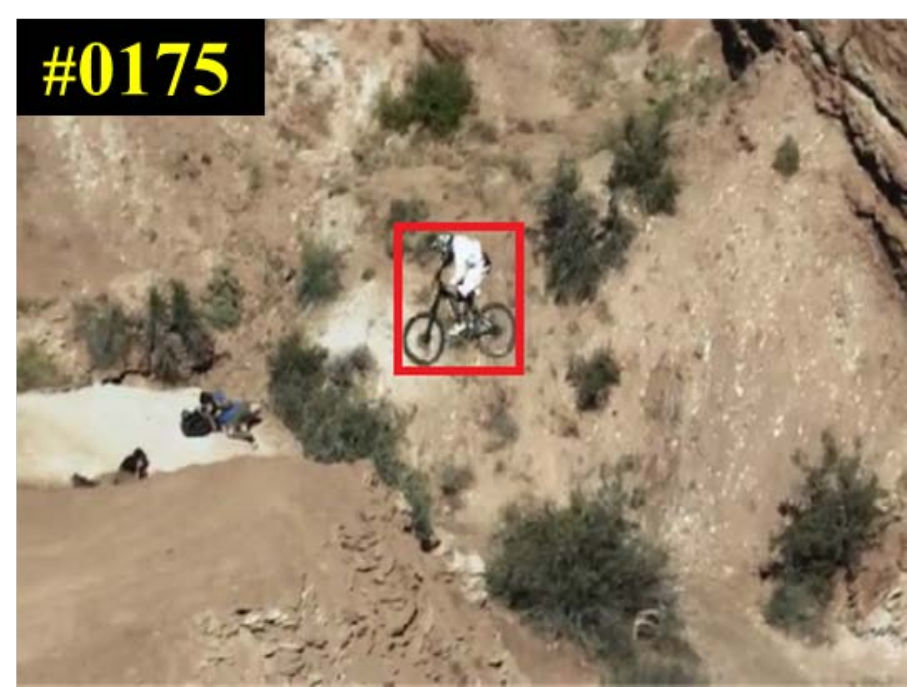} \ &
\includegraphics[width=0.14\linewidth, height=0.099\linewidth]{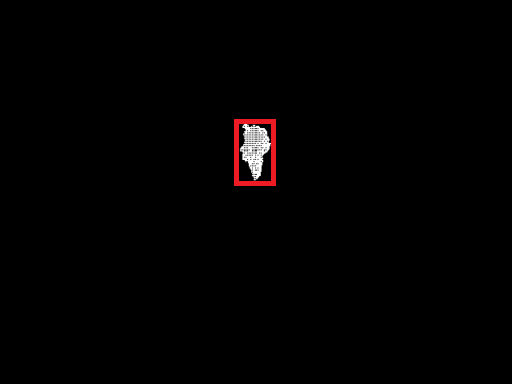} \ &
\includegraphics[width=0.14\linewidth, height=0.099\linewidth]{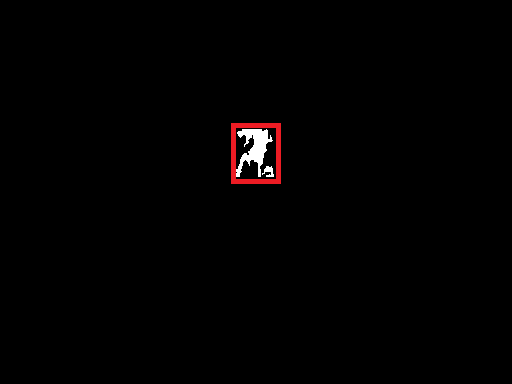} \ &
\includegraphics[width=0.14\linewidth, height=0.099\linewidth]{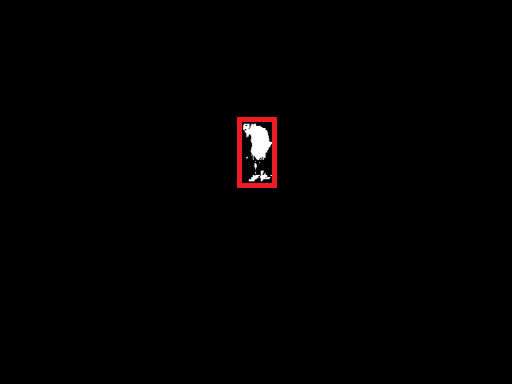} \ &
\includegraphics[width=0.14\linewidth, height=0.099\linewidth]{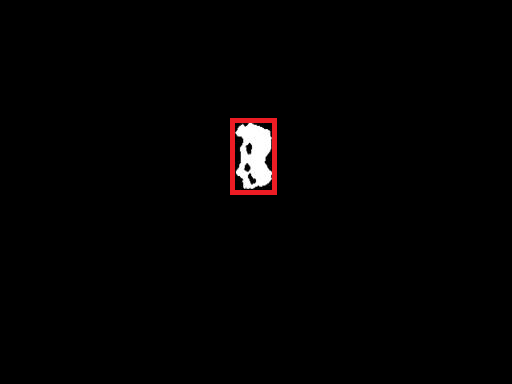} \ &
\includegraphics[width=0.14\linewidth, height=0.099\linewidth]{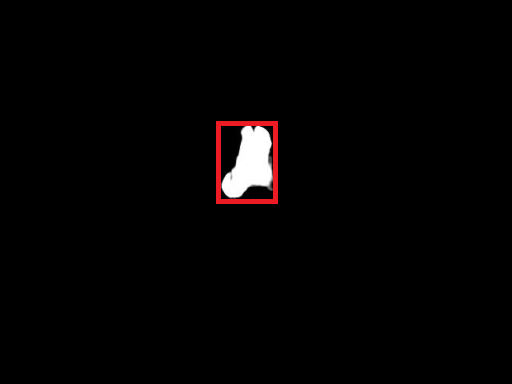} \ &
\includegraphics[width=0.14\linewidth, height=0.099\linewidth]{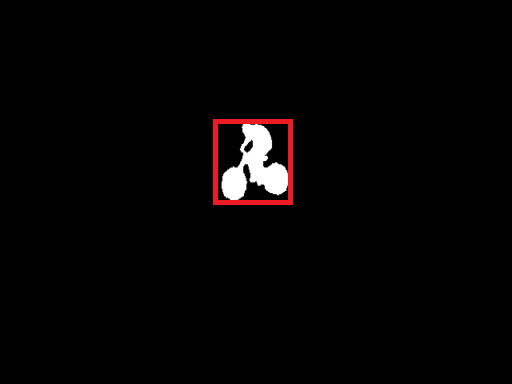} \ \\
\includegraphics[width=0.14\linewidth, height=0.099\linewidth]{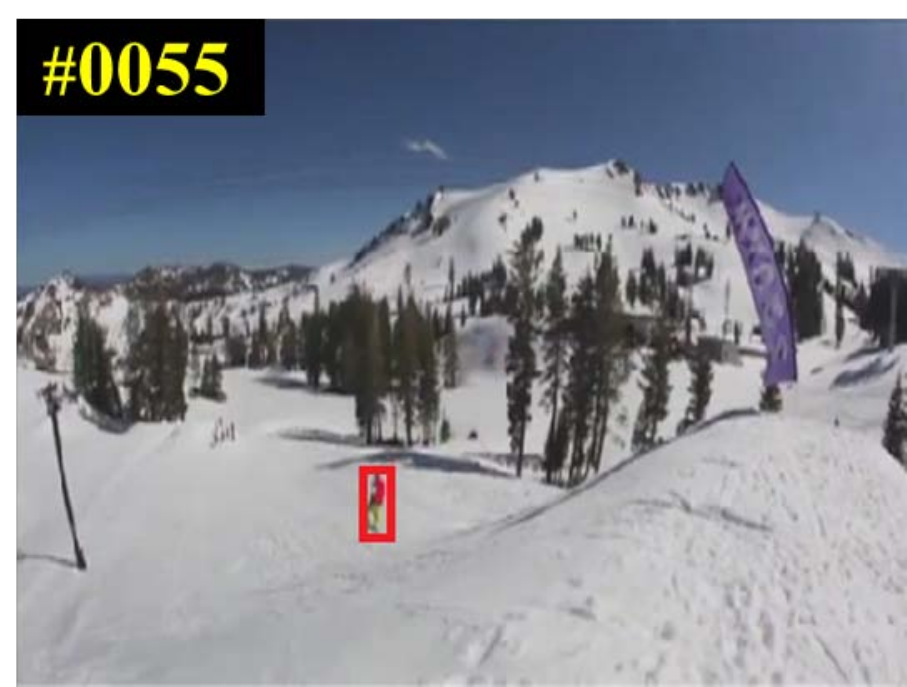} \ &
\includegraphics[width=0.14\linewidth, height=0.099\linewidth]{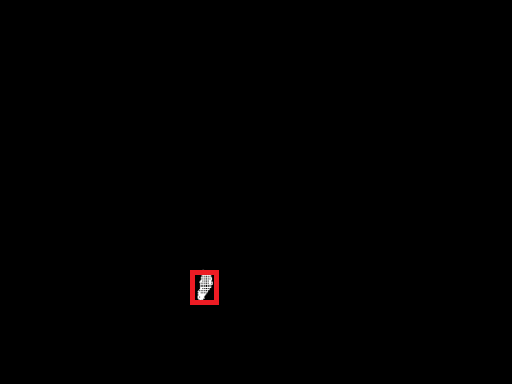} \ &
\includegraphics[width=0.14\linewidth, height=0.099\linewidth]{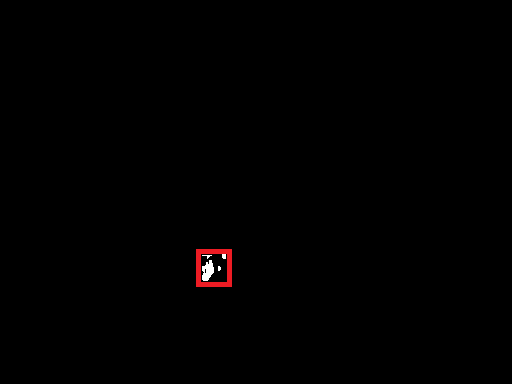} \ &
\includegraphics[width=0.14\linewidth, height=0.099\linewidth]{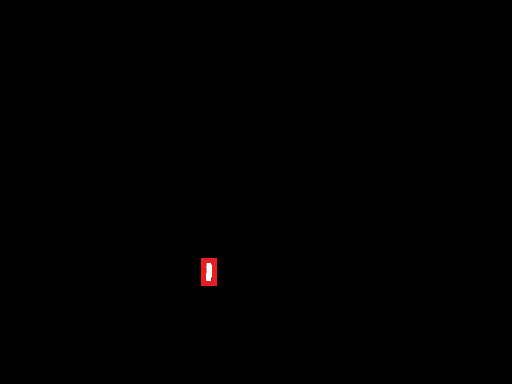} \ &
\includegraphics[width=0.14\linewidth, height=0.099\linewidth]{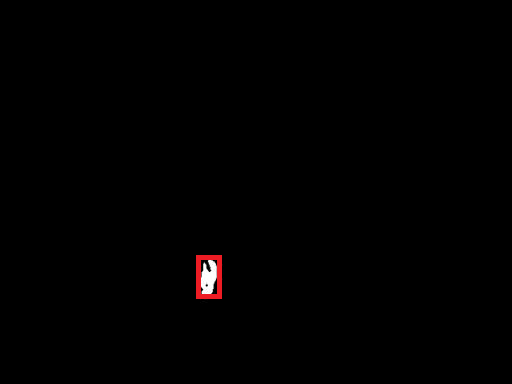} \ &
\includegraphics[width=0.14\linewidth, height=0.099\linewidth]{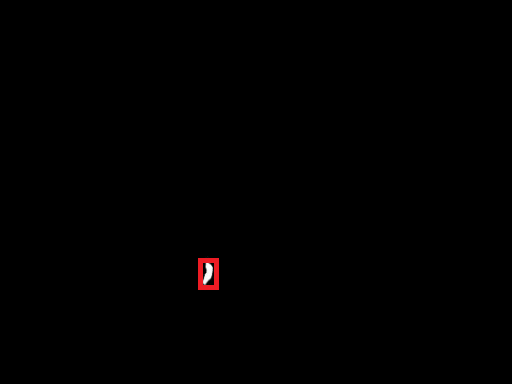} \ &
\includegraphics[width=0.14\linewidth, height=0.099\linewidth]{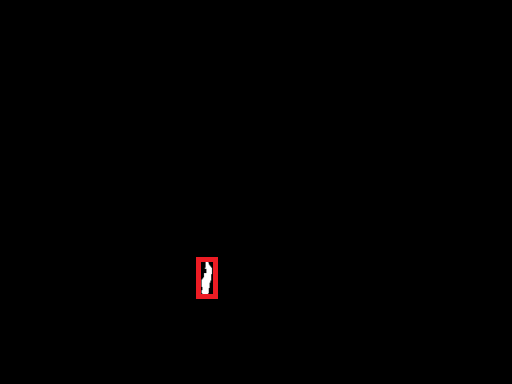} \ \\
\includegraphics[width=0.14\linewidth, height=0.099\linewidth]{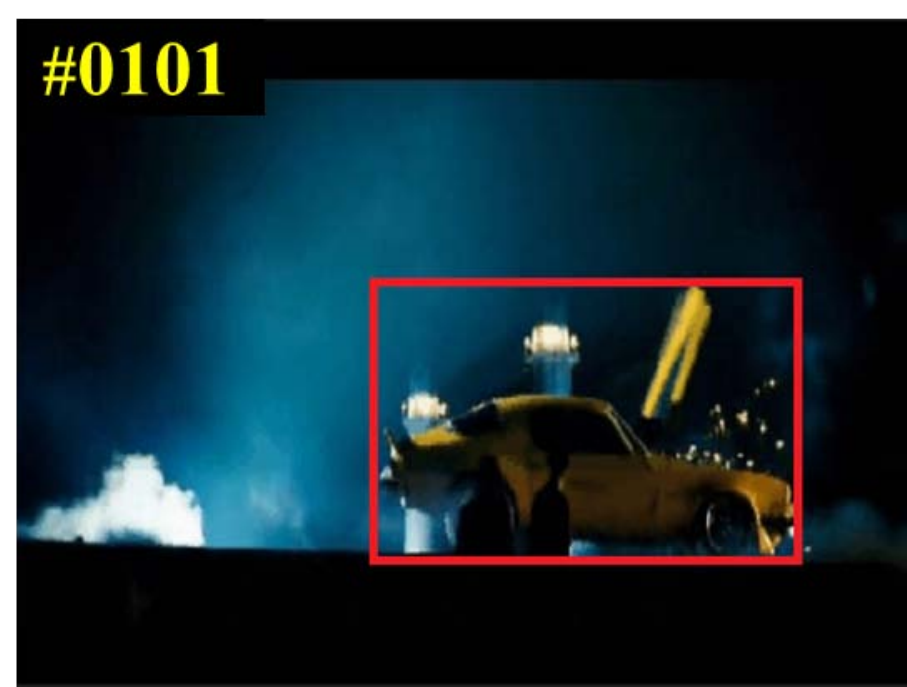} \ &
\includegraphics[width=0.14\linewidth, height=0.099\linewidth]{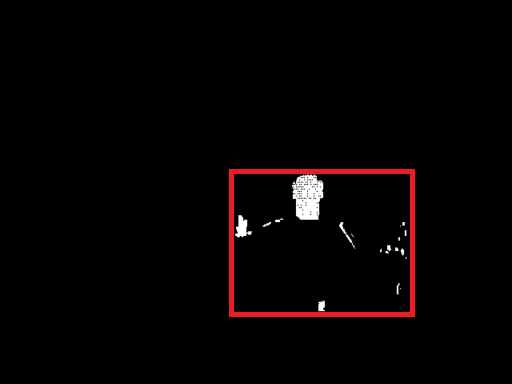} \ &
\includegraphics[width=0.14\linewidth, height=0.099\linewidth]{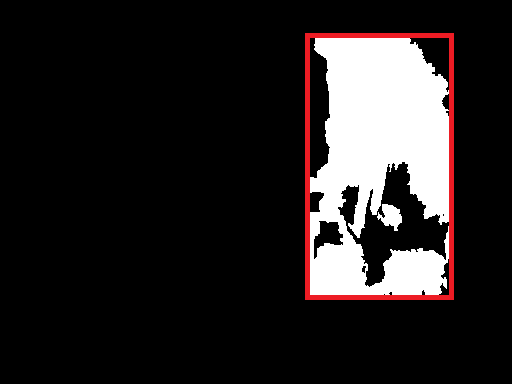} \ &
\includegraphics[width=0.14\linewidth, height=0.099\linewidth]{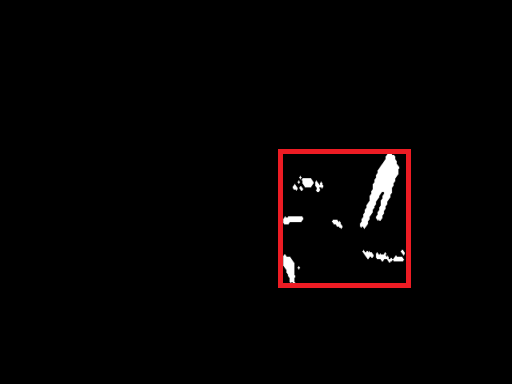} \ &
\includegraphics[width=0.14\linewidth, height=0.099\linewidth]{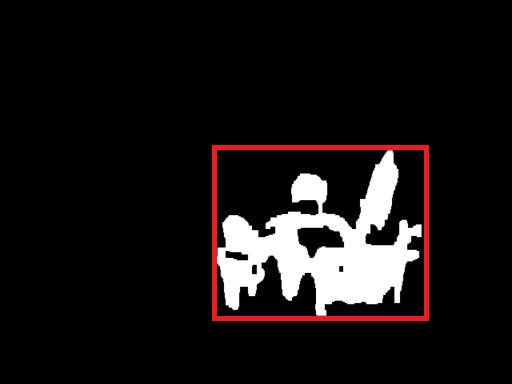} \ &
\includegraphics[width=0.14\linewidth, height=0.099\linewidth]{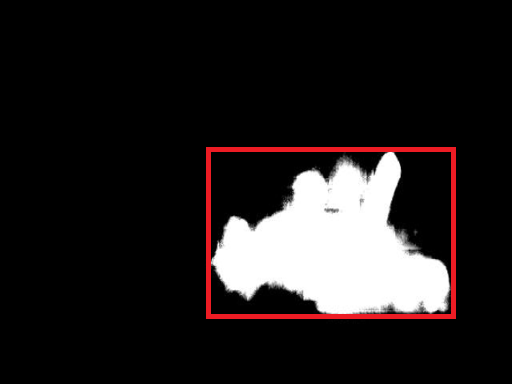} \ &
\includegraphics[width=0.14\linewidth, height=0.099\linewidth]{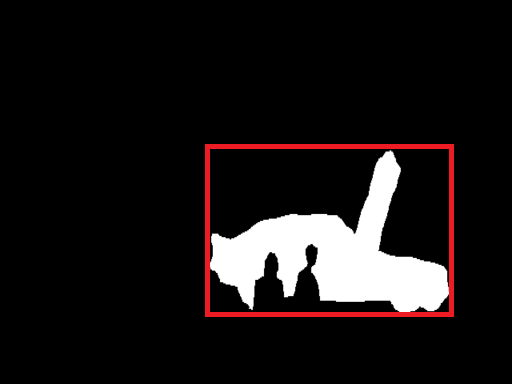} \ \\
\includegraphics[width=0.14\linewidth, height=0.099\linewidth]{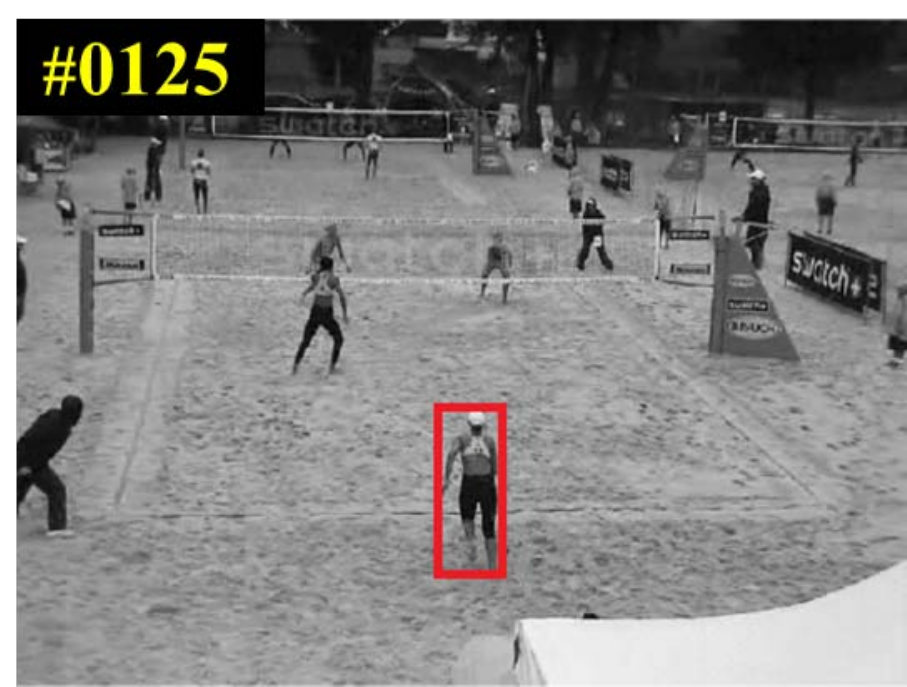} \ &
\includegraphics[width=0.14\linewidth, height=0.099\linewidth]{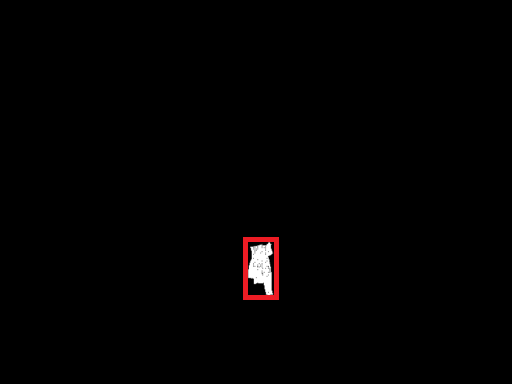} \ &
\includegraphics[width=0.14\linewidth, height=0.099\linewidth]{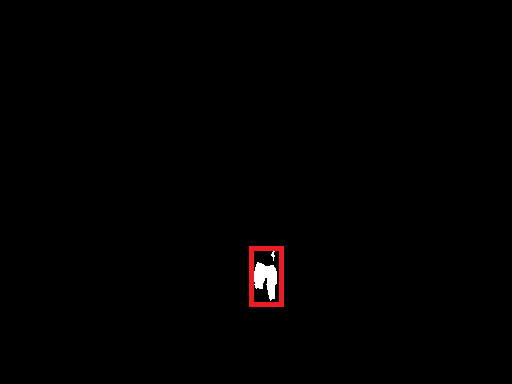} \ &
\includegraphics[width=0.14\linewidth, height=0.099\linewidth]{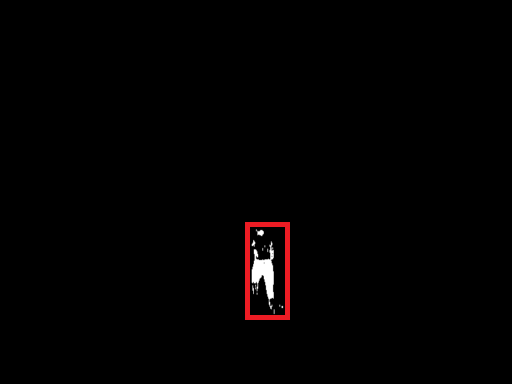} \ &
\includegraphics[width=0.14\linewidth, height=0.099\linewidth]{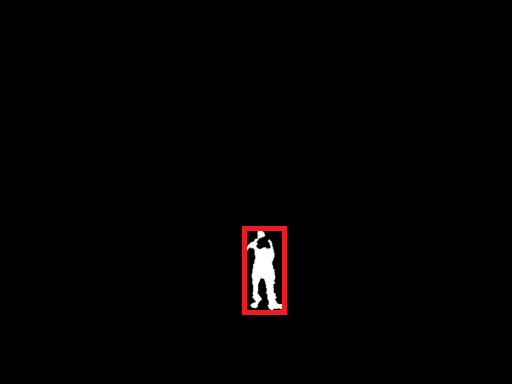} \ &
\includegraphics[width=0.14\linewidth, height=0.099\linewidth]{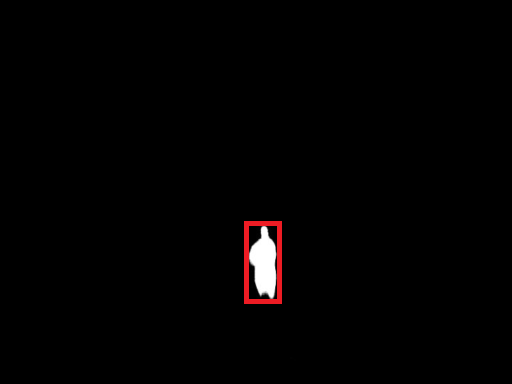} \ &
\includegraphics[width=0.14\linewidth, height=0.099\linewidth]{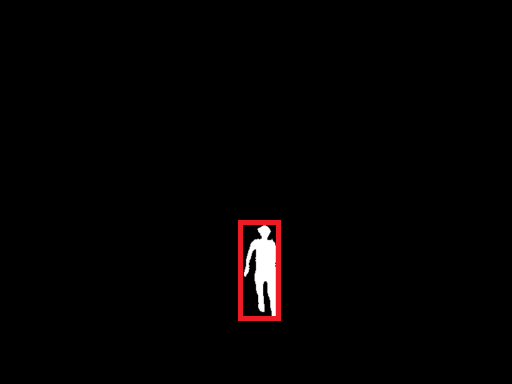} \ \\
(a) & (b) & (c) & (d) & (e) & (f) & (g)\\ \\
\end{tabular}
\caption{Qualitative evaluation of the proposed tracker and other segmentation-based methods.
From left to right: (a) input frames; (b) HT~\cite{hough}; (c) SPT~\cite{Superpixel};
(d) PT~\cite{Pixeltrack}; (e) OGBDT~\cite{ogbdt}; (f) our method; and
(g) ground truth.
\label{fig:trackingexamples}}
\end{figure*}

Similar to~\cite{ogbdt}, we adopt two overlap ratio rules to evaluate the proposed method and other competing ones.
The bounding box overlap ratio is used to compare all trackers; whereas the segmentation overlap ratio is adopted for
comparing non-rigid tracking algorithms with segmentation outputs.
The average overlap ratio results are demonstrated in Table.~\ref{table:bboverlap}.
%
From Table.~\ref{table:bboverlap}, we have two fundamental observations: (1) other deep learning-based trackers have not taken considered segmentation, however, their performance on this non-rigid dataset are still competitive; (2) the proposed method outperforms
all compared algorithms in both bounding box and segmentation overlap ratios of most sequences.
Therefore, our tracker is more suitable for tracking deformable and articulated objects.
Fig.~\ref{fig:trackingexamples} demonstrates the representative screenshots of the proposed trackers and other segmentation-based ones, which shows
that our tracker achieves smoother visual effects and includes more semantic contexts.
More specifically, the semantic information and clear boundary of the tracking
targets are highlighted in most sequences, such as \emph{Gymnastics, High-jump} and \emph{Motocross1}. It is worth noted that in the challenging sequences such as \emph{Diving, MotorRolling} and \emph{Transformer}, most methods fail to track targets well whereas our algorithm performs accurately in terms of either precision or overlap; For the most challenging sequence, i.e, \emph{diving}, other trackers fail to track target and can not re-locate the target. Our tracker can automatically recover the tracking target.
\subsubsection{Evaluation on Different Components}
To further verify the contribution of each component in our model, we also implement different variants of the proposed method and report their average segmentation overlap ratios in Table.~\ref{table:bboverlap}.
These variants include the following methods: (1) $Ours_{1}$ denotes the proposed algorithm without fusing scale-dependent saliency maps, which merely uses the ROI as the input. (2) $Ours_{2}$ stands for our method only using the current saliency map without the accumulated operation. (3) $Ours_{3}$ simultaneously removes the local saliency fusion and the accumulated operation.
Comparison results of $Ours_{1}$ and $Ours_{2}$ with $Ours_{3}$ show that our proposed local saliency map detection method can significantly improve the tracking performance.
Both strategies of scale-dependent saliency maps and accumulated operations boost the tracking performance from 36.8 to 55.0.
These results demonstrate the effectiveness of our method in both accuracy and robustness.
\subsubsection{Generalization Ability}
To demonstrate the generalization ability of our tracker, we evaluate the proposed algorithm by using the OTB-50 benchmark~\cite{otb} including both rigid and non-rigid objects (most of them are rigid or approximatively rigid).
Our tracker achieves comparable results (\emph{precision} = 0.842, \emph{success rate} = 0.561).
For one thing, our tracker performs better than the second best non-rigid tracker (i.e., OGBT~\cite{ogbdt}) in the OTB-50 benchmark.
The performance of the OGBT method is as follows: \emph{precision} = 0.748, \emph{success rate} = 0.524.
For another, the proposed method is not the best compared with the latest ones reported in the benchmark; nevertheless, it is still competitive.
We note that our tracker can generate pixel-wise saliency maps, which is more difficult and useful than the outputs of the
bounding box tracking.
\section{Conclusion}
In this paper, we present a novel non-rigid object tracking method based on STCSM.
First, we develop a TFCN model to introduce the local saliency prior for a given image region.
Second, a multi-scale multi-region mechanism is exploited to generate multiple local saliency maps, which are further effectively fused into a final discriminative saliency map through a weighted entropy method.
In addition, a non-rigid object tracker is designed by using the STCSM model.
Finally, the proposed model can be fine-tuned to update the tracker and capture the appearance change of
the tracked object.
The experimental results show that the proposed method not only performs significantly better than other related trackers for tracking non-rigid objects but also achieves competitive performance in dealing with traditional saliency detection and visual tracking problems.

\bibliographystyle{IEEEtran}

\begin{IEEEbiography}{\textbf{Pingping Zhang}}
received his B.E. degree in mathematics and applied mathematics, Henan Normal University (HNU), Xinxiang, China, in 2012. He is currently a Ph.D. candidate in the School of Information and Communication Engineering, Dalian University of Technology (DUT), Dalian, China. His research interests are in deep learning, saliency detection, object tracking and semantic segmentation.
\end{IEEEbiography}
\begin{IEEEbiography}{\textbf{Dong Wang}}
received the B.E. degree in electronic information engineering and the Ph.D. degree in signal and information processing from the Dalian University of Technology (DUT), Dalian, China, in 2008 and 2013, respectively. He is currently a Faculty Member with the School of Information and Communication Engineering, DUT. His current research interests include face recognition, interactive image segmentation, and object tracking.
\end{IEEEbiography}
\begin{IEEEbiography}{\textbf{Huchuan Lu}}
(SM'12) received the M.Sc. degree in signal and information processing, PhD degree in system engineering, Dalian
University of Technology (DUT), China, in 1998 and 2008 respectively. He has been a faculty since 1998 and a professor since 2012 in the School of Information and Communication Engineering of DUT. His research interests are in the areas of computer vision and pattern recognition. In recent years, he focus on visual tracking, saliency detection and semantic segmentation. Now, he serves as an associate editor of the IEEE Transactions On Systems, Man, and Cybernetics: Part B.
\end{IEEEbiography}
\begin{IEEEbiography}{\textbf{Hongyu Wang}}
(M'98) received the B.S. degree from Jilin University of Technology, Changchun, China, in 1990 and the M.S. degree from the Graduate School of Chinese Academy of Sciences, Beijing, China, in 1993, both in electronic engineering. He received the Ph.D. degree in precision instrument and optoelectronics engineering from Tianjin University, Tianjin, China, in 1997. He is currently a Professor with Dalian University of Technology, Dalian, China. His research interests include algorithmic, optimization, and performance issues in wireless ad hoc, mesh, and sensor networks.
\end{IEEEbiography}
\end{document}